%% file: Thesis.tex
\documentclass[10pt]{ociamthesis} 
\usepackage{amssymb}
\usepackage{sectsty}
\allsectionsfont{\sf\bfseries}

\title{\sf\bfseries{Two Projection Pursuit Algorithms for Machine Learning under Non-Stationarity}}

\author{Duncan~A.~J.~Blythe}             
\college{Machine Learning Group - TU Berlin\\
Bernstein Focus: Neurotechnology Berlin} 

\degree{MSc in Computational Neuroscience}  
\degreedate{September 2011}      
\usepackage[latin1]{inputenc}
\usepackage[T1]{fontenc}
\usepackage{graphicx}
\usepackage{bm}
\usepackage{algpseudocode}
\usepackage{amsmath}
\usepackage{dsfont}
\usepackage{hyperref}
\usepackage{amssymb}
\usepackage{graphicx}
\usepackage{pifont}
\usepackage{url}
\usepackage{amsfonts}
\usepackage{amscd,amsthm}
\usepackage{algorithm}
\usepackage{setspace}
\usepackage{fancyhdr}
\usepackage[margin=2.5cm]{geometry}
\newcommand{\tickYes}{\checkmark}
\newcommand{\tickNo}{\hspace{1pt}\ding{55}}
\newcommand{\G}{\ensuremath{\mathcal{G}}}
\newcommand{\HH}{\ensuremath{\mathcal{H}}}
\newcommand{\NNN}{\ensuremath{\mathcal{N}}}

\newcommand{\mmu}{\bm{\hat{\mu}_i}}

\newcommand{\covar}{\hat{\Sigma}_i}

\theoremstyle{plain}

\theoremstyle{definition}

\input ssa_defs.tex

\begin{document}

\baselineskip=18pt plus1pt

\setcounter{secnumdepth}{3}
\setcounter{tocdepth}{3}
\onehalfspace

\maketitle                  

\fancyhead[RO]{{\footnotesize\rightmark}\hspace{2em}\thepage}
\setcounter{tocdepth}{2}
\fancyhead[LE]{\thepage\hspace{2em}\footnotesize{\leftmark}}
\fancyhead[RE,LO]{}
\fancyhead[RO]{{\footnotesize\rightmark}\hspace{2em}\thepage}

\begin{romanpages}          

\chapter*{Acknowledgements}
\addcontentsline{toc}{chapter}{Acknowledgements}

I am most grateful to Professor K.-R.M{\"u}ller for having provided me the opportunity, not only to gain research experience in his IDA/Machine Learning
Laboratory but in addition for employing me to do so. Without his assistance I would not have been able to complete my degree at the BCCN
and continue to the PhD level in a dignified manner. I have greatly appreciated having been treated as an equal member of his research group,
despite my Master's student's status, in being assigned research tasks of genuine scientific import and interest.
In addition, I most heartily thank Franz Kiraly, Paul von B{\"u}nau, Frank Meinecke and Wojciech Samek for imparting to me their knowledge 
of Machine Learning and Neuroscience during my first two years at the TU Berlin
and for their enthusiasm for the topics which I have pursued in this thesis. 
Chapters 2 and 3 of this thesis are, in addition, the result of collaboration with Paul, Frank and Wojciech.
In addition I would like to thank Franz Kiraly and Wojciech Samek, once again, as well 
as Alex Schlegel and Danny Pankin for their comments on the manuscript. Moreover, 
I thank the remaining, numerous members of the Machine Learning Group and the BCCN 
with whom I have discussed ideas relating to Neuroscience, Machine Learning, Computer Science and Mathematics.
In addition, I thank Andrea Gerdes, Vanessa Casagrande and Margret Franke for assisting
me in the demanding task of completing my degree within the allotted two year period.
Finally, I would like to thank my parents, for their continued support during the transition I have made to this field and to my girlfriend Michaela
for helping me not to lose sight of the idealism which originally brought me to this field.

\tableofcontents            
\end{romanpages}            

\include{chapter2}
\include{conclusions}

\include{appendix1}
\include{appendix2}

\chapter{Introduction to Linear Algorithms Under Non-Stationarity}

Non-stationarity of a stochastic process is defined loosely as variability of probability distribution over time. Conversely, stationarity of a stochastic process corresponds to 
constancy of distribution. In machine learning, typical tasks include regression, classification or system identification. For regression and classification, no guarantee on 
generalization, from a training set to a test set, may be made under the assumption that the underlying process, yielding the training and test sets, is non-stationary.
Thus quantification of non-stationarity and algorithms for choosing features which are stationary are indispensable for these tasks. On the other hand, in system identification, a stationary 
or maximally non-stationary subsystem often carries important significance in terms of the primitives of the domain under consideration: for example, in 
neuroscience, identification of the subsystem of the dynamics over synaptic weights with a neural network with $n$ weights consisting of the subset of $m<n$ weights whose distribution is maximally non-stationary under learning is of vital interest to analysis of the neural substrate underlying the learning process.






\section{Survey on Projection Algorithms under the Non-Stationarity Assumption}

The classical literature on machine learning and statistical learning theory assumes that the samples used for training the parameters of, for instance, classifiers, are drawn from a single
probability distribution, rather than a, possibly non-stationary, process \cite{bb58133}. That is to say, the time series from which the data are taken is stationary over time. Thus, approaches to learning which 
relax this stationarity assumption have been sparse up until the present time within the Machine Learning literature. In particular,
the first algorithm, Stationary Subspace Analysis (SSA) for obtaining a linear stationary projection of a data set was published only recently in 2009 \cite{PRL:SSA:2009}. 
Since 2009 algorithms have published which refine SSA computationally 
\cite{HarKawWasBun10SSA} under assumptions, develop a maximum likelihood approach to SSA \cite{InfoSSA} and derive an algebraic algorithm for its solution \cite{JMLRpreprint}. In addition to SSA, linear algorithms
have been published for applications, for example, sCSP for Brain Computer Interfacing \cite{5946469}. However, non-stationarity in Machine Learning remains largely an open problem: for instance, no general technique for classification under non-stationarity has been proposed which is non-adaptive. (Adaptation has been studied in some detail, for instance, for an adaptive neural network for regression or classification, see \cite{AdaptNN}; otherwise only the covariate shift problem has been studied in detail by, for instance, \cite{CovariateShift}.) Methods which seek to perform robust classification under
non-stationarity remain to be fully investigated.

\section{Overview and Layout of this Thesis}
The present thesis's contribution is twofold: firstly in Chapter~\ref{CPD} we present a method based on Stationary Subspace Analysis (SSA)  \cite{PRL:SSA:2009} which addresses the system identification task described 
above, namely identifying a maximally non-stationary subsystem. We subsequently apply this method to a specific machine learning task, namely Change Point Detection. In particular, we show that
using the method based on SSA as a prior feature extraction step to Change Point Detection, boosts the performance of three representative Change Point Detection algorithms on synthetic data and data
adapted from real world recordings. Note that this chapter (Chapter~\ref{CPD}) consists in part of joint work of the present author and the authors of the following submitted paper: \cite{CPpreprint}, of which the Chapter is an adapted version.
Secondly, in Chapter~\ref{ChapSup}, we address the two-fold classification problem under non-stationarity and propose a method for adapting a commonly used method for two-fold classification, namely Linear Discriminant Analysis (LDA) for this setting: we call the resulting algorithm \emph{stationary} Linear Discriminant Analysis, or sLDA for short. 
We investigate the properties and performance
 of the algorithm on simulated data and on data recorded for Brain Computer Interface experiments. Finally having tested the algorithm we perform a rigorous investigation of the results obtained
 by means of statistical testing and comparison with base-line methods. Please note, in addition, that Chapter~\ref{ChapSup} includes joint work with Wojciech Samek.

\chapter{Maximizing Non-Stationarity with Applications to Change Point Detection}
\label{CPD}
\section{Introduction}

Change Point Detection is a task that appears in a broad 
range of applications such as biomedical signal processing \cite{Biomed2, EEGPrin, KohlmorgenKybernetics}, 
speech recognition \cite{Speech1,Speech2}, industrial process monitoring \cite{ChangePoint, Narendra}, 
fault state Detection \cite{Fault1} and econometrics \cite{Econ1}. The goal of Change Point 
Detection is to find the time points at which a time series changes from one macroscopic state to another. 
As a result, the time series is decomposed into segments \cite{ChangePoint} of similar
behavior. Change point detection is based on finding changes in the properties of 
the data, such as in the moments (mean, variance, kurtosis) \cite{ChangePoint}, in the spectral properties \cite{Spectral}, 
temporal structure \cite{DistBased2} or changes w.r.t.~to certain patterns \cite{Pattern}.
The choice of any of these aspects 
depends on the particular application domain and on the statistical type of the changes that one
aims to detect. 

\begin{figure}[ht]
 \begin{center}
  \includegraphics{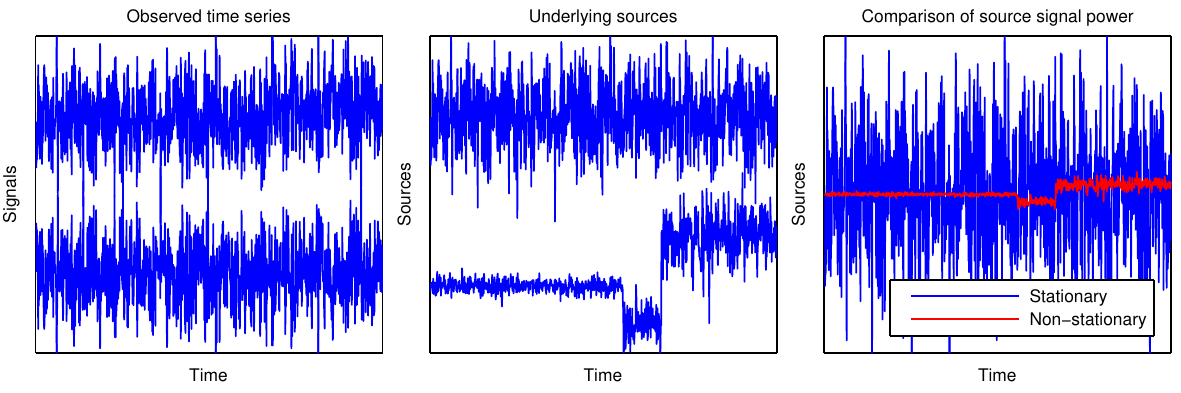}
  \caption{
  	Informative vs.~uninformative directions for Change Point Detection. The left panel shows
	the observed bivariate time series where no pronounced changes are visible. The middle panel shows
	the two underlying sources, where one of them exhibits clearly visible changes. In the right panel, 
	we see that the stationary sources has much higher signal power than the informative non-stationary 
	sources and thus masks the presence of change-points in the observed data. 
    \label{fig:tempseg_intro}
    }
 \end{center}
\end{figure}

For a large family of general segmentation algorithms, state changes are detected by
comparing the empirical distributions between windows of the time series 
\cite{DistBased1, DistBased4,DistBased2}. Estimating and comparing probability densities 
is a difficult statistical problem, particularly in high dimensions.
Often, however, many directions in a high dimensional signal space are uninformative for Change Point 
Detection: in many cases there exists a subspace in which the distribution of the data remains constant 
over time (i.e. is stationary). This subspace is irrelevant for Change Point Detection but increases the 
overall dimensionality. Moreover, stationary components with high signal power can make change
points invisible to the observer and also to detection algorithms. For example, there are no change 
points visible in the 
time series depicted in the left panel of Figure~\ref{fig:tempseg_intro}, even though there exists 
one direction in the two-dimensional signal space which clearly shows two change-points, as it can 
be seen in the middle panel. However, the non-stationary contribution is not visible in the observed signal
because of its relatively low power (right panel). In this example, we also observe that it does
not suffice to select channels individually, as neither of them appears informative. 
In fact, in many application domains such as biomedical engineering~\cite{ZieheBio, EEGPrin} or geophysical 
data analysis~\cite{Geophysics}, it is most plausible that the data is generated as a mixture of 
underlying sources which we cannot measure directly. 

In this chapter we show how to extract useful features for Change Point Detection
by finding the most non-stationary directions using a variant of Stationary Subspace Analysis \cite{PRL:SSA:2009}. Even 
though there exists a wide range of feature extraction methods for classification and 
regression \cite{Guyon:2003q}, to date, no specialized procedure for feature extraction or for general signal processing \cite{HaykinSig} has been proposed for Change Point Detection. In controlled
simulations on synthetic data, we show that for three representative Change Point Detection algorithms 
the accuracy is significantly increased by a prior feature extraction step, in particular 
if the data is high dimensional. This effect is consistent over various numbers of dimensions and strengths 
of change-points. In an application to Fault Monitoring, where the ground truth is available, 
we show that the proposed feature extraction improves the performance and leads to a 
dimensionality reduction where the desired state changes are clearly visible. Moreover, 
we also show that we can determine the correct dimensionality of the informative subspace. 

The remainder of this chapter is organized is follows. In the next Section~\ref{sec:FeEx}, we introduce 
our feature extraction method that is based on an extension of Stationary Subspace Analysis. 
Section~\ref{sec:Sim} contains the results of our simulations and in Section~\ref{sec:Real}
we present the application to Fault Monitoring. Our conclusions are outlined in Section~\ref{sec:conclusion}. 

\section{Feature Extraction for Change-Point Detection}
\label{sec:FeEx}
Feature extraction from raw high-dimensional data has been shown to be useful not only 
for improving the performance of subsequent learning algorithms on the derived 
features \cite{Guyon:2003q} but also for understanding high-dimensional complex physical systems. In many application areas such as 
Computer Vision~\cite{Foerstner:1994q}, Bioinformatics~\cite{Saeys:2007q, Morris:2005q} and
Text Classification~\cite{Lewis:1991q}, defining useful features is in fact the main 
step towards successful machine learning. General feature extraction methods for classification 
and regression tasks are based on maximizing the mutual information between features 
and target \cite{Torkkolla:2003q}, explaining a given percentage of the variance in 
the dataset \cite{Schoelkopf:1998q}, choosing features which maximize the margin between 
classes \cite{Li:2006q} or selecting informative subsets of variables through enumerative search (wrapper methods) \cite{Guyon:2003q}.
However, for Change-Point Detection no dedicated feature extraction has been proposed 
\cite{ChangePoint}. Unlike in classical supervised feature selection, where a target 
variable allows us to measure the informativeness of a feature, for Change-Point Detection we cannot tell 
whether a feature elicits the changes that we aim to detect since there is usually 
no ground truth available (the problem is unsupervised). Even so, feature extraction is feasible following the 
principle that a useful feature should exhibit significant distributional changes 
over time. 
Reducing the dimensionality in a pre-processing step should be particularly beneficial 
for the Change-Point Detection task: most algorithms either explicitly or implicitly make approximations to  
probability densities \cite{DistBased2, DistBased4} or directly compute a divergence 
measure based on summary statistics, such as the mean and covariance \cite{ChangePoint} 
between segments of the time series --- both are hard problems whose sample complexities 
grow exponentially with the number of dimensions. 

As we have seen in the example presented in Figure~\ref{fig:tempseg_intro}, selecting
channels individually (univariate approach) is not helpful or may lead to suboptimal
features. The data may be non-stationary overall despite the fact that each dimension 
seems stationary. Moreover, a single non-stationary source may be expressed across a large
number of channels. It is therefore more sensible to estimate a linear projection of the 
data which contains as much information relating to change-points as possible. In this chapter, 
we demonstrate that finding the projection to the most non-stationary directions
using a variant of Stationary Subspace Analysis significantly increases the performance of 
change-point detection algorithms. 

In the remainder of this section, we first review the SSA algorithm and show 
how to extend it towards finding the most non-stationary directions.
Then we show that this approach corresponds to finding the projection that is  
most likely to be non-stationary in terms of a statistical hypothesis test. 

\subsection{Stationary Subspace Analysis}

The following section uses material adapted from the supplementary material from the original SSA publication \cite{PRL:SSA:2009}
and, of course, the paper upon which this chapter is based (see \cite{CPpreprint}).

Stationary Subspace Analysis factorizes a multivariate 
time series $x(t) \in \R^D$ into stationary and non-stationary sources according to the linear 
mixing model,  
\begin{equation}
  x(t) = A {\mathbf s}(t) = \begin{bmatrix} A^{\s} & A^{\n} \end{bmatrix}
  \begin{bmatrix} s^{\s}(t) \\  s^{\n}(t) \end{bmatrix},
\label{eq:mixing_model}
\end{equation}
where $s^\s(t)$ are the $d_s$ stationary sources,  
$s^\n(t)$ are the $d_n$ ($d_n+d_s = D$) non-stationary sources, and $A$ is an unknown
time-constant invertible mixing matrix. The spaces spanned by the columns of the mixing 
matrix $A^{\s}$ and $A^{\n}$ are called $\s$- and $\n$-space respectively. Note that in contrast 
to Independent Component Analysis (ICA)~\cite{ICABook}, there is no independence 
assumption on the sources $s(t)$. 

The aim of SSA is to invert the mixing model (Equation~\ref{eq:mixing_model}) given only 
samples from the mixed sources $x(t)$, i.e.~we want to estimate the demixing matrix 
$\hat{B}$ that separates the stationary from the non-stationary sources. 
Applying $\hat{B}$ to the time series $x(t)$ yields the estimated stationary and 
non-stationary sources $\hat{s}^\s(t)$ and $\hat{s}^\n(t)$ respectively,
\begin{align}
\label{eq:applying_solution}
	\begin{bmatrix} \hat{s}^\s(t) \\ \hat{s}^\n(t) \end{bmatrix}
	=
	\hat{B} x(t)
	=
	\begin{bmatrix} \hat{B}^\s \\ \hat{B}^\n \end{bmatrix} x(t)
	=
	\begin{bmatrix} \hat{B}^\s A^\s & \hat{B}^\s A^\n \\ \hat{B}^\n A^\s &  \hat{B}^\n A^\n \end{bmatrix}
	\begin{bmatrix} \hat{s}^\s(t) \\ \hat{s}^\n(t) \end{bmatrix} .
\end{align}
The submatrices $\hat{B}^\s \in \R^{d_s\times D}$ and $\hat{B}^\n \in \R^{(d_n)\times D}$ of the estimated
demixing matrix $\hat{B}$ project to the estimated stationary and non-stationary sources and are called
\s-projection and \n-projection respectively. The estimated mixing matrix $\hat{A}$ is the inverse of
the estimated demixing matrix, $\hat{A} = \hat{B}^{-1}$.

The inverse of the SSA model (Equation~\ref{eq:mixing_model}) is not unique: given one demixing
matrix $\hat{B}$, any linear transformation \textit{within} the two groups of estimated sources leads
to another valid separation, because it leaves the stationary resp.~non-stationary nature of the sources
unchanged. 
In addition, the separation into \s- and \n-sources itself is not unique: adding stationary components to
a non-stationary source leaves the source non-stationary, whereas the converse is not true. That is, the
\n-projection can only be identified up to arbitrary contributions from the stationary sources.
Hence one cannot recover the true \n-sources, but only the true \s-sources (up to linear transformations).
Conversely, we can identify the true \n-space (because the \s-projection is orthogonal to it) but 
not the true \s-space. However, in order to extract features for change-point detection, our aim 
is not to recover the true non-stationary sources, per se (since as we will see below, defining \emph{the} non-stationary sources is problematic), but instead the \textit{most} non-stationary ones. 

An SSA algorithm depends on a definition of stationarity, which the \s-projection aims to satisfy. 
In the SSA algorithms~\cite{PRL:SSA:2009, HarKawWasBun10SSA},  a time series $X_t$ is considered
  stationary if its mean and covariance is constant over time, i.e.:~
\begin{align*}
  \E[X_{t_1}] & = \E[X_{t_2}] \\
  \E[X_{t_1} X_{t_1}^\top] & = \E[X_{t_2} X_{t_2}^\top],
\end{align*}
for all pairs of time points $t_1, t_2 \in \NN_0$. This is a variant of weak stationarity~\cite{Pri83Spectral} whereby
time structure is not taken into account. 
Following this concept of stationarity, the SSA algorithm~\cite{PRL:SSA:2009} 
finds the \s-projection $\hat{B}^\s$ that minimizes the difference between the first two moments
of the estimated \s-sources $\hat{s}^\s(t)$ across epochs of the time series. (Non-overlapping epochs of
the time series are used since one cannot
estimate the mean and covariance at a single time point.) Thus the samples from $x(t)$
are divided into $n$ non-overlapping epochs defined by the index sets $\mathcal{T}_1, \ldots, \mathcal{T}_n \subset \NN_0$
and the epoch mean and covariance matrices are estimated as:
\begin{align*}
\emu_i = \frac{1}{|T|} \sum_{t \in \mathcal{T}_i} x(t) \hspace{0.5cm} \text{ and }  \hspace{0.5cm}
\esi_i = \frac{1}{|T|-1} \sum_{t \in \mathcal{T}_i} \left( x(t)-\emu_i \right)\left( x(t)-\emu_i \right)^\top ,
\end{align*}
respectively for all epochs $1 \leq i \leq n$. Given an \s-projection, the epoch mean and 
covariance matrix of the estimated \s-sources in the $i$-th epoch are:
\begin{align*}
	\emu^\s_i = \hat{B}^\s \emu_i \hspace{0.5cm} \text{ and } \hspace{0.5cm}
	\esi^\s_i = \hat{B}^\s \esi_i  ( \hat{B}^\s )^\top .
\end{align*}
The difference in the mean and covariance matrix between two epochs is measured using 
the Kullback-Leibler divergence between Gaussians. (Due to the fact that we measure only the covariance and mean, the maximum entropy principle tells us that the most prudent model is the Gaussian.) The objective function is the sum of the information theoretic difference 
between each epoch and the average epoch. Since the \s-sources can only be determined up to an 
arbitrary linear transformation and since a global translation of the data does not change the difference
between epoch distributions, 
without loss of generality the data is centered and whitened \footnote{A whitening transformation is a basis 
transformation $W$ that sets the sample covariance matrix to the identity. 
It can be obtained from the sample covariance matrix $\hat{\Sigma}$ as $W = \hat{\Sigma}^{-\frac{1}{2}}$.
}
This implies that one may assume that the average epoch's mean and covariance matrix are:
\begin{align}
\label{eq:avg_epoch}
	\frac{1}{N} \sum_{i=1}^N \emu_i = 0 \hspace{0.5cm} \text{ and } \hspace{0.5cm} \frac{1}{N} \sum_{i=1}^N \esi_i = I .
\end{align}
Moreover, the search for the true \s-projection may be restricted to the set of
matrices with orthonormal rows, i.e.~$\hat{B}^\s (\hat{B}^\s)^\top = I$. Thus 
the optimization problem becomes:
\begin{align}
	\hat{B}^s & =
	\argmin_{B B^\top = I} \; \sum_{i=1}^N \KLD \Big[ \Gauss(\emu^\s_i,\esi^\s_i) \; \Big|\Big| \; \Gauss(0,I) \Big] \notag \\
	& = \argmin_{B B^\top = I} \; \sum_{i=1}^N \left(
			- \log\det\esi^\s_i
			+ (\hat{\mu}^\s_i)^\top \emu^\s_i 	\right) ,
\label{eq:ssa_objfun}
\end{align}
which can be solved efficiently by using multiplicative updates with orthogonal matrices
parameterized as matrix exponentials of antisymmetric matrices \cite{PRL:SSA:2009,Plu05} \footnote{An efficient implementation of SSA may be downloaded free of charge at \url{http://www.stationary-subspace-analysis.org/toolbox}}.

\subsection{Finding the Most Non-Stationary Sources} 

In order to extract useful features for change-point detection, we would like to find 
the projection to the most non-stationary sources. However, the SSA algorithms
\cite{PRL:SSA:2009,HarKawWasBun10SSA} merely estimate the projection to the
most stationary sources and choose the projection to the non-stationary sources to be
orthogonal to the found \s-projection, which means that all stationary contributions are 
projected out from the estimated \n-sources. In the case in which the covariance between the sources orthogonal to 
the \s-sources is constant over time, this implies that no information relating to non-stationarity 
is thereby lost by following this protocol to obtain sources containing non-stationarity. This constancy, however, may not always 
hold: non-stationarity may well reside in changing covariance
between \s- and \n-sources.
So, intuitively, we see that in order to find directions which do not lose information
relating to the non-stationarity contained in the data, we need to
propose a different method than simply taking the orthogonal complement of the stationary 
projection. 
The problem then remains whether we may frame a sensible way of defining \emph{the}
non-stationary sources, independently of the definition of \emph{the} stationary sources, since any non-stationary source remains non-stationary
when a stationary source is imposed onto that source: it is also not at all clear
if there is a sensible way to define non-stationarity of a data set which non-circularly guarantees
that the superposition of stationary noise onto a non-stationary direction yields
a less non-stationary direction and thus allows us to recover the orthogonal case mentioned above; neither is it clear that this should be the case from an information theoretic point of view\footnote{The project of describing non-stationarity canonically in terms of information theory, is, however, problematic. See Chapter~\ref{ChapSup} for details.}. Therefore, we simply take, as our definition 
of the non-stationary sources, those sources which maximize the 
SSA loss function. This definition makes, trivially, the non-stationary sources
unique; thus, as our working definition for non-stationarity, we are taking the measure which is
optimized for SSA (see Equation~\ref{eq:ssa_objfun}). There are various independent justifications for this measure, which 
we will not explore too deeply here: these include the fact that the minimum of the SSA loss 
function is a consistent estimator for the stationary projection, that the loss is information theoretic in the sense that it
is independent of parameterizations of the probability distribution and underlying vector space and 
yields positive results for the task at hand (\emph{a posteriori} justification). Thus the rationale 
behind the following approach is:

\begin{enumerate}
\item We \emph{define} a non-stationarity measure using the SSA loss function due to 
\\ its intuitive plausibility.
\item We thus define \emph{the} non-stationary sources as those maximizing 
the \\ non-stationarity.
\end{enumerate}

Thus, taking the SSA loss as our definition of non-stationarity, we aim to find the most non-stationary sources: this means optimizing 
the \n-projection instead of the \s-projection. Before we turn to the optimization problem, let us first of all analyze 
the situation more formally in order to develop some intuition for the difference between maximizing
non-stationarity and taking the orthogonal complement of the stationary projection.
 
\begin{figure}[ht]
 \begin{center}
  \includegraphics{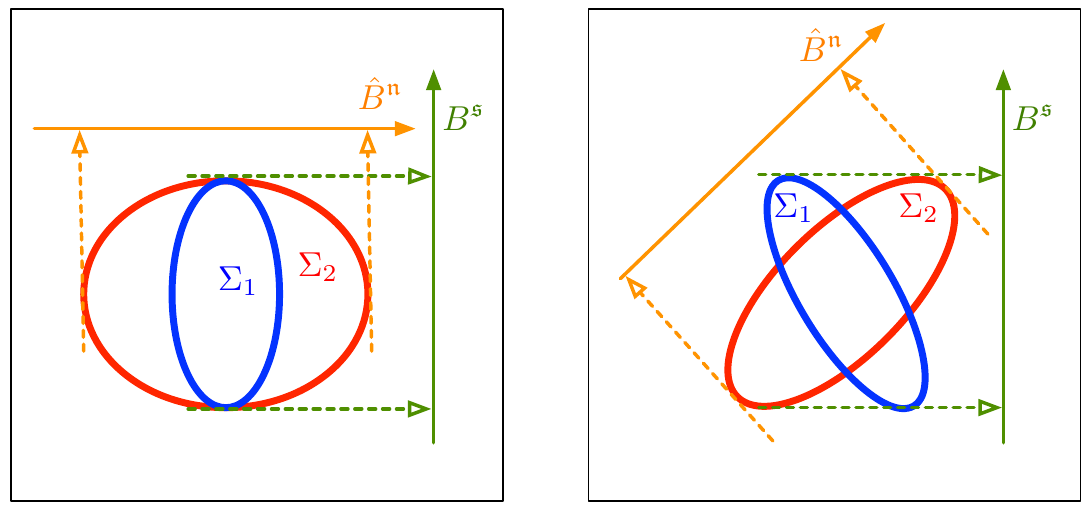}
  \caption{
  	The left panel shows two epoch covariance matrices $\Sigma_1$ and $\Sigma_2$ where the
	non-stationarity is confined to changes in the variance along one direction, hence the
	most non-stationary projection $\hat{B}^\n$ is orthogonal to the true stationary 
	projection $B^\s$. This is not the case in the situation depicted in the right panel: 
	here, the covariance of the two dimensions changes between $\Sigma_1$ and $\Sigma_2$, 
	so that we can find a non-stationary projection that is more non-stationary than 
	the orthogonal complement of the true \s-projection. 
    \label{fig:nstat_optim}
    }
 \end{center}
\end{figure}

We consider first a simple example comprising one stationary and one non-stationary source with 
corresponding normalized basis vectors $\| A^\s \| = 1$ and $\| A^\n \| = 1$ respectively, 
and we let $\phi$ be the angle between the two spaces, i.e.~$\cos \phi = A^{\s \top} A^\n$. We will 
consider an arbitrary pair of epochs, $\mathcal{T}_1$ and $\mathcal{T}_2$, and find 
the projection $\hat{B}^\n$ which maximizes the difference in mean $\Delta_\mu$ and variance $\Delta_\sigma$ 
between $\mathcal{T}_1$ and $\mathcal{T}_2$.

Let $X_1$ and $X_2$ be bivariate random variables modeling the distribution of 
the data in the two epochs respectively. According to the linear mixing model (Equation~\ref{eq:mixing_model}), 
we can write $X_1$ and $X_2$ in terms of the underlying sources:
\begin{align*}
	X_1 & = A^\s X_s + A^\n X_{n_1} \\
	X_2 & = A^\s X_s + A^\n X_{n_2} 
\end{align*}
where the univariate random variable $X_s$ represents the stationary source and the two 
univariate random variables $X_{n_1}$ and $X_{n_2}$ model the non-stationary sources, 
in the epochs $\mathcal{T}_1$ and $\mathcal{T}_2$ respectively. Without loss of generality, we will 
assume that the true \s-projection $B^\s = (A^\n)^\perp$ is normalized, $\| B^\s \| = 1$. In order
to determine the relationship between the true \s-projection and the most non-stationary
projection, we write $\hat{B}^\n$ in terms of $B^\s$ and $A^\n$:
\begin{align}
\label{eq:defnproj}
	\hat{B}^\n = \alpha B^\s + \beta A^{\n \top}, 
\end{align}
with coefficients $\alpha, \beta \in \R$ such that $\| \hat{B}^\n \| = 1$.
In the next step, we will observe which \n-projection maximizes the difference in 
mean $\Delta_\mu$ and covariance $\Delta_\sigma$ between the two epochs  $\mathcal{T}_1$ and $\mathcal{T}_2$. 
Let us first consider the difference in the mean of 
the estimated \n-sources:
\begin{align*}
	\Delta_\mu = \E[ \hat{B}^\n X_1 ] - \E[ \hat{B}^\n X_2 ] =  \hat{B}^\n A^\n ( \E[X_{n_1}] -   \E[X_{n_2}] ) .
\end{align*}
This is maximal for $\hat{B}^\n A^\n = 1$, i.e.~when $\hat{B}^\n$ is orthogonal to $B^\s$.  
Thus, with respect to the difference in the mean, choosing the \n-projection $\hat{B}^\n$ to be orthogonal to the \s-projection is 
always optimal, irrespective of the type of distribution change between epochs.

Let us now consider the difference in variance $\Delta_\sigma$ of the estimated \n-sources
between epochs. This is given by: 
\begin{multline*}
	\Delta_\sigma = \Var[ \hat{B}^\n X_1 ] - \Var[ \hat{B}^\n X_2 ] =  \beta^2 ( \Var [ X_{\n_1} ] - \Var [ X_{\n_2} ] )
												\\ + 2  \left[ \alpha \cos \left(\phi + \frac{\pi}{2}\right) + \beta \cos \phi \right] \underbrace{( \Cov[ X_\s,  X_{\n_1} ] - \Cov[ X_\s,  X_{\n_2} ] )}_{= \Delta_{\sigma_{\s \n}}} .
\end{multline*}
Clearly, when there is no change in the covariance of the \s- and the \n-sources between the two epochs, 
i.e.~$\Delta_{\sigma_{\s \n}} = 0$, the difference $\Delta_\sigma$ is maximized for
 $\hat{B}^\n = (B^\s)^\perp$. See the left panel of Figure~\ref{fig:nstat_optim} for an example.
However, when the covariance between \s- and \n-sources does 
vary, i.e.~$| \Delta_{\sigma_{\s \n}}|>0$, the projection $(B^\s)^\perp$ is no longer 
the most non-stationary. To see this, consider the derivative of $\Delta_\sigma$ 
with respect to the $\alpha$ at $\alpha = 0$:
\begin{align*}
	\partial \Delta_\sigma / \partial \alpha |_{\alpha=0} = 2  \cos \left(\phi + \frac{\pi}{2}\right) \Delta_{\sigma_{\s \n}} . 
\end{align*}
Since this derivate does not vanish, $\alpha = 0$ (see Equation~\ref{eq:defnproj}) is not an extremum when 
$|\Delta_{\s \n}|>0$, which means that the most non-stationary
projection is not orthogonal to the true \s-projection. This is the case in the right panel of
Figure~\ref{fig:nstat_optim}. 

Thus here we have seen an example where clearly maximizing non-stationarity is equivalent to maximizing variance
difference and we have seen that this is not equivalent to taking the orthogonal complement of the stationary projection.
Of course, maximizing variance differences is, in general, not an appropriate definition for non-stationarity, which 
is where the SSA loss comes into play as a more general measure\footnote{There are other, arguably canonical,
characterizations of non-stationarity, for example, as the empirical entropy over the parameter space of the non-stationary process;
this is a topic of current research.}.

Thus, in order to find the projection to the most non-stationary sources, we need to
maximize the non-stationarity of the estimated \n-sources. To that end, we propose maximizing the 
SSA objective function (Equation~\ref{eq:ssa_objfun}) for the \n-projection:
\begin{align}
	\hat{B}^\n = \argmax_{B B^\top = I} \; \sum_{i=1}^N \left(
			- \log\det\esi^\n_i
			+ (\hat{\mu}^\n_i)^\top \emu^\n_i 	\right) ,
\label{eq:ssa_nobjfun}
\end{align}
where $\esi_i^\n = \hat{B}^\n \esi_i (\hat{B}^\n)^\top$ and $\hat{\mu}^\n_i =  \hat{B}^\n \emu_i$ for 
all epochs $1 \leq i \leq N$. 

\subsection{Relationship to Statistical Testing}

In this section we show that maximizing the SSA objective function to find the most 
non-stationary sources can be understood from a statistical testing point-of-view, 
in that it also maximizes a test statistic which minimizes the $p$-value for rejecting the null hypothesis that the estimated directions are stationary. 
In doing so, we provide an alternative rationale for the loss function we maximize. In addition, the interpretation of the loss
function in terms of testing will allow us to detect the number of \emph{actually} non-stationary directions in 
dataset.

More precisely, we maximize a test statistic which thereby minimizes a $p$-value for a statistical hypothesis test that compares 
two models for the data: the null hypothesis $H_0$ that each epoch follows a standard normal 
distribution vs.~the alternative hypothesis $H_A$ that each epoch is Gaussian distributed with 
individual mean and covariance matrix. Let $X_1, \ldots, X_N$ be random variables modeling 
the distribution of the data in the $N$ epochs. Formally, the hypothesis can 
be written as follows:
\begin{align*}
	& H_0 : X_1, \ldots, X_N \sim \mathcal{N}(0, I) \\
	& H_A : X_1 \sim \mathcal{N}(\mu_1, \Sigma_1), \ldots, X_N \sim \mathcal{N}(\mu_N, \Sigma_N)
\end{align*}
In other words, the statistical test tells us whether we 
should reject the simple model $H_0$ in favor of the more complex model $H_A$. This decision 
is based on the value of the test statistics, whose distribution is known under the null hypothesis 
$H_0$. Since $H_0$ is a special case of $H_A$ and since the parameter estimates are obtained by 
Maximum Likelihood, we can use the likelihood ratio test statistic $\Lambda$ \cite{Likelihood},
which is the ratio of the likelihood of the data under $H_0$ and $H_A$, where the parameters are
their maximum likelihood estimates. 

Let $\mathcal{X} \subset \R^{d_n}$ be the data set which is divided into 
$N$ epochs $\mathcal{T}_1, \ldots, \mathcal{T}_n$ and let $\emu^\n_1, \ldots, \emu^\n_N$ 
and $\esi^\n_1, \ldots, \esi^\n_N$ be the maximum likelihood estimates of the mean and covariance 
matrices of the estimated \n-sources respectively. Let $p_{\mathcal{N}}(x ; \mu, \Sigma)$ be the probability density 
function of the multivariate Gaussian distribution. The likelihood ratio test statistic is given by:
\begin{align}
  \Lambda(\mathcal{X}) = - 2 \log \frac{ \prod_{x \in \mathcal{X}} p_{\mathcal{N}}(x ; 0, I)  } 
  { 
  	\prod_{i=1}^N \prod_{x \in \mathcal{T}_i} p_{\mathcal{N}}(x ; \emu^\n_i , \esi^\n_i) 
  }
\end{align}
which is approximately $\chi^2$ distributed with $\frac{1}{2} N d_n (d_n+3)$ degrees of freedom \cite{Likelihood}. 
Using the facts that we have set the average epoch's mean and covariance matrix to zero and 
the identity matrix respectively, i.e.:
\begin{align}
	\frac{1}{N} \sum_{i=1}^N \emu^\n_1  = 0 \hspace{0.5cm} \text{ and } \hspace{0.5cm} \frac{1}{N} \sum_{i=1}^N \esi^\n_i = I, 
\end{align}

Letting, in addition,
$M$= \emph{no. of data points in the entire dataset}, gives the following difference of logarithms\footnote{Throughout we require the following identity: $z^T A z = tr(zz^T A)$.}. 
\begin{equation}
 \Lambda(\mathcal{X})  = \text{log}(\prod_{i=1}^N \prod_{j \in T_i} (2\pi)^{-d/2} |\covar|^{-\frac{1}{2}}e^{-\frac{1}{2}(x_j-\mmu)^T \covar^{-1} (x_j-\mmu)}) - \text{log}(\prod_{j=1}^M (2\pi)^{-d/2} e^{-\frac{1}{2}x_j^T x_j})
\end{equation}

The simplicity of the right hand term is because we are testing the hypothesis of whether the data is generated from a normal distribution with constant covariance and mean 
which we can assume w.l.o.g. to be resp. white and 0.

This gives us:
\begin{equation}
\sum_{i=1}^N \sum_{j \in T_i}( \text{log}(2\pi^{-d/2}) -\frac{1}{2}\text{log}(|\covar|)-\frac{1}{2}(x_j-\mmu)^T \covar^{-1} (x_j-\mmu)) -( \sum_{j=1}^M( \text{log}(2\pi^{-d/2}) -\frac{1}{2}x_j^T x_j))
\end{equation}

The $\pi$ terms cancel and we can multiply out the rest to get:

 \begin{equation}
  \sum_{i=1}^N -\frac{N_i}{2} \text{log}(|\covar|)-\frac{1}{2}\sum_{i=1}^N \sum_{j \in T_i}(x_j^T\covar^{-1}x_j-\mmu^T \covar^{-1}x_j  -x_j^T\covar^{-1}\mmu +\mmu^T \covar^{-1}\mmu) + \frac{1}{2}\sum_{j=1}^Mx_j^T x_j
  \label{Complicated}
\end{equation}

Take the second term without the factor of $-1/2$ in front and neglecting the outer sum: the inner sum distributes over the multiplication with the inverses to give:
\begin{equation}
(\sum_{j \in T_i}x_j^T\covar^{-1}x_j)-N_i(\mmu^T \covar^{-1}\mmu + \mmu^T\covar^{-1}\mmu - \mmu^T \covar^{-1}\mmu)
\end{equation} 
Which is:
\begin{equation}
(\sum_{j \in T_i}x_j^T\covar^{-1}x_j)-N_i\mmu^T \covar^{-1}\mmu
\end{equation} 
Which is:
\begin{equation}
(\sum_{j \in T_i}\text{tr}(x_jx_j^T\covar^{-1})-N_i\text{tr}(\mmu\mmu^T \covar^{-1})
\end{equation} 
Which gives:
\begin{equation}
\text{tr}((\sum_{j \in T_i}x_jx_j^T)-N_i\mmu\mmu^T\covar^{-1})=\text{tr}(N_i \covar \covar^{-1})=\text{tr}(N_i I) = dN_i
\end{equation} 

Now take the final term from equation \ref{Complicated} with the factor in front:
\begin{equation}
\sum_{j=1}^Mx_j^T x_j = \sum_{i=1}^N \sum_{j\in T_j} x_j^T x_j = \sum_{i=1}^N \sum_{j\in T_j} (x_j^T x_j - \mmu^T\mmu + \mmu^T\mmu)
\end{equation}
Then we use the same trick as before but using the identity $I$ as $A$ to get:
\begin{equation}
\sum_{j=1}^Mx_j^T x_j = \sum_{i=1}^N( N_i \text{tr}(\covar) + N_i \mmu^T\mmu))
\end{equation}

So the term in equation \ref{Complicated} becomes:
\begin{equation}
 \sum_{i=1}^N -\frac{N_i}{2} \text{log}(|\covar|)-\frac{1}{2}\sum_{i=1}^N dN_i  + \frac{1}{2} \sum_{i=1}^N( N_i \text{tr}(\covar) + N_i \mmu^T\mmu))
\end{equation}

Which in all cases simplifies to:
\begin{equation}
 \sum_{i=1}^N -\frac{N_i}{2} \text{log}(|\covar|)-\frac{d}{2}M  + \frac{1}{2} \sum_{i=1}^N( N_i \text{tr}(\covar) + N_i \mmu^T\mmu))
\end{equation}

So the test statistic simplifies to:
\begin{align}
\label{eq:teststat}
	\Lambda(\mathcal{X}) = -d_s N  + \sum_{i=1}^N N_i \left( - \log \det \esi^\n_i  + \| \emu^\n_i \|^2 + \text{tr}(\esi^{\n}_i) \right),  
\end{align}
where $N_i = | \mathcal{T}_i |$ is the number of data points in the $i$-th epoch. If 
every epoch contains the same number of data points ($N_1 = \cdots = N_N$), then 
maximizing the SSA objective function (Equation~\ref{eq:ssa_nobjfun}) is equivalent 
to maximizing the test statistic (Equation~\ref{eq:teststat}) and hence minimizing the $p$-value 
for rejecting the simple (stationary) model for the data. 

As we will see in the application to Fault Monitoring (Section~\ref{sec:Real}), the $p$-value of this test furnishes a useful indicator for the number of informative
directions for Change Point Detection. More specifically to obtain an upper bound on the optimal number of directions for Change Point Detection, we use SSA to find the stationary sources, increasing the number of stationary sources until the test returns that the projection is significantly non-stationary.
These sources may safely be removed without loss of informativeness for Change Point Detection. Removing more directions may sacrifice information for Change Point Detection;
on the other hand, depending on the particular data set, removing additional directions may lead to increases in performance as a result of the reduced dimensionality.
Notice here, that the procedure for obtaining the upper bound uses the standard SSA algorithm, whereas in the final preprocessing step for Change Point Detection we optimize for
non-stationarity.

\begin{figure}[ht]
 \begin{center}
  \includegraphics[width = 90mm]{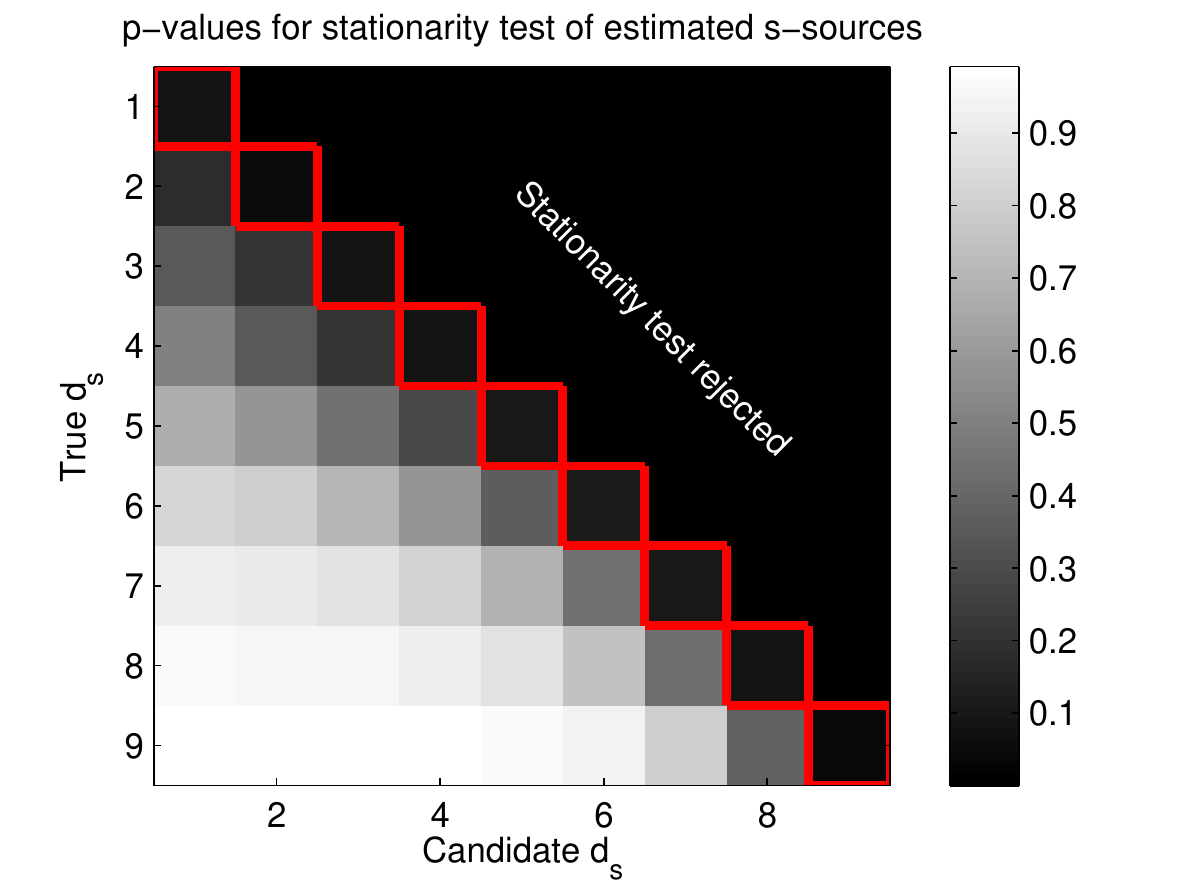}
  \caption{Average p-values obtained over 100 realizations of the dataset for each setting of the actual $d_s$. The number of dimensions is $D=10$ and the $y$-axis displays the actual number for $d_s$.
  The $x-$axis displays the value for the parameter $d_s$ used to compute the stationary projection using SSA. The red box shows the decision made at the $p=0.01$ confidence level. The red box displays
  the choice made using this decision rule for choosing the parameter $d_s$ which occurs most often. The results show that the method picks the correct parameter on averages, over simulations and thus 
  represents a feasible procedure for choosing the number of non-stationary sources, $d_n$.
     \label{fig:p_values}
    }
 \end{center}
\end{figure}

To demonstrate the feasibility of using the test statistic to select the parameter $d_s$ we display, in Figure~\ref{fig:p_values}, $p$-values obtained using SSA for a fixed value for simulated data's dimensionality $D=10$, the number of stationary sources ranging from $d_1 = 1, \dots, 9$ and the chosen parameter ranging from $1, \dots, 9$.  The confidence level $p = 0.01$ for rejection of the null hypothesis $H_0 =$ \emph{The projected data is stationary} returns the correct $d_s$, on average, in all cases. For each simulation, a dataset was synthesized (according to the details described in Section~\ref{sec:syndata}) of length 20,000 with 200 epochs. Then for each possible parameter setting for $d_s$, a stationary projection $\hat{P}^{\s}$ was computed. Finally the value of the test statistic together with the $p-$value were computed on the estimated stationary sources. The dataset is described in Section~\ref{sec:Sim}.

\section{Simulations}
\label{sec:Sim}

In this section we demonstrate the ability of SSA to enhance the segmentation performance of three change-point 
detection algorithms on a synthetic data setup. The algorithms are single linkage clustering with divergence (SLCD) \cite{Gower69SingleLinkage}  which uses the mean and 
covariance as test statistics, CUSUM \cite{Page:1954fk}, which uses a sequence of hypothesis tests and the Kohlmorgen/Lemm \cite{DistBased2}, using a kernel density measure and a hidden Markov model. For each segmentation algorithm we compare the performance of 
the baseline case in which the dataset is segmented without preprocessing, the case in which the data is 
preprocessed by projecting to a random subspace and the case in which the dataset is preprocessed 
using SSA. We compare performance with respect to the following schemes of parameter variation:

\begin{enumerate}
\item The dimensionality $D$ of the time series is fixed and $d_n$, the number of non-stationary sources is varied.
\item The number $d_n$ of non-stationary sources is fixed and $d_s$, the number of the stationary sources is varied.
\item $D$, $d_n$ and $d_s$ are fixed and the power $q$ between the changes in the non-stationary sources is varied.
\end{enumerate}

For two of the Change-Point Detection algorithms which we test, SLCD and Kohlmorgen/Lemm, all three parameter variation schemes are 
tested. For CUSUM the second scheme does not apply as the method is a univariate method.

For each setup and for each realization of the dataset we perform segmentation on the raw dataset, 
the estimated non-stationary sources after SSA preprocessing for that dataset and on a $d_n$ 
dimensional random projection of the dataset. The random projection acts as a comparison measure for 
the accuracy of the SSA-estimated non-stationary sources for segmentation purposes. 

\begin{table}
\begin{center} 
\begin{tabular}{| c | c | c | c | c || c | c | c | } 
\hline 
Setup & $D$ & $d_n$ & $d_s$ & $q$ & SLCD & Kohl./Lemm & CUSUM \\
\hline \hline
(1) & \tickNo & \tickYes & \tickYes & \tickNo & Fig. \ref{fig:LinkageROC}, Pa. 1 & Fig. \ref{fig:KL_ROC}, Pa. 1 & Fig. \ref{fig:CUSUM_ROC}, Pa. 1 \\
(2) & \tickYes & \tickNo & \tickYes & \tickNo & Fig. \ref{fig:LinkageROC}, Pa. 2 & Fig. \ref{fig:KL_ROC}, Pa. 2 & Fig. \ref{fig:CUSUM_ROC}, Pa. 2 \\
(3) & \tickNo & \tickNo & \tickNo & \tickYes & Fig. \ref{fig:LinkageROC}, Pa. 3 & Fig. \ref{fig:KL_ROC}, Pa. 3 & Fig. \ref{fig:CUSUM_ROC}, Pa. 3 \\
\hline
\end{tabular} 
\caption{Overview of simulations performed and corresponding figures reporting the results. A tick denotes that the corresponding parameter was varied in the experiment. A cross denotes that the corresponding parameter is kept fixed. ("pa." denotes panel within the respective figures.) }
\end{center}
\end{table}

\subsection{Synthetic Data Generation}
\label{sec:syndata}
The synthetic data which we use to evaluate the performance of Change Point Detection methods is generated as a linear mixture of stationary and non-stationary sources.
The data is further generated epoch-wise: each epoch has fixed length and each dataset consists of a concatenation of epochs.
The $d$ stationary sources are distributed Normally on each epoch according to $\Gauss(0,I_{d_s})$. The other $d_n$ (non-stationary) source signals $s^\n(t)$ are distributed according to the active model $k$ of this epoch; this active model is one of five Gaussian distributions $\G_k = \Gauss(0,\Sigma_k)$: the covariance $\Sigma_k$ is a diagonal matrix whose eigenvalues are chosen at random from five log-spaced values between $\sigma_1^2 = 1/q$ and $\sigma_5^2 = q$; thus five covariances, corresponding to the $\G_k$ of the Markov chain are then chosen in this way. The transition between models over consecutive epochs follows a Markov model with transition probabilities:
\begin{equation}
	P_{ij} = 
	\begin{cases} 
		0.9   & i=j \\ 
		0.025 & i\neq j .
	\end{cases}	
\end{equation} 
Here, the indices $i$ and $j$ correspond to the $ith$ and $jth$ epoch, respectively of the dataset, so that $P_ij$ describes the probability of transition between the $ith$ and $jth$ epochs. 

In our experiments, we vary the parameters $D$, the total number of sources, $d_n$ (or equivalently, $d_s$), the number of non-stationary sources and $q$, the power change in the non-stationary sources.

\subsection{Performance Measure}
\label{sec:perform}
In our experiments we evaluate the algorithms based on an estimation of the area under the ROC curves (AUC) across realizations of the dataset. The true positive rate (TPR) and false positive rate (FPR) are defined with respect to the fixed epochs which constitute the synthetic dataset; a change-point may only occur between two such epochs of fixed length. Each of the change-point algorithms, which we test, reports changes with respect to the same division into epochs as per the synthetic dataset: thus the TPR and FPR are well defined.

We use the AUC because it provides information relating to a range of TPR and FPR.  In signal detection the trade off achieved between TPR and FPR depends on operational constraints: cancer diagnosis procedures must achieve a high TPR perhaps at the cost of a higher than desirable FPR. Network intrusion detection, for instance, may need to compromise the TPR given the computational demands set by too high an FPR. In order to assess detection performance across all such requirements the AUC provides the most informative measure: one integrates over all possible trade offs.
More specifically, each algorithm is accompanied by a parameter $\tau$ which controls the trade off between TPR and FPR. For SLCD this is the number of clusters, for CUSUM this is the threshold set on the log likelihood ratio and for the Kohlmorgen/Lemm this is the parameter controlling how readily a new state is assigned to the model.  In each case we vary $\tau$ to obtain AUCs.

\subsection{Single Linkage Clustering with Symmetrized Divergence Measure (SLCD)}

Single Linkage Clustering with a Symmetrized Distance Measure is a simple algorithm for Change Point Detection which has, however, the advantage of efficiency and of segmentation based on a parameter independent distance matrix (thus detection may be repeated for differing trade offs between TPR and FPR without reevaluating the distance measure).
In particular, segmentation based on Single Linkage Clustering \cite{Gower69SingleLinkage} computes a distance measure based on the covariance and mean over time windows to estimate the occurrence of change-points:
the algorithm consists of the following three steps:
\begin{enumerate}
	\item The time series is divided into 200 epochs for 
				which we estimate the epoch-mean and epoch-covariance matrices 
				$\{ ( \hat{\boldsymbol \mu}_i, \hat{\Sigma}_i ) \}_{i=1}^{200}$.

	\item The dissimilarity matrix $D \in \R^{200 \times 200}$ between the epochs is computed 
				as the symmetrized Kullback-Leibler divergence $\KLD$ between the estimated distributions
				(up to the first two moments),			  
				\begin{align*}
					D_{ij} = 	\frac{1}{2} \KLD\left[ \Gauss(\hat{\boldsymbol \mu}_i, \hat{\Sigma}_i ) \; || \; \Gauss(\hat{\boldsymbol \mu}_j, \hat{\Sigma}_j )  \right] + 
	\frac{1}{2} \KLD\left[ \Gauss(\hat{\boldsymbol \mu}_j, \hat{\Sigma}_j ) \; || \; \Gauss(\hat{\boldsymbol \mu}_i, \hat{\Sigma}_i )  \right],  
				\end{align*}
				where $\Gauss({\boldsymbol \mu}, \Sigma)$ is the Gaussian distribution.

	\item Based on the dissimilarity matrix $D$, Single Linkage Clustering \cite{Gower69SingleLinkage} 
				(with number of clusters set to $k=5$) returns an assignment of epochs to clusters such that 
				a change-point occurs when two neighbouring epochs do not belong to the same cluster.

\end{enumerate}

\begin{figure}[ht]
 \begin{center}
  \includegraphics[width = 120mm]{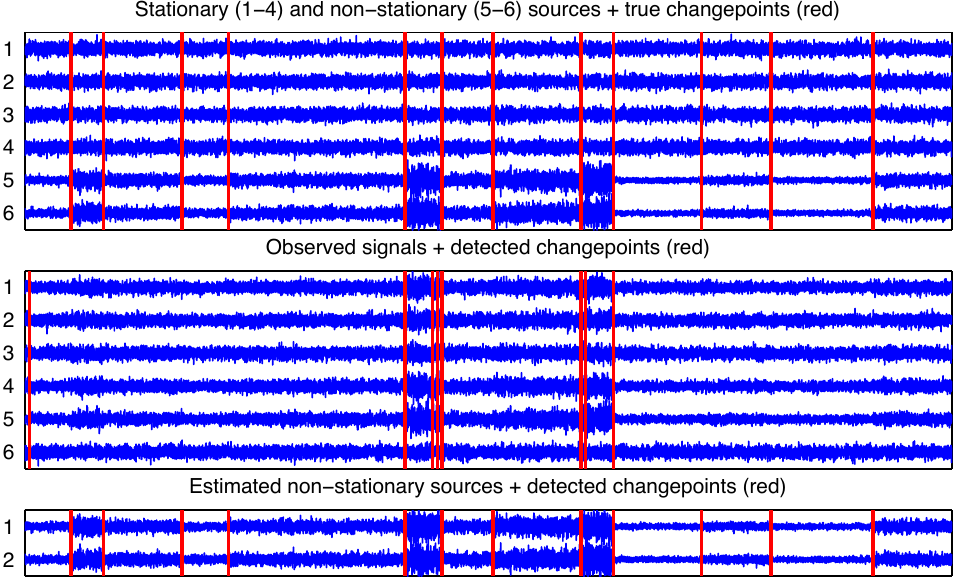}
  \caption{An Illustration of a case in which SSA significantly improves single linkage clustering with divergence: $d_s = 4$ (no. of stat. sources), $d_n = 2$ (no. non-stat.), $q = 2.3$ (power change in non-stat. sources). 
  The top panel displays the true decomposition into stationary and non-stationary sources with the true change-points marked. The middle panel displays the
  change-points which SLCD finds on the entire data set (sources mixed): clearly some change-points are left undetected. The bottom panel displays the change-points found by SLCD on the estimated non-stationary
  sources.
    \label{fig:seg_illustration}
    }
 \end{center}
\end{figure}




\subsubsection{Results}

\begin{figure}[ht]
 \begin{center}
  \includegraphics[width = 130mm]{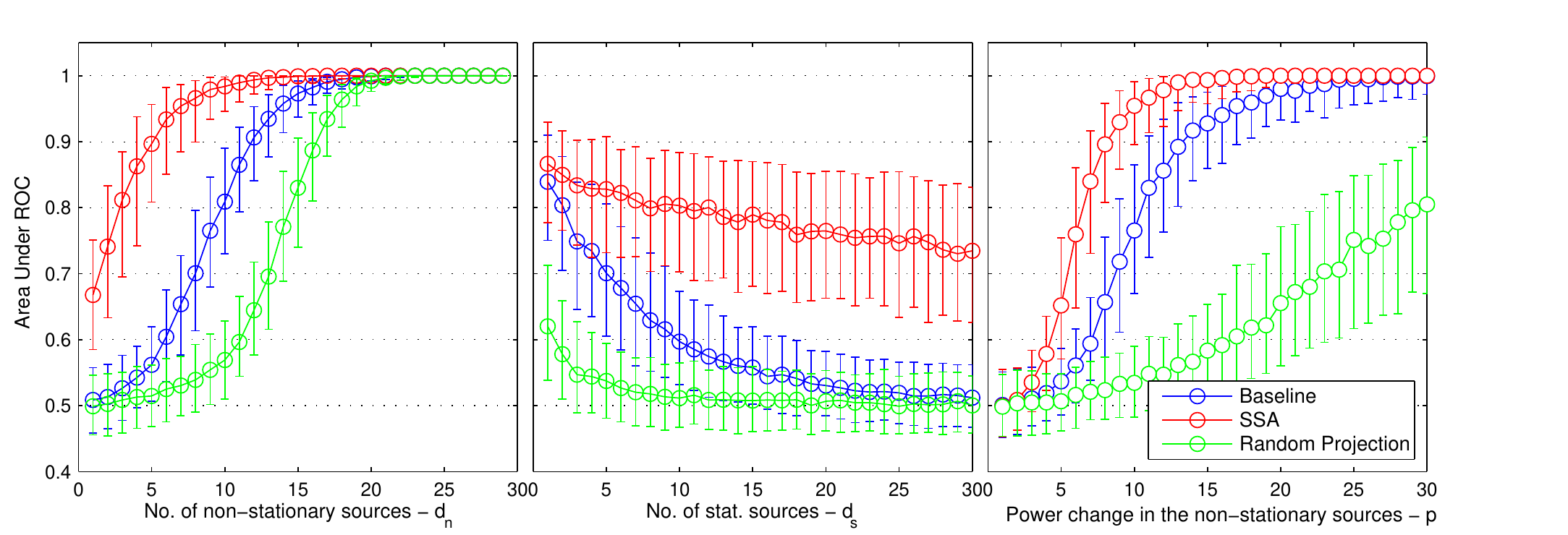}
  \caption{Results of the simulations for Single Linkage Clustering with Symmetrized Divergence (SLCD). The left panel displays the results for a fixed dimensionality of the time series, $D = 30$ and varying $d_n$, the number of stationary sources with $q=1.8$.The middle panel displays the results for a fixed number of non-stationary sources, $d_n = 2$ and varying  $d_s$, the number of stationary sources and with the power change $q=1.8$. The right panel displays the results for fixed $D = 20$, $d_s = 16$ and $d_n = 4$ and for varying $q$, the power change in the non-stationary sources. Each displays the results in terms of the area under the ROC curve computed as per Section \ref{sec:perform}.
  The error bars extend from the 25th to the 75th percentiles.
    \label{fig:LinkageROC}}
 \end{center}
\end{figure}

The results of the simulations for varying numbers of non-stationary sources in  a dataset of 30 channels are shown in Figure~\ref{fig:LinkageROC} in the first panel. When the degree to which the changes are visible is lower (i.e. there are fewer non-stationary directions in the data setup), SSA preprocessing significantly outperforms the baseline method, even for a small number of irrelevant stationary sources. 

The results of the simulations for a varying number of stationary dimensions with 2 non-stationary dimensions are displayed in Figure~\ref{fig:LinkageROC} in the second panel. For small $d_s$ the performance of the baseline and SSA preprocessing are similar: SSA's performance is more robust with respect to the addition of higher numbers of stationary sources, i.e. noise directions. The segmentations produced using SSA preprocessing continue to carry information relating to change-points for $d_s = 30$, whereas, for $d_s \geq 12$, the baseline's AUC approaches $0.5$, which corresponds to the accuracy of randomly chosen segmentations.

The results of the simulations for varying power $q$ in the non-stationary sources with $D = 20$, $d_s = 16$ (no. of stat. sources) and $d_n = 4$ are displayed in Figure~\ref{fig:LinkageROC} in the third panel.
Both the performance of the baseline and of the SSA preprocessing improves with increasing power change $q$. This effect is evident for lower $q$ for the SSA preprocessing
than for the baseline.

An illustration of a case in which SSA preprocessing significantly outperforms the baseline is displayed in Figure~\ref{fig:seg_illustration}. The estimated non-stationary sources exhibit a far clearer illustration of the change-points than the full dataset: the corresponding segmentation performances reflect this fact.

\subsection{Weighted CUSUM for changes in variance}

In statistical quality control,  CUSUM (or cumulative sum control chart) is a sequential analysis technique developed in 1954 \cite{Page:1954fk}. CUSUM is one of the most widely used and oldest methods for Change Point Detection; the algorithm is an online method for Change Point Detection based on a series of log-likelihood ratio tests.
Thus CUSUM algorithm detects a change in parameter $\theta$ of a process $p_\theta(y)$ \cite{Page:1954fk} and is asymtotically optimal when the pre-change and post-change parameters are known \cite{ChangePoint}.
For the case in which the target value of the changing parameter is unknown, the \emph{weighted} CUSUM algorithm is defined as a direct extension of CUSUM \cite{ChangePoint}, by integrating over a parameter interval. The following statistics ${\tilde \Lambda}_j^k$ constitutes likelihood ratios between the currently estimated parameter of the non-stationary process and differing target values (values to which the parameter may change), integrated over a measure $F$:
 
 \begin{equation}
{\tilde \Lambda}_j^k  = \Bigg{(} \int _{-\infty}^{\infty}\frac{p_{\theta_1}(y_j,...,y_k)}{p_{\theta_0}(y_j,...,y_k)}dF(\theta_1) \Bigg{)}
 \end{equation}  
 
 Here $y_j,...,y_k$ denote the timepoints lying inside a sliding window of length $k$ whereby $y_k$ indicates the latest time point received.
 The stopping time is then given as follows:
 
 \begin{equation}
 t_a = \mathrm{min}\{ k : \mathrm{max}\{ j \leq k : \mathrm{ln}({\tilde \Lambda}_j^k ) \geq h\} \}
 \end{equation}
 
The function $F$ serves as a weighting function for possible target values of the changed parameter. In principle the algorithm can thus be applied to multi-dimensional data. However, as per \cite{ChangePoint}, the extension of the CUSUM algorithm to higher dimensions is non-trivial, not just because integrating over possible values of the covariance is computationally expensive but also because various parameterizations can lead to the same likelihood function. Given this, we test the effectiveness of the algorithm in computing one-dimensional segmentations. In particular we compare the segmentation performed on the one dimensional projection chosen by SSA with the best segmentation of all individual dimensions with respect to hit-rate on each trial. 
In accordance with \cite{ChangePoint} we choose $F$ to comprise a fixed uniform interval containing all possible values of the process's variance. We approximate the integral above as a sum over evenly spaced values on that interval. We approximate the stopping time by setting:
\begin{equation}
t_a \approx \mathrm{min}\{ k : \mathrm{ln}({\tilde \Lambda}_{k-W+1}^k ) \geq h\}
\end{equation}
The exact details of our implementation are as follows. Let $T$ be the number of data points in the data set $X$.

 \begin{enumerate}
	\item We set the window size $W$, the sensitivity constant $h$ and the current time step as $t_c=W+1$ and $\theta_0 = \text{var}(\{x_1,...,x_W\})$ and $\Theta = \{ \theta_1,\ldots,\theta_r \} = \{ c, c + b, c + 2b,\ldots, d\}$. 		
	\item ${\tilde \Lambda}_j^k = \frac{1}{b} \sum_{i = 1}^{r}\frac{p_{\theta_i}(y_j,\ldots,y_k)}{p_{\theta_0}(y_j,\ldots,y_k)}$
	\item If $\mathrm{ln}({\tilde \Lambda}_j^k) \geq h$ then a change-point is reported at time $t_c$ and $t_c$ is updated so that $t_c = t_c + W$ and $\theta_0 = \text{var}(\{x_{t_c - W +1},\ldots,x_{t_c}\})$. We return to step 2.
	\item Otherwise if ${\tilde \Lambda}_j^k < h$ no change-point is reported and $t_c = t_c + 1$. We return to step 2.
\end{enumerate}
 
 \subsubsection{Results}
  
\begin{figure}[ht]
 \begin{center}
  \includegraphics[width = 120mm]{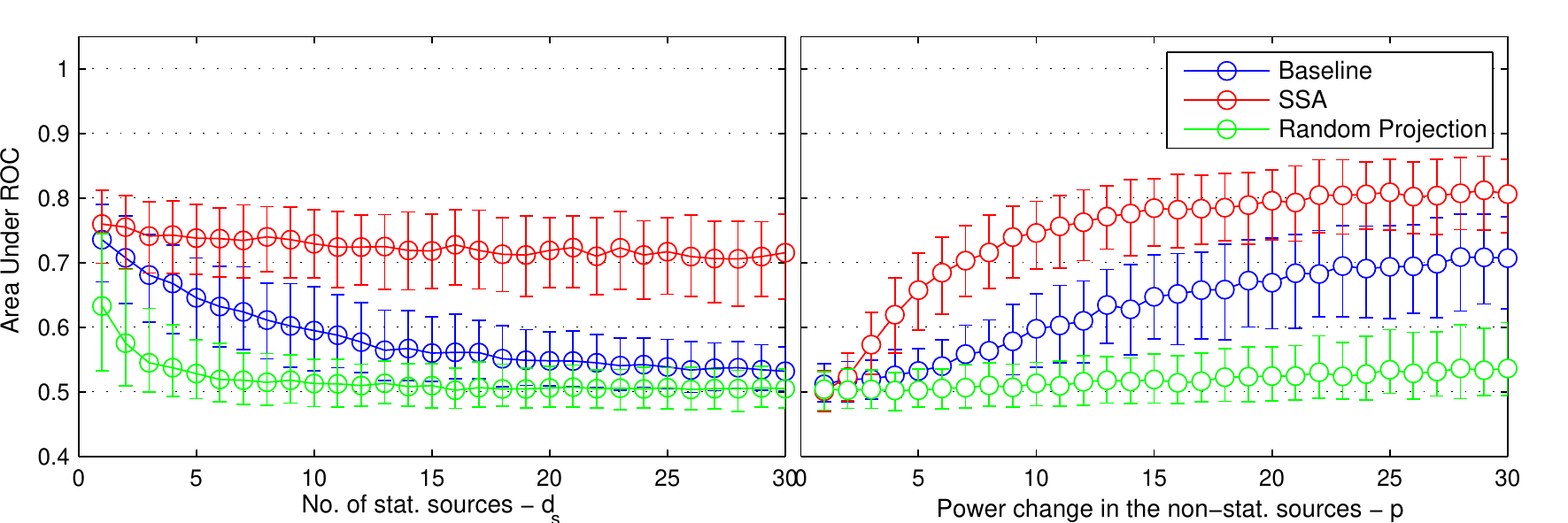}
  \caption{Results of the simulations for CUSUM. The left panel displays the results for a fixed number of non-stationary sources, $d_n = 1$, fixed power change $q=1.8$ and varying  $d_s$, the number of stationary sources. The right panel displays the results for fixed $D = 16$ and $d = 15$ and for varying $q$, the power change in the non-stationary sources. Each displays the results in terms of the area under the ROC curve computed as per Section \ref{sec:perform}.
  The error bars extend from the 25th to the 75th percentiles.
    \label{fig:CUSUM_ROC}
    }
 \end{center}
\end{figure}

 In Figure~\ref{fig:CUSUM_ROC}, in the left panel, the results for varying numbers of stationary sources are displayed. Weighted CUSUM with SSA preprocessing significantly outperforms the baseline for all values of D (dimensionality of the time series). Here we set $d_n=1$, the number of non-stationary sources, for all values of $d_s$, the number of stationary sources.
 
 In Figure~\ref{fig:CUSUM_ROC}, in the right panel, the results for changes in the power change between ergodic sections $q$ are displayed for $D = 16$, $d_s=15$ and $d_n = 1$. SSA outperforms the baseline for all except very low values of $q$, the power level change, where all detection schemes fail. The simulations show that SSA represents a method for choosing a one dimensional subspace to render uni-dimensional segmentation methods applicable to higher dimensional datasets: the resulting segmentation method on the one dimensional derived non-stationary source will be simpler to parametrize and more efficient. If the true dimensionality of the non-stationary part is $d_n=1$ then no information loss should be observed.

%
 
\subsection{Kohlmorgen/Lemm Algorithm}

The Kohlmorgen/Lemm algorithm is a flexible non-parametric and multivariate method which may be applied in online and offline operation modes.
Distinctive about the Kohlmorgen/Lemm algorithm is that a kernel density estimator, rather than a simple summary statistic, is used to estimate the occurence of change-points. In particular the algorithm is based on a standard Kernel Density Estimator with Gaussian kernels and estimation of the optimal segmentation based on a Hidden Markov Model  \cite{DistBased2}. More specifically if we estimate the densities on two arbitrary epochs $E_i,E_j$ of our dataset $X$ with Gaussian kernels then we can define a distance measure
$d$ between epochs via the $L2$-Norm yielding:

\begin{eqnarray}
d(E_i,E_j) &=& \frac{1}{W^2 (4\pi \sigma^2)^{d/2}} \sum_{w,v = 0}^{W-1} \Bigg( \mathrm{exp}\left( - \frac{(Y_w - Y_v)^2}{4\sigma^2})\right)  \\  
 &-&  2\mathrm{exp}\left( - \frac{(Y_w - Z_v)^2}{4\sigma^2})\right) + \mathrm{exp}\left( - \frac{(Z_w - Z_v)^2}{4\sigma^2})\right)\Bigg)
\end{eqnarray}

Here, $Y_w$ corresponds to the $w$th point of epoch $E_i$ and $Z_v$ to the $v$th point of epoch $E_j$. The formula for the distance measure $d(E_i,E_j)$ is derived analytically based on the
expression for the kernels used for density estimation and simplifies the computations over calculating the densities explicitly (see the corresponding paper \cite{DistBased2} for details).
The final segmentation is then based on the distance matrix generated between epochs calculated with respect to the above distance measure $d$.
As per the weighted CUSUM, it is possible to define algorithms whose sensitivity to distributional changes in reporting change-points is related to the value of a parameter $C$: $C$ controls the probability of transitions to new states in the fitting of the hidden markov model. However, in \cite{Kohlmorgen:2003fk} it is shown that in the case when all change-points are known then one can also derive an algorithm which returns exactly that number of change-points: in simulations we evaluate the performance on the first variant over a full range of parameters to obtain an ROC curve. In addition we choose the parameter $\sigma$ according to the rule of thumb given in \cite{DistBased2}, which sets $\sigma$ proportional to the mean distance of each data point to its $D$ nearest neighbours, where $D$ is the dimensionality of the data: this is evaluated on a sample set. The exact implementation we test is based on the papers \cite{Kohlmorgen:2003fk} and \cite{DistBased2}.  The details are as follows:

\begin{enumerate}
\item The time series is divided into epochs 
\item A distance matrix is computed between epochs using kernel density estimation and the $L2$-norm as described above.
\item The estimated density on each epoch corresponds to a state of the Markov Model. So a state sequence is a sequence of estimated densities.
\item Finally, based on the estimated states and distance matrix, a hidden Markov model is fitted to the data and a change-point reported whenever consecutive epochs have been fitted with differing states.
\end{enumerate}

\subsubsection{Results}

\begin{figure}[ht]
 \begin{center}
  \includegraphics[width = 130mm]{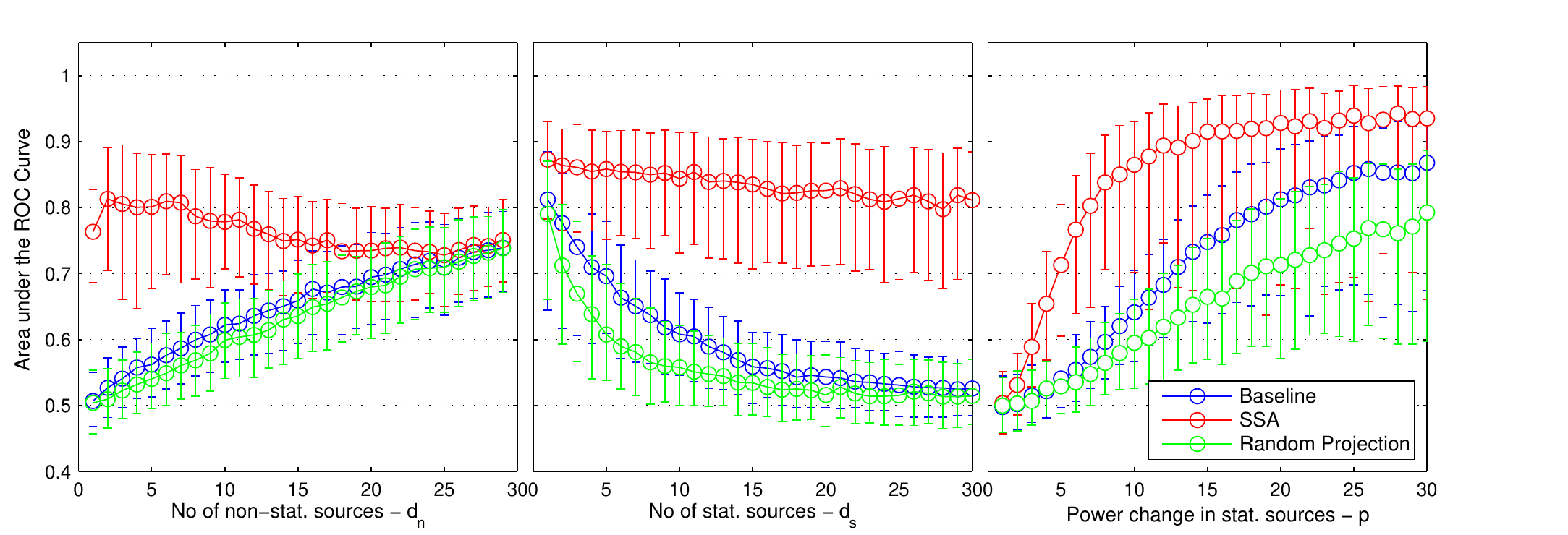}
  \caption{Results of the simulations for the Kohlmorgen/Lemm.The left panel displays the results for a fixed dimensionality of the time series, $D = 30$, varying $d_n$, the number of stationary sources and fixed power change $q=1.8$.The middle panel displays the results for a fixed number of non-stationary sources, $d_n = 2$, varying  $d_s$, the number of stationary sources and fixed power change $p = 1.8$. The right panel displays the results for fixed $D = 20$, $d_s = 16$, $d_n=4$ and for varying $q$, the power change in the non-stationary sources. Each displays the results in terms of the area under the ROC curve computed as per Section \ref{sec:perform}. The error bars extend from the 25th to the 75th percentiles.
    \label{fig:KL_ROC}
    }
 \end{center}
\end{figure}

SSA preprocessing improves the segmentation obtained using the Kohlmorgen/Lemm algorithm for all three schemes of parameter variation of the dataset.
In particular:
the area under the ROC (AUC) for varying $d_s$ and fixed $D$ are displayed in Figure~\ref{fig:KL_ROC}, in the first panel, with $D=30$.
The area under the ROC (AUC) for varying $d_s$ and fixed $d_n$ are displayed in Figure~\ref{fig:KL_ROC}, in the second panel, with $d_n=2$.
The area under the ROC (AUC) for varying power change in the non-stationary sources $p$ and fixed $D$ and $d_s$ are displayed in Figure~\ref{fig:KL_ROC}, in the third panel, with p ranging between 1.1 and 4.0 at increments of 0.1.
Of additional interest is that for varying $d_n$ and fixed $D$ the performance of segmentation with SSA  preprocessing is superior for higher values of $d_s$: this implies that the improvement of Change Point Detection of the Kohlmorgen/Lemm algorithm due to the reduction in dimensionality to the informative estimated n-sources outweighs the difficulty of the problem of estimating the n-sources in the presence of a large number of noise dimensions.

%
%

\section{Application to Fault Monitoring}
\label{sec:Real}

In this section we apply our feature extraction technique to Fault Monitoring. The dataset consists of multichannel 
measurements of machine vibration. The machine under investigation is a pump, driven by an electromotor. The incoming shaft is reduced in speed by 
two delaying gear-combinations (a gear-combination is a combination of 
driving and a driven gear). Measurements are repeated for two identical machines, where the first shows a progressed pitting in both gears, and the second machine is virtually fault free. The rotating speed of the driving shaft is measured with a tachometer\footnote{The dataset can be downloaded free of charge at \url{http://www.ph.tn.tudelft.nl/~ypma/mechanical.html}.}.

\begin{figure}[!h]
\centering
\includegraphics[width = 122mm]{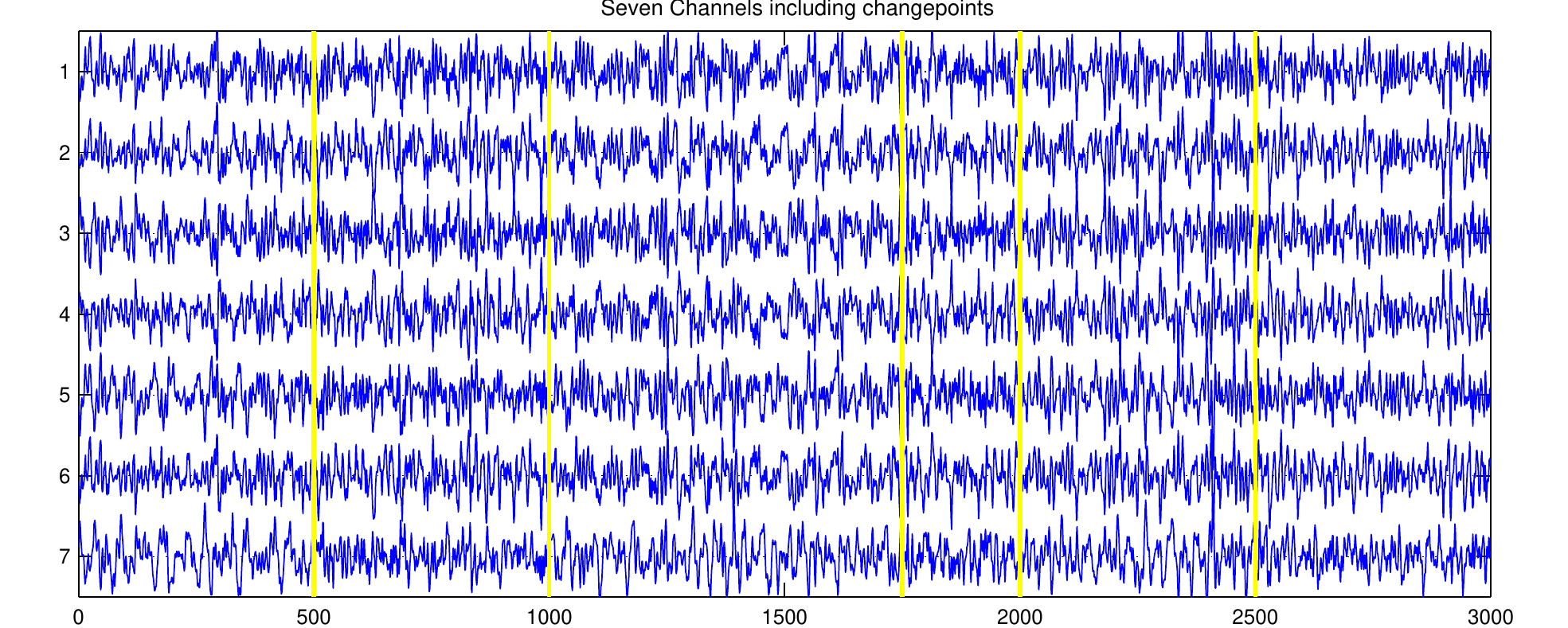}
\caption{Pump Dataset: the machine under investigation is a pump, driven by an electromotor. The measurements made are of machine vibration at seven sensors. The data alternates between two conditions: normal functionality and pitting in both gears. \label{fig:CLOSERLOOKDATA}}
 \end{figure}

The pump data set is semi-synthetic insofar as we juxtapose non-temporally consecutive sections of data between the two pump conditions. Sections of data from the first and second machine are spliced randomly (with respect to the time axis) together to yield a dataset with 10,000 time points in seven channels.
An illustration of the dataset is displayed in Figure~\ref{fig:CLOSERLOOKDATA}.
                                                                                
\subsection{Setup}

We preprocessed with SSA using a division of the dataset into 30 equally sized epochs and for $d_s$, the number of stationary sources ranging between $1$ and $6$, where $D=7$ is the dimensionality of the dataset: 
subsequently we ran the Kohlmorgen/Lemm algorithm on both the preprocessed and raw data using a window size of $W=50$ and a separation of 50 datapoints between non-overlapping epochs.


%
 
\subsection{Parameter Choice}
To select the parameter, $d_s$ (and thus $d_n = D-d_s$), the number of stationary sources, we use the following scheme: the measure of stationarity over which we optimize for SSA is given by the loss function in equation (2).
For each $d_s = \text{dim}(V^s)$ we compute the estimated projection to the stationary sources using SSA on the first half of the data available and computed this loss function on the estimated stationary sources on the second half and compared the result to the values of the loss function obtained on the dataset obtained by randomly permuting the time axis. This random permutation should produce, on average, a set of approximately stationary sources regardless of non-stationarity present in the estimated stationary sources for that $d_s$. In addition a measure of the information relating to non-stationarity lost in choosing the number of stationary sources to be $d_s$, we define the Baseline-Normalized Integral Stationary Error (BNISE) as follows:
\begin{equation}
\text{BNISE}(d) :=_{def} \sum_{d'<d} \frac{L_{d'}(\hat{A}^{-1},X) - \mathbf{E}_{X'}(L_{d'}(\hat{A}^{-1}),X')}{\sigma_{X'}(L_{d'}(\hat{A}^{-1},X'))}
\end{equation}
Here, $L_{d'}(\hat{A}^{-1},X)$ denotes the loss function given in equation~\ref{eq:ssa_objfun} on the original dataset with stationary parameter $r$ and  $L_{r}(\hat{A}^{-1},X')$ the same measure on a random permutation $X'$ of the same dataset. That is using the notation of Equation~\ref{eq:ssa_objfun}, here on the right hand side, $L_{d'}(\hat{A}^{-1},X) = L(\hat{A}^{-1})$ evaluated when specifying
the free parameter to SSA $d_s=d$. 

The motivation of the BNISE is that due to random fluctuations, corresponding to sample sizes, even a stationary dataset should be measured as slightly non-stationary by the SSA loss function. Thus
we measure the deviation in loss between the dataset at hand and the expected loss of a stationary dataset estimated by using the same dataset shuffled on the time axis. This difference is then normalized by the standard deviation over losses estimated under such shuffles.

\subsection{Results}

\begin{figure}[!h]
 \centering
 \includegraphics[width = 110mm]{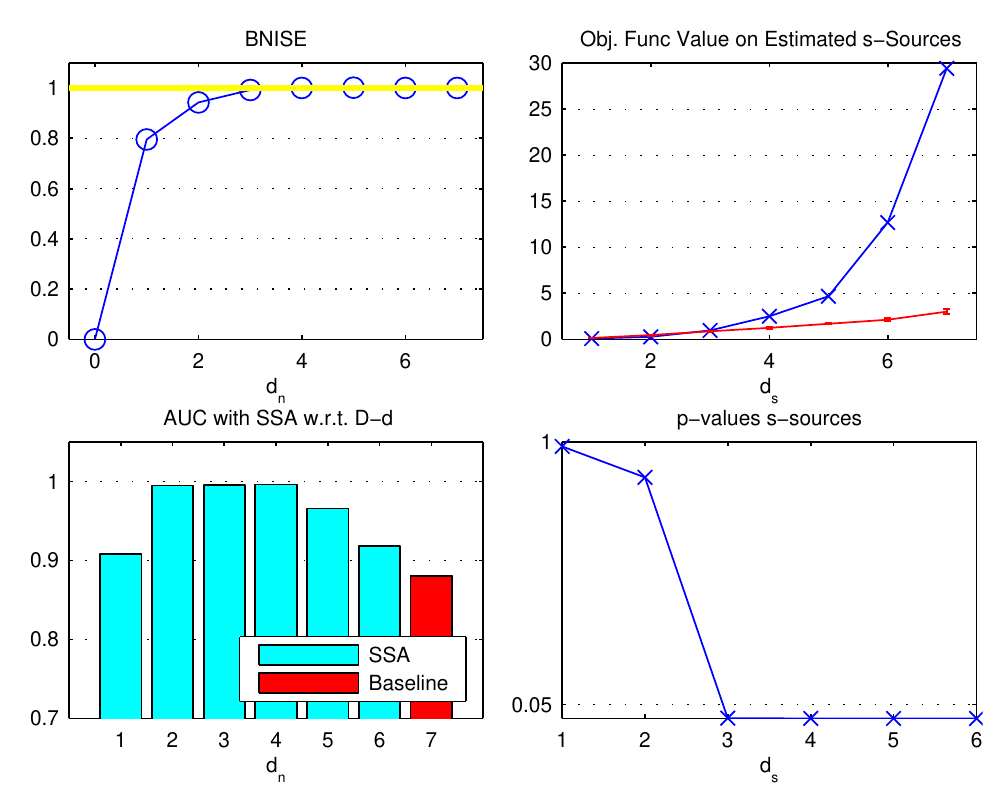}
 \caption{Pump Dataset: schemes for selecting the parameter $d_s$. Top left: the measure BNISE for increasing values of $d_n$. Top right the value of
 the error function as compared to randomly generated data. Bottom left: the AUC performances for various values of $d_n$. Bottom left: $p$-values
 on the estimated $s$-sources.\label{fig:ParameterPlots}}
 \end{figure}

\begin{figure}[!h]
 \centering
 \includegraphics[width = 40mm]{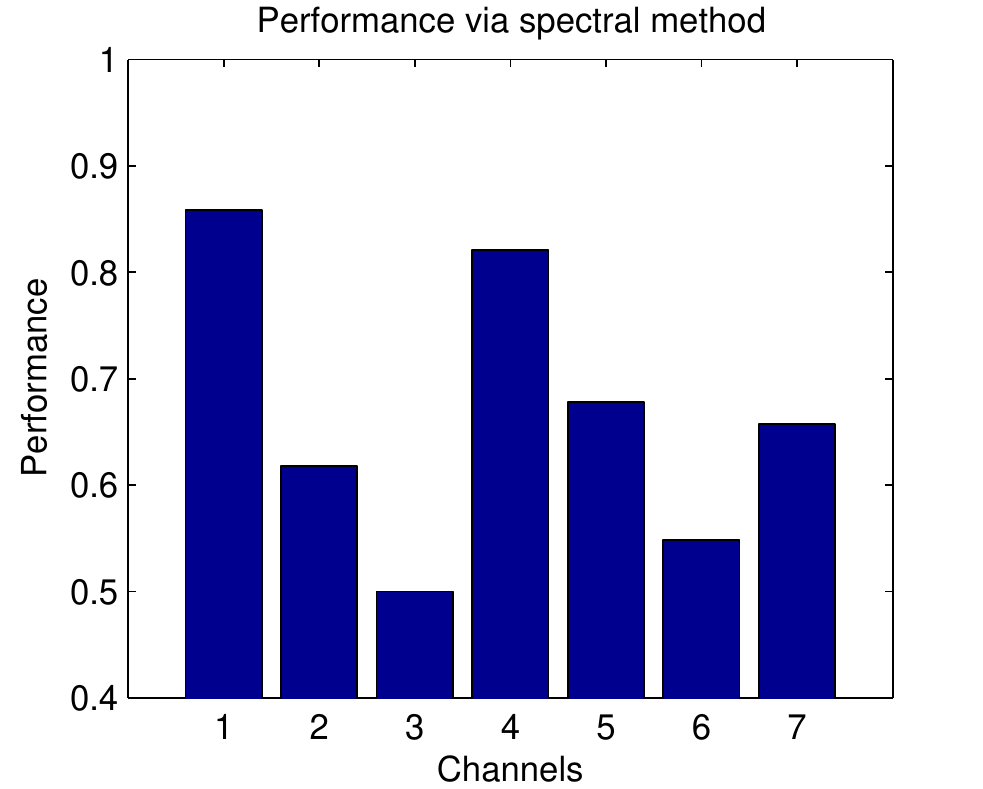}
 \caption{Pump Dataset: performance on spectograms computed on individual channels of the datatset. Each spectrogram is computed with a window length of 50 datapoints and overlap of 49 datapoints.
 7 frequency band windows are used to compute a timeseries of size $7 \times 10,000$.
  \label{fig:freq_pump}
}
 \end{figure}

The results of this scheme and the segmentation are given in Figure~\ref{fig:ParameterPlots}. For $d_s = 6$ we observe a clearly visible difference between the expected loss function value due to small sample sizes and the loss function value present in the estimated stationary sources. Similarly, looking at the p-values, we observe that for $d_s=1,2$ we do not reject the hypothesis that the estimated $\s$-$sources$ are stationary, whereas for higher values of $d_s$ we reject this hypothesis. This implies that $d_n \geq 5$. To test the effectiveness of this scheme, segmentation is evaluated for SSA preprocessing at all possible values of $d_s$.
The AUC values obtained using the parameter choices $d_s = 1,\ldots,6$ for SSA preprocessing as compared to the baseline case are displayed in Figure~\ref{fig:ParameterPlots}. An increase in performance with SSA preprocessing is robust, as measured by the AUC values, with respect to varying choices for the parameter $d_s$ as long as $d_s$ is not chosen $\leq 2$.
Note that, although, for the dataset at hand, there exists information relating to change-points in the frequency spectrum taken over time, this information cannot be used to bring the baseline method onto
par with preprocessing with SSA. We display the results in Figure~\ref{fig:freq_pump} for comparison. Here, segmentation based on a 7-dimensional spectrogram evaluated on each individual channel of the dataset is computed. The best performance over channels for segmentation on each of these spectrograms is lower than the worst performance achieved on the entire dataset without using spectral information, with or without SSA.

\begin{figure}[!h]
 \centering
 \includegraphics[width = 130mm]{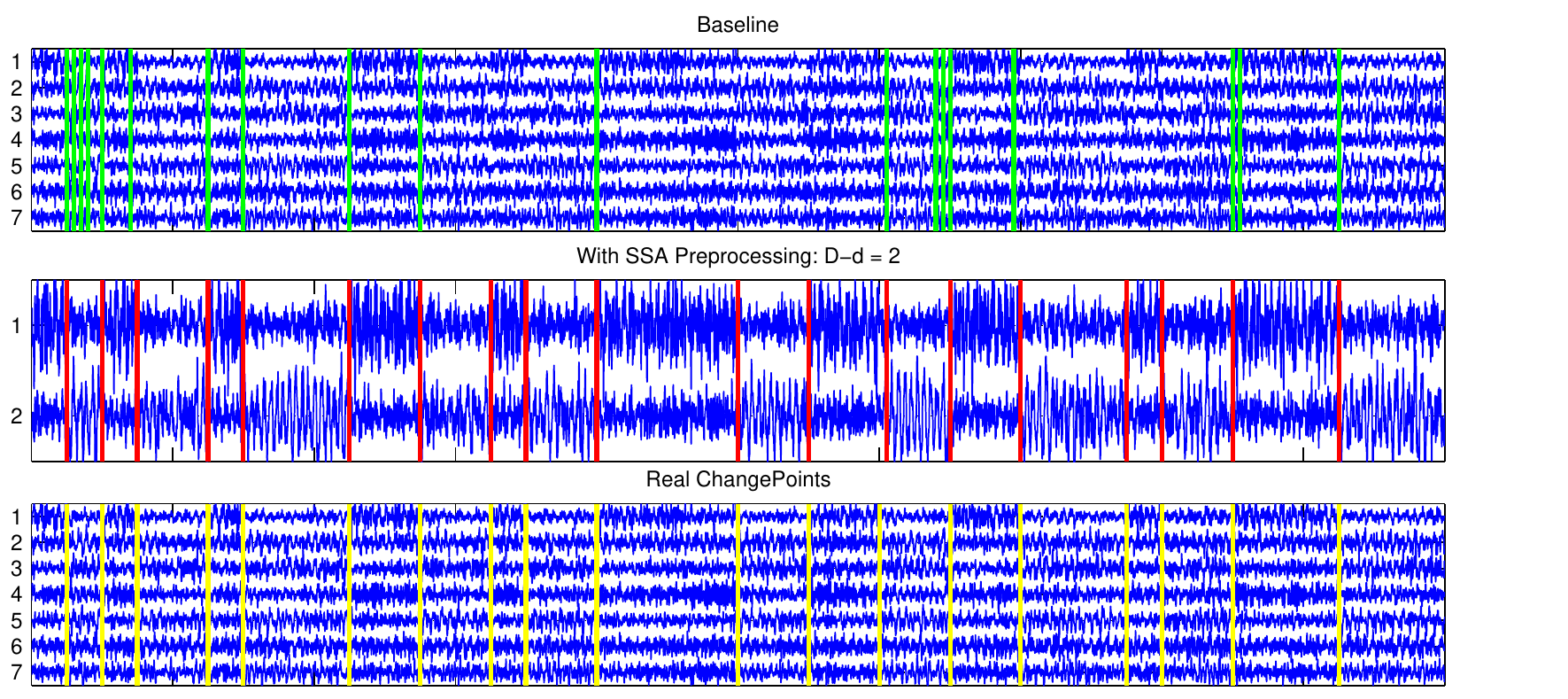}
 \caption{Pump Dataset: all segmentations are computed using Kohlmorgen/ Lemm with the number of change-points $N$ specified \label{fig:PumpPicture}. The baseline corresponds to segmentation
 without SSA preprocessing. The middle panel displays segmentation with SSA preprocessing. The bottom panel displays the real change-points superimposed over the raw dataset.
}
 \end{figure}

\section{Conclusion}
\label{sec:conclusion}

Unsupervised segmentation and identification of time series is a hard problem even in the univariate case and has received considerable attention in science and industry due to its broad applicability ranging from process control and finance to biomedical data analysis.
In high dimensional segmentation problems, different subsystems of the multivariate
time series may exhibit clearer and more informative signals for segmentation than others. We have shown that it may be beneficial to decompose the overall system into stationary and 
non-stationary parts by means of SSA and use the non-stationary subsystem to determine the segmentation. 

This generic approach yields excellent results in simulations and we expect that the proposed dimensionality reduction will 
be useful on a wide range of datasets, because the task of discarding irrelevant stationary information is 
independent of the dataset-specific distribution within the informative non-stationary subspace. Moreover, 
using SSA for preprocessing is a highly versatile because it can be combined with any subsequent segmentation 
method. 

Applications made along the same lines as in the present thesis are effective only when the non-stationary part of the data is visible in the mean and covariances.
The method we have considered may be thus made applicable to general datasets whose changes consist in the spectrum or temporal domain of the data by computing the score function as a further preprocessing step \cite{ChangePoint}. 

\chapter{Classification under Non-Stationarity}

\label{ChapSup}

Classification is a widely studied task in Machine Learning \cite{bb57382,LeCun90GG8}. Less well studied, however, is classification under the assumption that training and test sets differ in distribution. 
The situation under which training and test sets differ arises naturally when both are drawn from a non-stationary time series. In particular, classification under non-stationarity
poses a serious challenge for, for instance, the design of Brain Computer Interfaces \cite{oai:biomedcentral.com:1471-2202-10-S1-P85}. In a wider setting, classification under the assumption of difference of distribution between 
test and training domains but without the assumption that both are drawn from a single time series has been studied, for example, by \cite{oai:repository.ust.hk:1783.1/6830}. Moreover, online learning has been studied in the contribution of Murata et al.~\cite{MurataOnline}. In the following, however, we assume that the training and test distributions are drawn from a single time series. In particular, we focus on the two way classification problem under non-stationarity whereby the class distributions on each epoch may be modeled as Gaussians with differing means. No algorithm has been proposed in the non-stationarity literature which aims to robustify classification against non-stationarity for this setting.
This is exactly the aim of the present chapter.

Some progress has been made, however, in the Brain Computer Interfacing literature in two way classification under non-stationarity. In Brain Computer Interfacing (BCI), non-stationarity may be imposed by artifacts and learning related adaptation \cite{oai:biomedcentral.com:1471-2202-10-S1-P85}. With a view to improving classification perfomance for BCI, the contributions: \cite{oai:biomedcentral.com:1471-2202-10-S1-P85, 5946469,journals/neco/VidaurreSMB11} present
classification based, variance based and adaptation based methods for classification under non-stationarity. The method presented in \cite{5946469} adapts the standard prior feature extraction step for BCI, namely
CSP \cite{SpatFilt}, to yield sCSP, i.e.~\emph{stationary} CSP. CSP functions on the assumption that the variance of two class distributions differ: a projection of the data is then sought which maximizes the difference
between the variance of the two classes. A linear discriminant may subsequently be trained on the data consisting of variances within single trials \cite{SpatFilt}. The data thus extracted as variances on CSP features should be suitable for classification, under the stationarity assumption and the assumption that the classes differ in covariance. sCSP adapts this methodology by extracting features which
are discriminative in variance and simultaneously stationary. This approach has been shown to alleviate, to some extent, the problem of non-stationarity in BCI.

A drawback of the sCSP method is that it is specifically tailored to extracting discriminative features when the class distributions differ in covariance: thus sCSP is highly relevant to classification 
in Brain Computer Interfacing but does not necessarily extend to the general classification setting. In particular, sCSP is not applicable when the class distributions differ on each epoch but not
necessarily in covariance. We aim to address this problem by designing a corresponding method below.
Suppose, therefore, instead, that we only assume that we have set features which are separable but non-stationary; suppose, further, that the classes are discriminable with a given probability without further processing and may be modeled by their mean and covariance and the covariance on both classes is identical: that is to say that Linear Discriminant Analysis (LDA) \cite{LDA} is the correct ansatz. We will propose a method which improves classification under non-stationary for the LDA setting: we assume that the homoscedastic Gaussian assumption is valid within a transient time window but that the data may be non-stationary over a longer time frame; the algorithm will then find a classification direction which is obtained via optimization of a loss function whose minimization rewards separability and penalizes non-stationarity. We will test the resulting algorithm in Section~\ref{sec:Simulations} and Section~\ref{sec:experiments} on BCI data, since BCI represents
a case-study where non-stationarity poses a serious problem: subsequently, the results will be rigorously analyzed and tested. Furthermore, we briefly consider the theoretical background to classification under non-stationarity in Section~\ref{sec:theory}.

\section{SSA for classification problems}

The question now arises, before attempting to design new algorithms: may we use SSA in order to extract discriminative features which are stationary? The answer is: not in general. Firstly, because SSA does not use explicit class information. One may argue that we may choose epochs intelligently and thus use SSA to boost classification: equal numbers of samples from each class are used in each epoch \cite{PRL:SSA:2009}. Thus, in this fashion, SSA may be used to look for stationary directions. However this approach fails whenever non-stationarity is present which leaves the mixture distribution (with a balanced mixture) over classes stationary.

One may avoid, in one respect, this deficiency brought on by an unmodified application of SSA by treating each class separately but deriving a loss function which stipulates that the distribution of each class,
but not the joint distribution over classes, should be stationary at a zero of the loss function \cite{GroupWiseSSA}; the resulting algorithm has been aptly named Group-Wise SSA. This approach, however, has the drawback that, although differences between classes are not treated as non-stationarities, there is no guarantee that the class information is preserved in the features which are thus derived. A second drawback of this Group-Wise SSA is that there may be non-stationarities projected out of the data which are not detrimental to classification.

\begin{figure}[ht]
\begin{center}
\includegraphics[width=140mm]{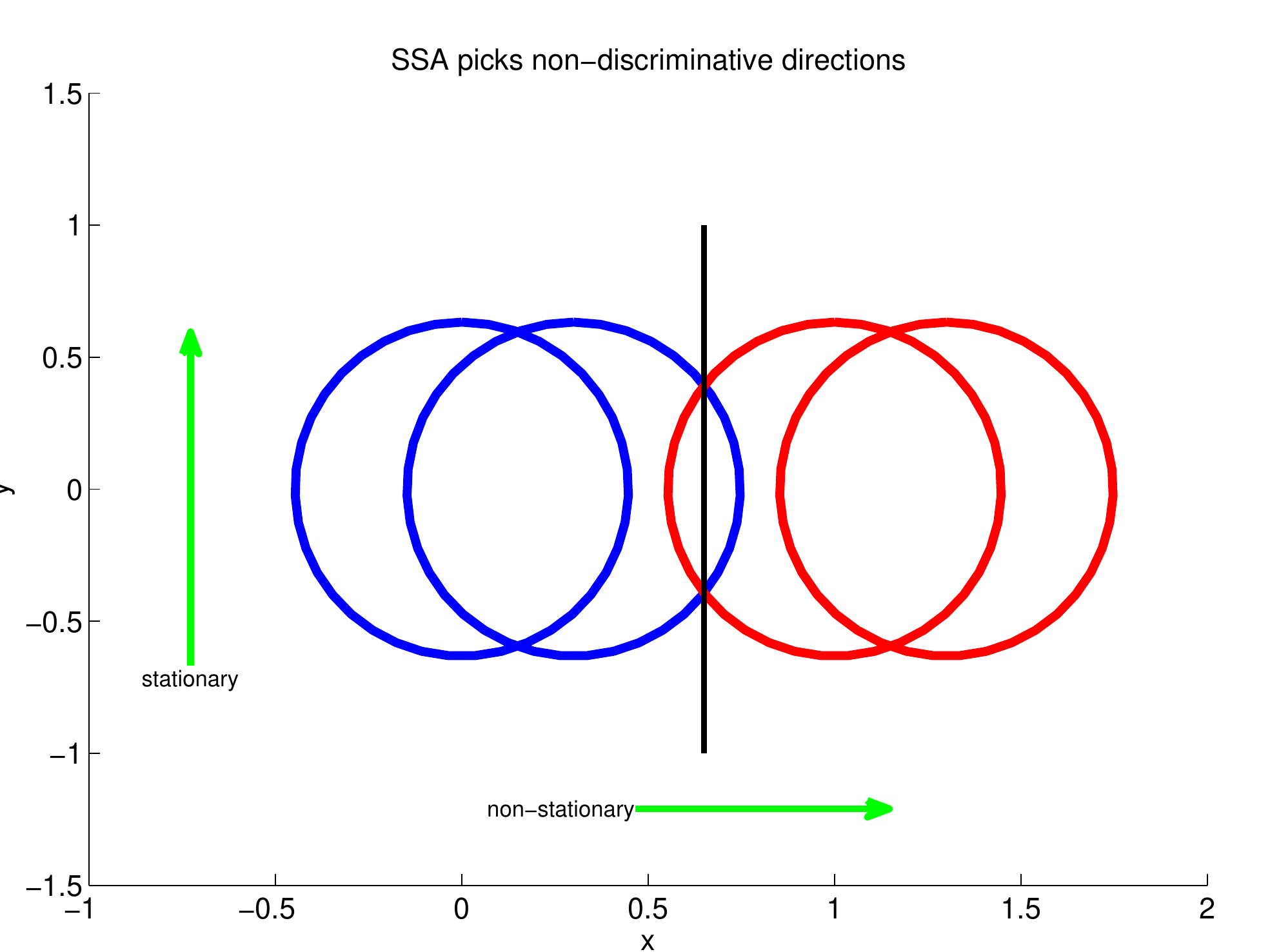}
\end{center}
\caption{\label{fig:SSA_nondiscrim_illustration} The figure provides an illustration of the unsuitability of SSA for classification. The ellipses correspond to the class covariances in each class over two epochs. The $y$-direction corresponds to the direction chosen by SSA and is not discriminative. The $x$-direction is non-stationary but provides discriminative information. Thus the illustration demonstrates that using SSA may discard information which is important for classification.}
\end{figure}

\begin{figure}[ht]
\begin{center}
\includegraphics[width=140mm]{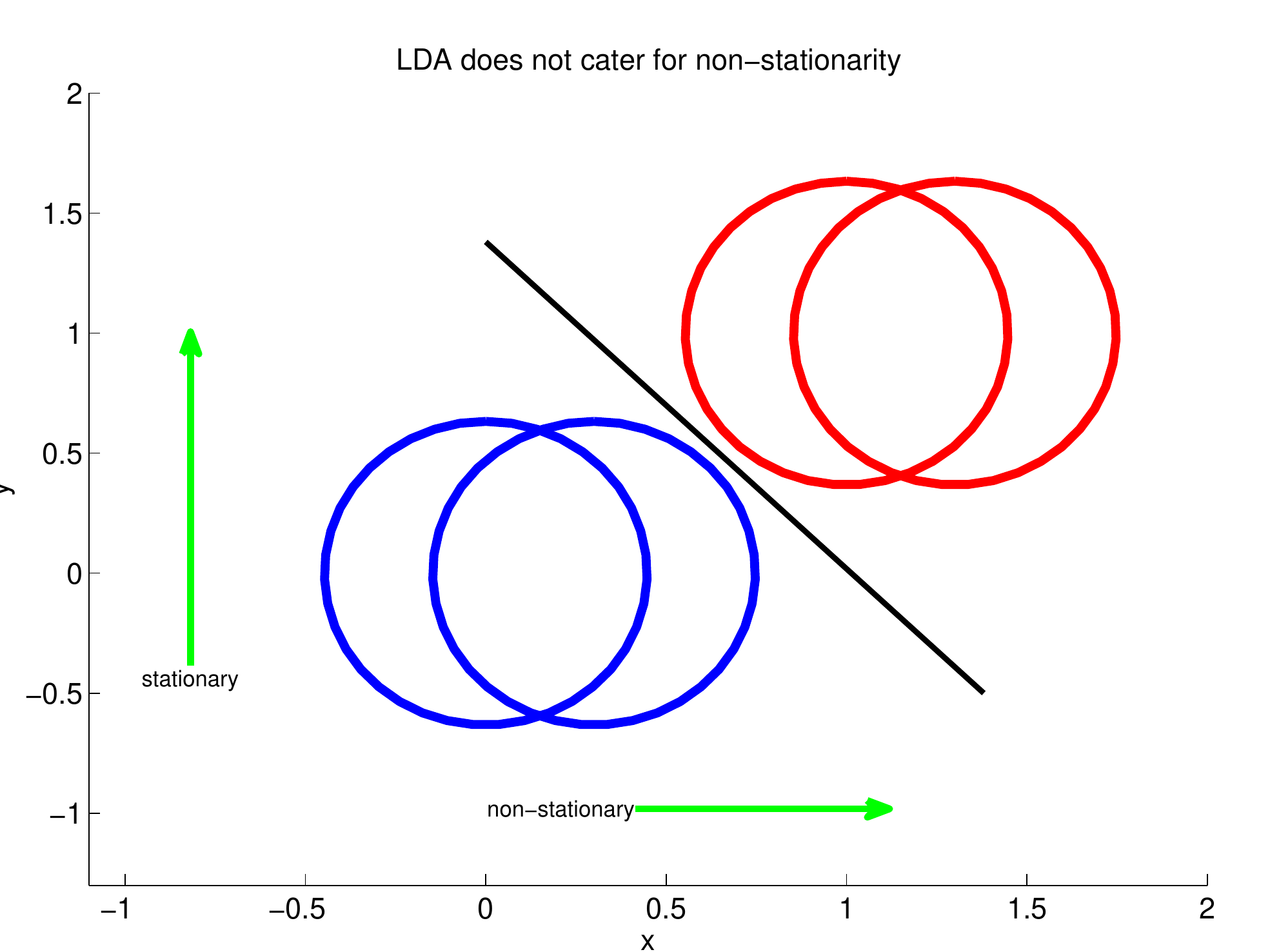}
\end{center}
\caption{\label{fig:LDA_nonstat_illustration} The figure provides an illustration demonstrating the unsuitability of LDA for classification under non-stationarity. The ellipses are schematics of covariances in a two dimensional space in each class (denoted by color) over two epochs. The diagonal direction is the most discriminative direction. The $y$-direction is stationary and discriminative and thus may provide a more robust classification direction than the LDA solution. Thus the illustration demonstrates that by using LDA may choose non-stationary directions, which provide poor generalization error on the test set.}
\end{figure}

Figure~\ref{fig:SSA_nondiscrim_illustration} illustrates the difficulty with using SSA for classification and Figure~\ref{fig:LDA_nonstat_illustration} illustrates the difficulty using LDA. In the first case, SSA may 
select the stationary direction but this direction contains no discriminative information. In the second case, LDA chooses the most discriminative direction on the training data which is, however, not-orthogonal to
the most non-stationary direction. If we assume, that the non-stationarity which is present in the transition from training to testing, is poorly reflected in the training non-stationarity, then, LDA provides a highly suboptimal solution. Thus. neither SSA nor LDA may sufficiently cater for classification under non-stationarity in a unified manner. We aim to address these deficits in Section~\ref{sec:sLDA}.

The solution we propose, therefore, is that we trade off stationarity against discriminability. We describe this ansatz in detail in the following sections.

\section{LDA}
\label{sec:LDA}
One standard approach for training a classifier is Linear Discriminant Analysis (LDA), first proposed in Fisher's landmark article of 1937 \cite{LDA}; LDA outputs a hyperplane with a decision bias given classes modeled by only their mean and covariances:
given that each of the two classes are normally distributed with identical covariance parameters, LDA is the Bayes optimal classifier \cite{BayesOptimal}. The decision criterion given by the hyperplane with normal vector $w$ and bias $b$ \cite{LDA} is given as follows:

\begin{eqnarray}
w &=& (\Sigma_1+\Sigma_2)^{-1}(\mu_1-\mu_2) \\
b &=& w^T\frac{(\mu_1+\mu_2)}{2}
\end{eqnarray}

Here, the $\mu_i$ refer to the class means and the $\Sigma_i$ to the class covariances. The decision rule is then $sign(w^Tx + b)$.
According to the LDA ansatz, $w$ is computed as the direction which maximizes the ratio between the within class covariances $(\mu_1-\mu_2)^T(\mu_1-\mu_2)$ and between class covariance,
$\Sigma_1+\Sigma_2$. The ratio maximized is thus:

\begin{equation}
\frac{(w^T (\widehat{\mu}_1-\widehat{\mu}_2)^2)}{w^T (\widehat{\Sigma}_1 + \widehat{\Sigma}_2) w}
\end{equation}

Optimizing a projection to maximize this ration will often fail for robust classification under non-stationarity, because the ratio does not take this non-stationarity into accound; however, this ratio will be incorporated into the error function described in the following section.

\section{sLDA}
\label{sec:sLDA}

We will now propose a loss function which finds a hyperplane which, in the ideal scenario, performs classification (of course at error-rates above chance) but is robust against non-stationarities. Because
we incorporate the LDA ratio into the loss function, we call the method sLDA, where the `s' stands for `stationary'. Thus sLDA is derived as classification performed on the direction given by Equation~\ref{eq:sLDA_loss} and using the bias described below.

A hyperplane, as a decision boundary, performs a one dimensional projection of the data ($w$) and makes a decision based on a threshold ($b$). It is thus only of interest that the data be as stationary
as possible on this one dimensional projection. This implies that, although the data may be very non-stationary, we only need to find a single dimension which is both discriminative
and stationary to succeed in our task. Using sLDA we aim to choose a direction for classification using a trade off loss function based on the ratio used by LDA but catering for non-stationarity. For the class decision we subsequently use the same threshold as for LDA. 
\vspace{5mm}
\begin{center}
\framebox{\emph{sLDA chooses a decision direction which is discriminative and stationary}.}
\end{center}
\vspace{5mm}

The trade off function, which we maximize in order to choose a direction for classification $w$ (a vector), is:
 
 \begin{equation}
 L_\alpha(w) = \alpha \sqrt{\frac{(w^T (\widehat{\mu}_1-\widehat{\mu}_2)^2)}{w^T (\widehat{\Sigma}_1 + \widehat{\Sigma}_2) w}} + (1-\alpha) \sum_{i=1}^N\sum_{j=1}^2\Phi_{ns}(w;\mu^i_j,\Sigma^i_j,\widehat{\mu}_j,\widehat{\Sigma}_j)
 \label{eq:sLDA_loss}
 \end{equation}
 
 $\Phi_{ns}$ is the Kullback-Leibler divergence between the average epoch mean and the $i$th epoch gaussian estimated from the epoch data: this does not use the assumption of normalized 
 means and covariances. The parameters $\Sigma_j$ and $\mu_j$ denote the empirical mean and covariance of the $j^\text{th}$ class evaluated on the training data. Thus:
 
 \begin{eqnarray*}
\Phi_{ns}(w;\mu^i_j,\Sigma^i_j,\widehat{\mu}_j,\widehat{\Sigma}_j) &=& D_{KL}(E_i || \hat{E})\\
&=& \frac{(w^T(\mu^i_j - \widehat{\mu}_j))^2}{2w^T \widehat{\Sigma}_jw} + \frac{1}{2}\left(\frac{w^T\Sigma^i_jw}{w^T\widehat{\Sigma}_jw}\right) -1 - \text{log}\left(\frac{w^T\Sigma^i_jw}{w^T\widehat{\Sigma}_jw}\right)
 \end{eqnarray*} 

Now we calculate the derivative of $L_{\alpha}$ w.r.t.~$w$. This is a simple matter as we are no longer dealing with rotation matrices as in the first chapter of this thesis.
The derivative is calculated as follows:

\begin{eqnarray*}
\frac{\partial}{\partial w}  \sqrt{\frac{(w^T (\widehat{\mu}_1-\widehat{\mu}_2)^2)}{w^T (\widehat{\Sigma}_1 + \widehat{\Sigma}_2) w}} 
&=& \frac{1}{2} \left( \frac{(w^T (\widehat{\mu}_1-\widehat{\mu}_2)^2)}{w^T (\widehat{\Sigma}_1 + \widehat{\Sigma}_2) w} \right)^{-1/2} \frac{2(w^T(\widehat{\mu}_1-\widehat{\mu}_2))(\widehat{\mu}_1-\widehat{\mu}_2)}{w^T (\widehat{\Sigma}_1 + \widehat{\Sigma}_2) w} \\
&-& \frac{1}{2} \left( \frac{(w^T (\widehat{\mu}_1-\widehat{\mu}_2)^2)}{w^T (\widehat{\Sigma}_1 + \widehat{\Sigma}_2) w} \right)^{-1/2} \frac{2 (\widehat{\Sigma}_1 + \widehat{\Sigma}_2) w (w^T (\widehat{\mu}_1-\widehat{\mu}_2)^2)}{(w^T (\widehat{\Sigma}_1 + \widehat{\Sigma}_2) w)^2}
\end{eqnarray*}

\begin{eqnarray*}
\frac{\partial}{\partial w} D_{KL}(E_i || \widehat{E}) &=&  \frac{2(w^T(\mu^i_j-\widehat{\mu}_j))(\mu^i_j-\widehat{\mu}_j)}{w^T \widehat{\Sigma}_j w} - \frac{2 (\widehat{\Sigma}_j) w (w^T (\widehat{\mu^i}_j-\widehat{\mu}_j)^2)}{(w^T \widehat{\Sigma}_j w)^2} \\
&+& \frac{1}{2} \frac{w^T\widehat{\Sigma^j}w 2\Sigma^i_jw -w^T\Sigma^i_jw 2 \widehat{ \Sigma^j} w}{(w^T\widehat{\Sigma^j}w)^2} - \left(\frac{w^T\widehat{\Sigma^j}w}{w^T\Sigma^i_jw}\right) \times \frac{1}{2} \frac{w^T\widehat{\Sigma^j}w 2\Sigma^i_jw -w^T\Sigma^i_jw 2 \widehat{ \Sigma^j} w}{(w^T\widehat{\Sigma^j}w)^2}
\end{eqnarray*} 

In practice we may attempt choose the regularization parameter $\alpha$ by cross validation. So the final procedure for computing an sLDA classifier given set features is as follows:

\subsection{Final Procedure For Training sLDA}

Here is the final procedure for training the sLDA classifier on set features:

\begin{enumerate}
\item For each of $k$ cross validation folds compute $w_i$ as the maximizer, by gradient ascent \footnote{We use the function {\tt fmincon} included in the matlab optimization toolbox to obtain this minimum: see:-  \url{http://www.mathworks.de/products/optimization/index.html}}, of $L_{\alpha_i}(w)$ with $||w|| = 1$ for each choice of set parameter values $\alpha_1,...,\alpha_i,...,\alpha_r$. 
\item Set $b_i = w_i^T(\mu_1+\mu_2)$ for each of these settings.
\item Perform classification for each of these settings via the standard decision rule: $x$ in class 1 if $w_i^Tx+b_i<0$, class 2 otherwise.
\item Choose $\alpha$ to be the parameter setting performing best on average.
\item Retrain $w$ and $b$ on the entire training set for this fixed optimal $\alpha$ as the maximizer of $L_\alpha(w)$.
\end{enumerate}

Of course, for fixed $\alpha$ the cross-validation folds may be omitted and the direction $w$ is simply selected as the direction maximizing the function $L_\alpha(w)$. To select $\alpha_i,...,\alpha_r$, in practice, a small set of sensible 
parameter values for $\alpha_i$ are chosen by hand, from which the best is chosen by cross validation.

\section{Simulations}
\label{sec:Simulations}

The Simulations in the section aim to test the performance of sLDA on toy data and investigate the conditions under which sLDA should yield improvements in classification under non-stationarity.

\subsection{Overview of Simulations}
\label{sec:Overview_Sim}
The simulations are divided into 3 subgroups. The first subgroup, described in Section~\ref{sec:sanitysim}, is designed to check that any improvements observed using sLDA are not due to implicit regularization or robustness, for instance, due to the use of gradient based optimization. (Regularization as a result of early stopping in gradient based descent is a well documented phenomenon in Machine Learning;
see, for example, \cite{EarlyStopping,TricksOfTheTrade}, for details.) Robustness refers to the ability of estimator to not produce drastically different results in the presence of outliers (see, for example, \cite{oai:eprints.pascal-network.org:3317} for details). In order to test whether regularization may be provided as a result of gradient based optimization we compare closed form LDA (see Section~\ref{sec:LDA}) with the maximum argument $w$ for the Fisher ratio: $\frac{(w^T (\widehat{\mu}_1-\widehat{\mu}_2)^2)}{w^T (\widehat{\Sigma}_1 + \widehat{\Sigma}_2) w}
$ obtained via gradient based optimization. We call this method for obtaining $w$, gradLDA. The various simulations within this regularization and robustness analysis section examine if and when exactly these phenomena may be expected through the comparison of the performance of gradLDA and LDA. In the first simulation (\ref{en:Simple}~\emph{Simple}), the data are generated as a mixture
of sources each of which has approximately equal separation. The second simulation (\ref{en:Outliers}~\emph{Outliers}) uses the same dataset as the first but including outliers. The third simulation (\ref{en:Hard}~\emph{Hard}) tests performance using datasets where very few directions in the data are discriminative, that is to say, the problem of finding a good discriminative direction is harder.
The fourth simulation (\ref{en:Tapered Difficulty} \emph{Tapered Difficulty}) tests the transition which occurs when moving from a dataset similar to \ref{en:Hard} \emph{Hard} to a dataset similar to \ref{en:Simple} \emph{Simple}.

The second group of simulations (see Section~\ref{sec:subspacesim}) tests the performance of sLDA in choosing a specific discriminative yet stationary directions within the dataset when non-stationary and non-discriminative yet stationary directions are also present. The first simulation (\ref{en:Simple2} \emph{Simple}) uses only three epochs and 3 dimensions whereas the second simulation (\ref{en:Realistic} \emph{Realistic}) uses 7 epochs and 6 dimensions. The motivation for the section is simply to test the accuracy of sLDA over LDA in picking a stationary yet discriminative subspace without considering the 
more complicated issue of whether this implies higher generalization issue, which we discuss in Section~\ref{sec:res_subspacesim}. For both of these simulations there is a stationary yet discriminative direction, but also non-stationary and discriminative directions and stationary yet non-discriminative directions. In each case, we test the performance of whether sLDA is more effective at picking the 
stationary and discriminative direction.

The third and final group of simulations are designed to test what difference between test and training distributions is necessary in order to guarantee that the stationary and discriminative direction chosen
by sLDA provides a classification direction which produces higher test error than the direction chosen by LDA. To this end, the first simulation (\ref{sec:transfersim} \emph{\s-space small}) investigates the difference between sLDA and LDA when 
the number of stationary directions is small, whereas the second simulation (\ref{sec:transfersim} \emph{\s-space large}) investigates the difference between sLDA and LDA when the number of stationary directions is larger.

\subsection{Data Setups}

\subsubsection{Sanity Checks for Regularization}

\label{sec:sanitysim}

The first set of simulations test for the presence of regularization and robustness effects of gradient based LDA (gradLDA) versus standard, closed form, LDA. In particular, the following simulations are reported upon:

\begin{itemize}

\item{\emph{Simple:}} \label{en:Simple} The training data consists of 75 points drawn randomly from distinct Gaussians for each class and the test data, 150 points from each class. On each source the classes are generated in class 1 as i.i.d. samples from $\NNN(0,1)$ and in class 2 from $\NNN(0.7,1)$. The dimensionality of the data is $D = 6$. We test the performance in terms of classification error between LDA and gradLDA, i.e. sLDA with $\alpha = 1$.
The results are displayed in the top left panel of Figure~\ref{fig:OO}. 

\item{\emph{Outliers:}} \label{en:Outliers} We next frame a data set which includes outliers added to the data used in the previous simulation (outliers are one type of non-stationarity). Outliers are added sparsely at random time points uniformly to each class; the outliers are generated as samples from a Gaussian with mean $20$ times the size of the original data set and are added to the data on each class.
As before, the training data consists of 75 points from each class and the test data, 150 points from each class. The dimensionality of the data set is $D = 6$.
The results are displayed in the top right panel of Figure~\ref{fig:OO}.

\item{\emph{Hard:}} \label{en:Hard} We further investigate the difference between LDA and gradLDA (see above in Section~\ref{sec:Overview_Sim}) as follows: we set the variance of each class to 1 on each source. 5 of the sources are chosen with class
means differing by $0.2$. For the remaining source we vary the degree of separation observed.  In this simulation, outliers are not included. The results reported as classification errors are displayed in the bottom left panel of Figure~\ref{fig:OO}.

\item{\emph{Tapered Difficulty}} \label{en:Tapered Difficulty} In the final simulation in this section, we retain the data setup from the previous simulation but we hold the separability (absolute value of the difference in means as Gaussians) of the most separable source constant (difference in means $=1.1$), whilst increasing the separability of the second and third most separable sources from $0.1$ to $1.1$. The remaining source means are separated by $0.2$ as before. The results are displayed in the bottom right panel of Figure~\ref{fig:OO}. In all cases the data are randomly mixed orthogonally. 

\end{itemize}

\subsubsection{Subspace Based Comparison of sLDA and LDA}

\label{sec:subspacesim}

In the second set of simulations we investigate simple non-stationary toy data setups and evaluate performance in terms of the angle between the normal vectors to the decision hyperplanes and stationary directions within the data sets. 

\begin{figure}[ht]
 \begin{center}
  \includegraphics[width = 105mm]{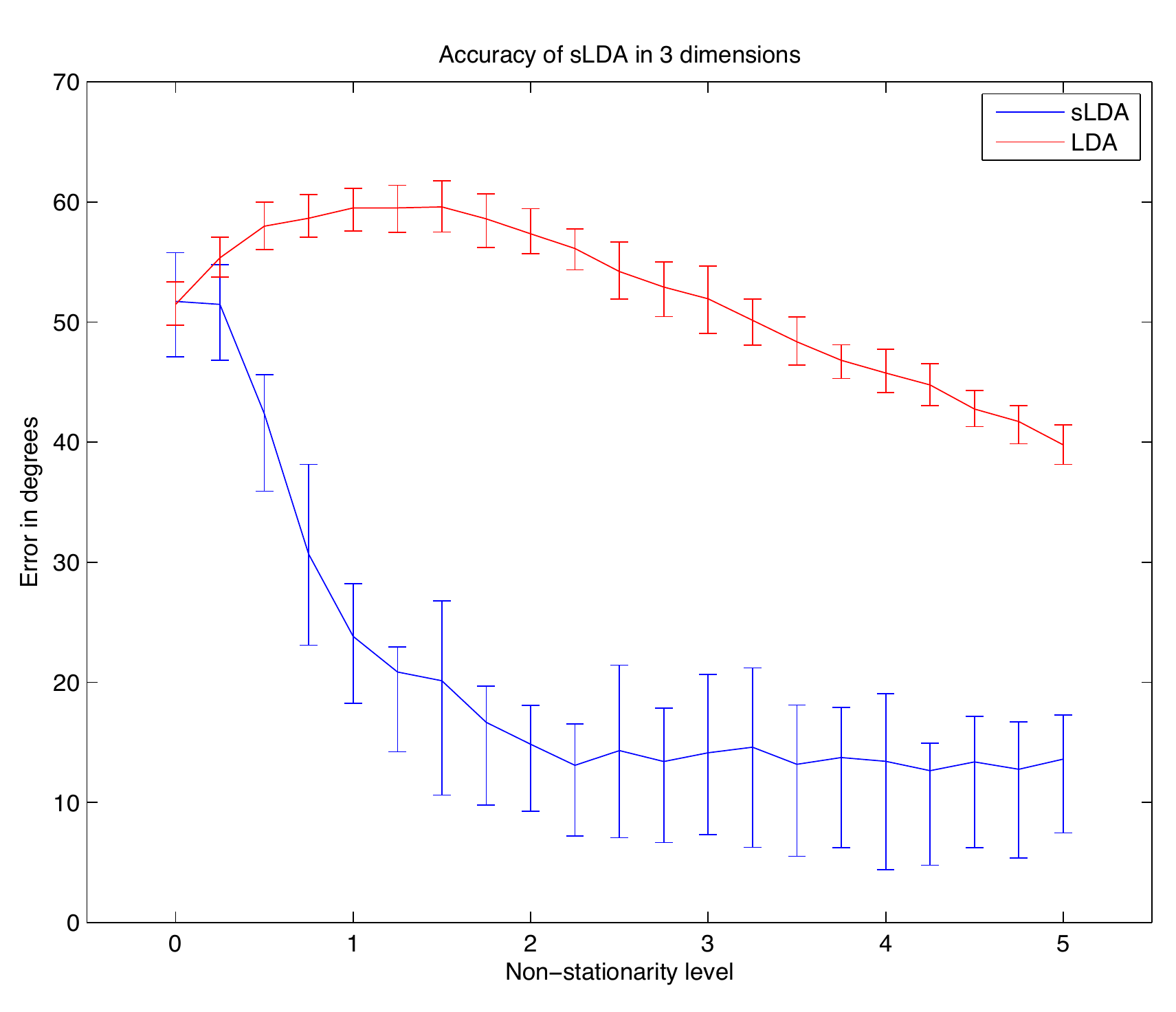}
  \caption{The figure displays the accuracy in degrees between the stable and separable direction and the direction chosen by resp. LDA and sLDA with $\alpha = 0.1$ for the 1st simulation (\ref{sec:subspacesim} \emph{Simple}) described in Section~\ref{sec:subspacesim}. The goal is to pick a single direction which is separable and stationary from a three dimensional space which contains non-separable, stationary directions and separable, non-stationary directions. The $x$-axis displays the non-stationarity level: this corresponds to $a_{ns}$ in the text above and corresponds roughly to the average deviation of the epoch means from the mean over all epochs. The $y$-axis corresponds to the accuracy of methods in choosing the correct direction measured in degrees. From the figure we can see that the higher the non-stationarity level, the more accurate sLDA is. For high non-stationarity levels, LDA is inaccurate. However when the non-stationarity level is very high, LDA becomes
  less inaccurate, because lower separation is induced on the non-stationary source, through this non-stationarity.  \label{fig:sLDA_sim}
    }
 \end{center}
\end{figure}

\begin{itemize}

\item{\emph{Simple:}} \label{en:Simple2} The first data set has three dimensions: $D=3$. In each case, all epoch distributions are Gaussian with variance 1 on each source. The first source is stationary and has no separation between classes. The second source has separation and is stationary; the classes are drawn from Gaussians with unit variance and means, $0$ and $0.7$ respectively.
The third dimension has separation and non-stationarity; the data on this 3rd dimension consists of 3 epochs; the outer 2 epochs are drawn from normal distributions with unit variance and means
$0,1$ respectively. The inner epoch is drawn from a Gaussian distribution which has in class one mean randomly chosen from the uniform distribution on $[-a_{ns}-1, 0]$ and in class 2 randomly chosen from the uniform distribution on $[1,a_{ns}]$ where $a_{ns}$ corresponds to the non-stationarity level.
The data are subsequently mixed orthogonally.  

The aim is to find the second dimension as a subspace and not the first or the third; the second has lower but stable separation as opposed to the 3rd dimension which has higher but unstable separation.
The results are displayed in Figure~\ref{fig:sLDA_sim}.

\item{\emph{Realistic:}} \label{en:Realistic} We next investigate the effect observed in the previous simulation in further detail: in particular, we increase the number of dimensions to $D=6$. We investigate, again, for simplicity, non-stationarity 
in mean. We mix 6 sources orthogonally, which consist of samples from both classes. In each case, on each source and each epoch, 50 samples are drawn from each class. 
On each epoch, on source 1, we add 2 independent random numbers to the mean of both classes. The remaining sources are stationary. In addition, each source is generated with a separation 
between the means of the classes, as follows. For each epoch $i$, the classes on source 1, $\n_i^1$ and $\n_i^2$ are distributed according to Gaussians, where $a_i \sim \NNN(0,\kappa)$.

\begin{eqnarray}
\n_i^1(t) &\sim& \NNN(a_i,1) \\
\n_i^2(t) &\sim& \NNN(\tau+a_i,1) \\
\end{eqnarray}

On source 2, viz. $\s_{sep}$, the classes are distributed according to stationary distributions. 

\begin{eqnarray}
\s_{sep}^1(t) &\sim& \NNN(0,1) \\
\s_{sep}^2(t) &\sim& \NNN(b,1) 
\end{eqnarray} 

And, for $j = 1, \dots, 4$, the 3rd to 6th sources are distributed as follows:
\begin{eqnarray}
\s_{n_j}^1(t) &\sim& \NNN(0,1) \\
\s_{n_j}^2(t) &\sim& \NNN(c,1) 
\end{eqnarray}

The data are randomly mixed orthogonally. On each realization of the data set, we compute directions $w_{LDA}$ an $w_{sLDA}$ and measure the angle between each $w$ and the projection to the second source, $P^{\s_{sep}}$. We set $\tau = 2$ and use epochs $i = 1,...,7$, each of which contains 11 data points per class.
The exact parameter values we fix are as follows: $\tau = 2$, $b = 1.2$, $c = 0.2$. 
The results over 100 realizations of the data set for each value over varying trade-off parameter $\alpha$ and non-stationarity level $\kappa$ are displayed in terms of subspace angle in Figure~\ref{fig:cv5_complex_simulation}.

\end{itemize}

\subsubsection{Investigation of transfer non-stationarity to test-phase}

\label{sec:transfersim}

Finally, we assess the relationship between the non-stationarity observed on the training data with the non-stationarity observed in the transfer between training and test data in terms of classification test error: we aim 
to assess the discrepancy between these non-stationarities which guarantees, on average, an improvement in classification performance using sLDA over LDA.

\begin{itemize}

\item{\emph{\s-space small:}} In the first of these transfer simulations, we retain the second data set used in the fourth set of simulations (Section~\ref{sec:subspacesim}), using the first 7 epochs as the training data each comprising $11$ data points per class. The 8th and final epoch generated is held back as a test set, comprising 150 data points per class and the parameter $a_8$ is now fixed and varied between $0$ and $3$. In addition, the parameter value $\kappa = 0.5$ is fixed. We evaluate average classification performance over the range $a_8 = 0.2, 0.4, \dots, 3.0$.

\item{\emph{\s-space large:}} In the second of these transfer simulations, we retain the same data set as in the previous simulation, with the only difference being that we enlarge the number of significantly separable but stationary directions to 2.
That is, we exchange $\s_{noise_4}$ for an independent copy of $\s_{sep}$ and additionally increase $b$ so that $b = 2$. In addition, the parameter $\tau$ is reduced so that $\tau = 1$ and $c$ is decreased so that $c = 0$. 

\end{itemize}

In both transfer simulations, we study gradLDA as a sanity check against regularization.

\subsection{Results and Discussion}

In the current section we describe the results for the simulations described in Section~\ref{sec:Simulations}.Ä

\subsubsection{Results for Sanity Checks for Regularization}

\begin{figure}[ht]
 \begin{center}
  \includegraphics[width = 140mm]{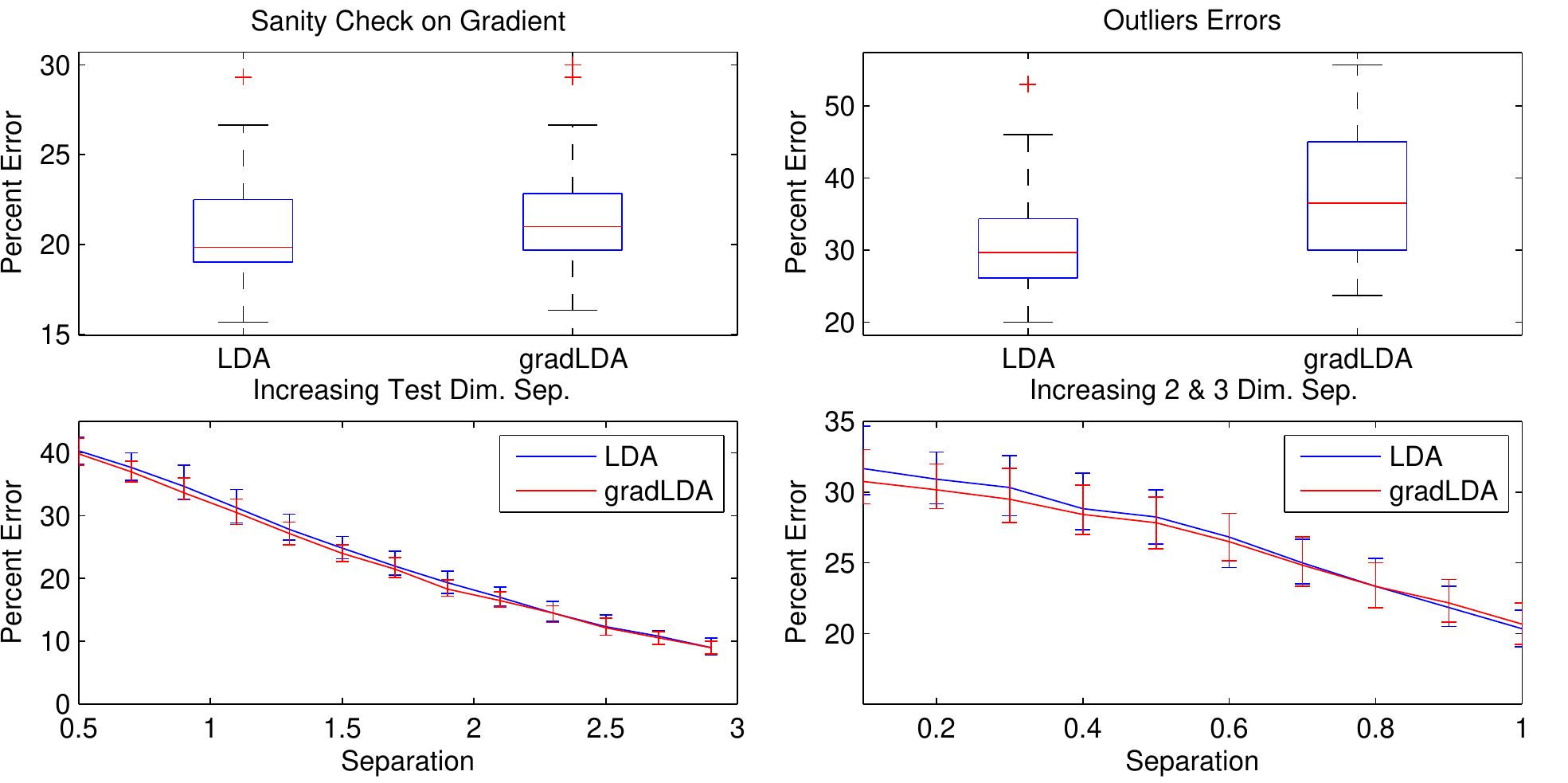}
  \caption{The figure displays the results from the simulations described in Section~\ref{sec:sanitysim}. The top left panel displays the results from  the first simulation (\emph{Simple}) and demonstrates that LDA outperforms
  gradLDA (sLDA + $\alpha = 1$) for stationary data generated by sources which are uniformly separable. The top right panel displays the results from the second simulation (\emph{Outliers}) and shows that gradLDA does not robustify LDA against outliers.  The bottom left panel displays the results from the third simulation (\emph{Hard}) and the bottom right displays the results from the fourth simulation (\emph{Tapered Difficulty}) and show that when the discriminative subspace is smaller (the problem is harder), then regularization effects may be obtained using gradLDA. All performances reported in given in terms of errorates on the $y$-axis. In the boxplots, the whiskers describe the full extent of the data, the box corresponds to the lower quartile, median and upper quartile. In the bottom left panel, the $x$-values represent the level of separation on the single separable source, whereas in the second panel, the $x$-values represent the level of separation on the second and third separable sources. \label{fig:OO}}
 \end{center}
\end{figure}

The results for the simulations described in Section~\ref{sec:sanitysim} are displayed in Figure~\ref{fig:OO}. The comparison between gradLDA and LDA on the first data set (\ref{en:Simple}~\emph{Simple}) show that when 
there are multiple discriminative directions in the data, gradLDA is outperformed by LDA; see the top left panel of the figure. Similarly, for the second data set (\ref{en:Simple}~\emph{Outliers}), see the top right panel, sLDA provides no robustness effect in the presence of outliers.
Thus, gradLDA does not necessarily regularize or robustify LDA through, for instance, early stopping. However, the results from the third and fourth data sets (\ref{en:Simple}~\emph{Hard} and \emph{Tapered Difficulty}) show that when 
the classification problem is more difficult,  improvements in classification performance of up to 1$\%$ are possible using gradLDA. Thus, some regularization 
effects may be obtained, when using gradLDA, when the classification task is difficult.

\subsubsection{Subspace Based Comparison of sLDA and LDA}

\label{sec:res_subspacesim}

\label{sec:subspaceres}

\begin{figure}[ht]
 \begin{center}
  \includegraphics[width = 160mm]{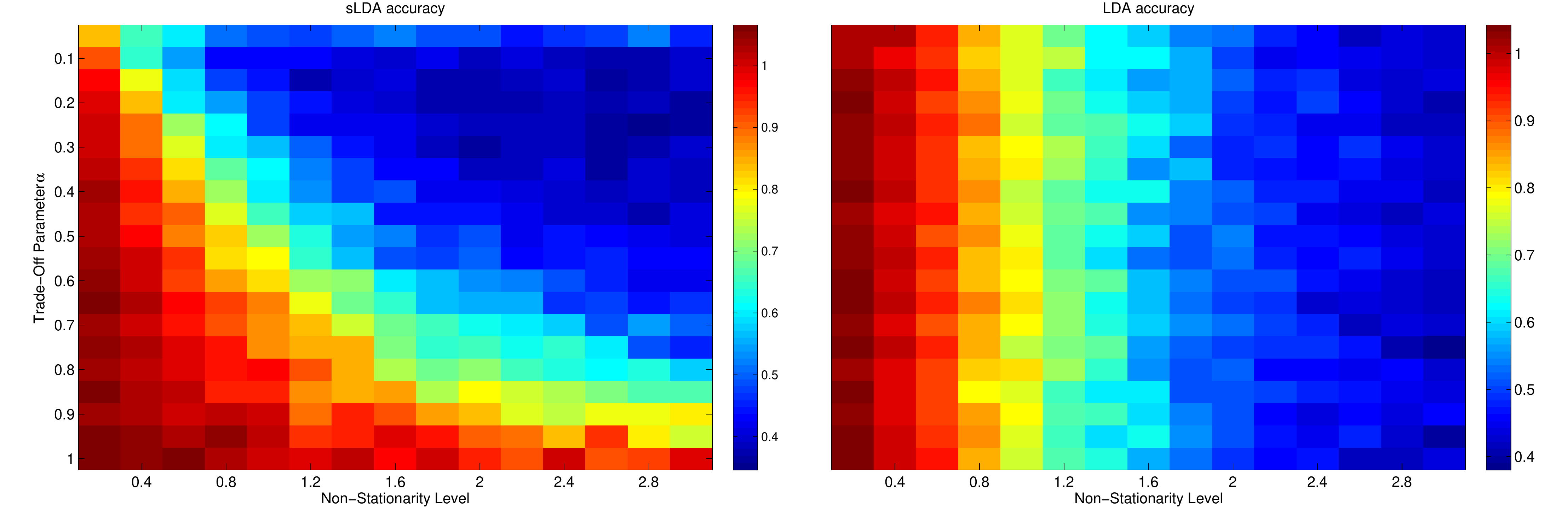}
  \caption{The figure displays the accuracy in radians between the stable and separable direction and the direction chosen by resp. LDA and sLDA for the 2nd simulation described in Section~\ref{sec:subspacesim} (\emph{Realistic}). The data set consists of six dimensions composed of a single stationary and separable source a separable and non-stationary source and four noise sources. The goal is to recover the single stationary and separable source. The $x$-axis displays the non-stationarity level: this corresponds to the average deviation of the epoch means from the mean over all epochs. The $y$ axis corresponds to varying values of the trade-off parameter $\alpha$ in the sLDA loss function $L_{\alpha}(w)$. The color displayed on the panels corresponds to the accuracy in radians in choosing the correct direction for each of the methods. For low non-stationarity levels, sLDA only chooses the correct direction for smaller values of $\alpha$ but for higher non-stationarity levels, sLDA at higher values of $\alpha$ chooses the correct direction. Similarly to the results displayed in Figure~\ref{fig:sLDA_sim}, LDA is more accurate at higher non-stationarity levels due to the lower separation, on average, induced on the non-stationary directions in the data setup. \label{fig:cv5_complex_simulation}
    }
 \end{center}
\end{figure}

Note that we evaluate the quality of the performance of sLDA first in terms of angles between subspaces, rather than in terms of classification accuracy. This is because
classification accuracy is dependent on the difference in distribution between training and test sets: given set training data, we may choose the test data arbitrarily.
We can produce test data which induces an arbitrarily low test performance corresponding to the LDA solution over the sLDA solution: for example, we can frame data sets
whereby a slightly non-stationary subspace in training corresponds to a highly non-stationary subspace in testing. This is achieved by choosing a non-stationarity in 
testing which does not reflect the level of non-stationarity observed in training: see Figure~\ref{fig:ToyExampleLDAnonstat}. Thus, any evaluation scheme based on classification accuracy for a fixed distribution of non-stationarity is data dependent and lacks
objectivity. 

In the first simulation, (\ref{sec:subspacesim}~\emph{Simple}) sLDA finds the correct direction to within 10 to 20 degrees, whereas LDA often chooses the wrong direction. The imperfection in performance of sLDA is due to the fact that the non-stationarity on the first dimension
may not, in every case, be sufficient to outweigh the discriminability available in that direction.

In the second simulation (\ref{sec:subspacesim}~\emph{Realistic}), see Figure~\ref{fig:cv5_complex_simulation}, for the data-set used, sLDA with low values of the hyperparameter $\alpha$ achieve lower angles with the discriminative yet stationary source in the data set
than LDA. As the non-stationarity in training grows, however, LDA improves, since the high level of non-stationarity implies low discriminability.

A further subtlety to note is that, if the parameters of the epoch distributions on the non-stationarity directions of data space are drawn themselves from a probability distribution, then the non-stationarity may be ignored
provided enough epochs are available. For example, if one assumes that the data on each epoch is distributed according to a Gaussian distribution and
that the non-stationarity consists of non-stationarity in mean, then the distribution over the data $r(x;\alpha)$ obtained by integrating out the distribution
over parameters for the mean is also a Gaussian:

\begin{eqnarray}
p(x;\theta) &=& \NNN(x;\theta,\sigma_0^2)\\
q(\theta;\alpha,\beta) &=& \NNN(\theta;\alpha;\beta^2)\\
r(x;\alpha,\beta) &=& \frac{1}{Z} \int p(x;\theta) q(\theta;\alpha,\beta) d\theta\\
&=& \frac{1}{Z} \int \frac{1}{\sqrt{2\pi\sigma_0^2}} e^{-\frac{(x-\theta)^2}{2\sigma_0^2}} \frac{1}{\sqrt{2\pi\sigma_0^2}} e^{-\frac{(\theta-\alpha)^2}{2\beta^2}} d\theta \\
&=& \frac{1}{Z} \int \frac{1}{\sqrt{2\pi\sigma_0^2}} e^{-\frac{(\theta-x)^2}{2\sigma_0^2}} \frac{1}{\sqrt{2\pi\sigma_0^2}} e^{-\frac{(\theta-\alpha)^2}{2\beta^2}} d\theta \\
&=& \NNN(x; \alpha,\sigma_0^2+\beta^2)
\end{eqnarray}	

Thus in the case whereby non-stationarity can be construed as draws of parameters from a probability distribution, given enough data, no extra allowance for non-stationarity
should be made. Thus the sLDA ansatz requires that such a formulation is invalid. In addition, for sLDA to aid classification, the topographies of the non-stationarities
should be constant over time: i.e. the spatial location of the non-stationary sources affecting classification should be the same in the test and in the training data.
These requirements, taken together, constitute very strong assumptions on the data generating process. This requirement is investigated in the classification simulations.

\subsubsection{Results of Transfer Non-Stationarity to Test-Phase Simulations}

\begin{figure}[ht]
 \begin{center}
  \includegraphics[width = 160mm]{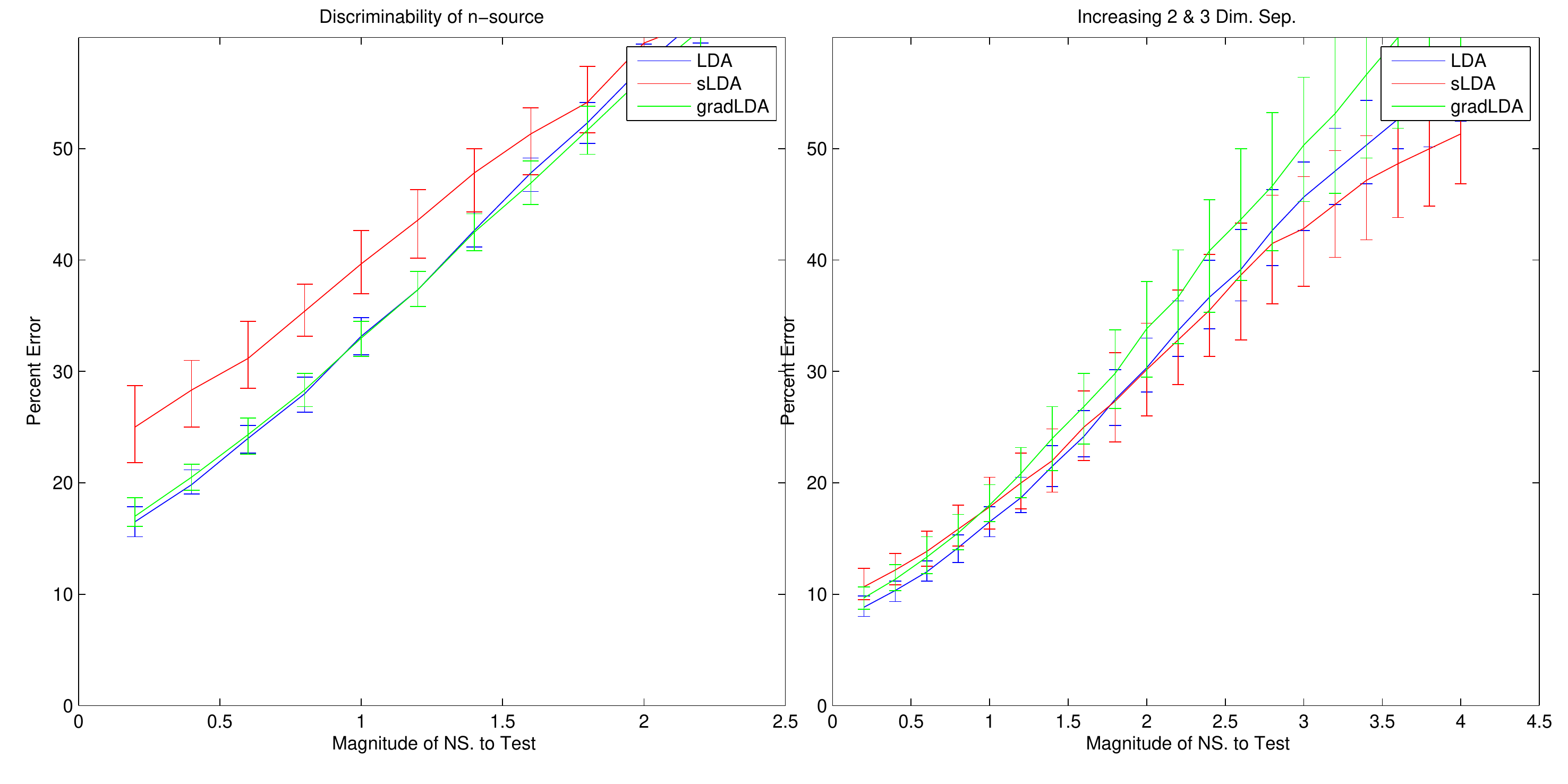}
  \caption{The figure displays the classification accuracy for resp. LDA and sLDA whilst varying the transfer non-stationarity from training to test. For both panels, the data set consists of six sources mixed orthogonally. One of these sources is non-stationary and separable, several are noise directions and resp. 1 or 2 of these sources are separable and stationary. See Section~\ref{sec:transfersim} for details of the data setup. For each panel, the $x$-axis displays the non-stationarity level: this corresponds to the average deviation of the epoch means from the mean over all epochs, whilst the $y$-axis
  corresponds to the classification error achieved.
  The left hand figure displays classification accuracy when the discriminative subspace has dimension approximately 2 (~\ref{sec:transfersim} \emph{\s-space small}) whilst in the right hand figure, the discriminative subspace has 
  dimension of approximately 3 (~\ref{sec:transfersim} \emph{\s-space large}) and the n-source provides less discriminative information than in the simulation corresponding to the left hand panel.  \label{fig:ClassificationSimulations}
    }
 \end{center}
\end{figure}

\begin{figure}[ht]
 \begin{center}
  \includegraphics[width = 120mm]{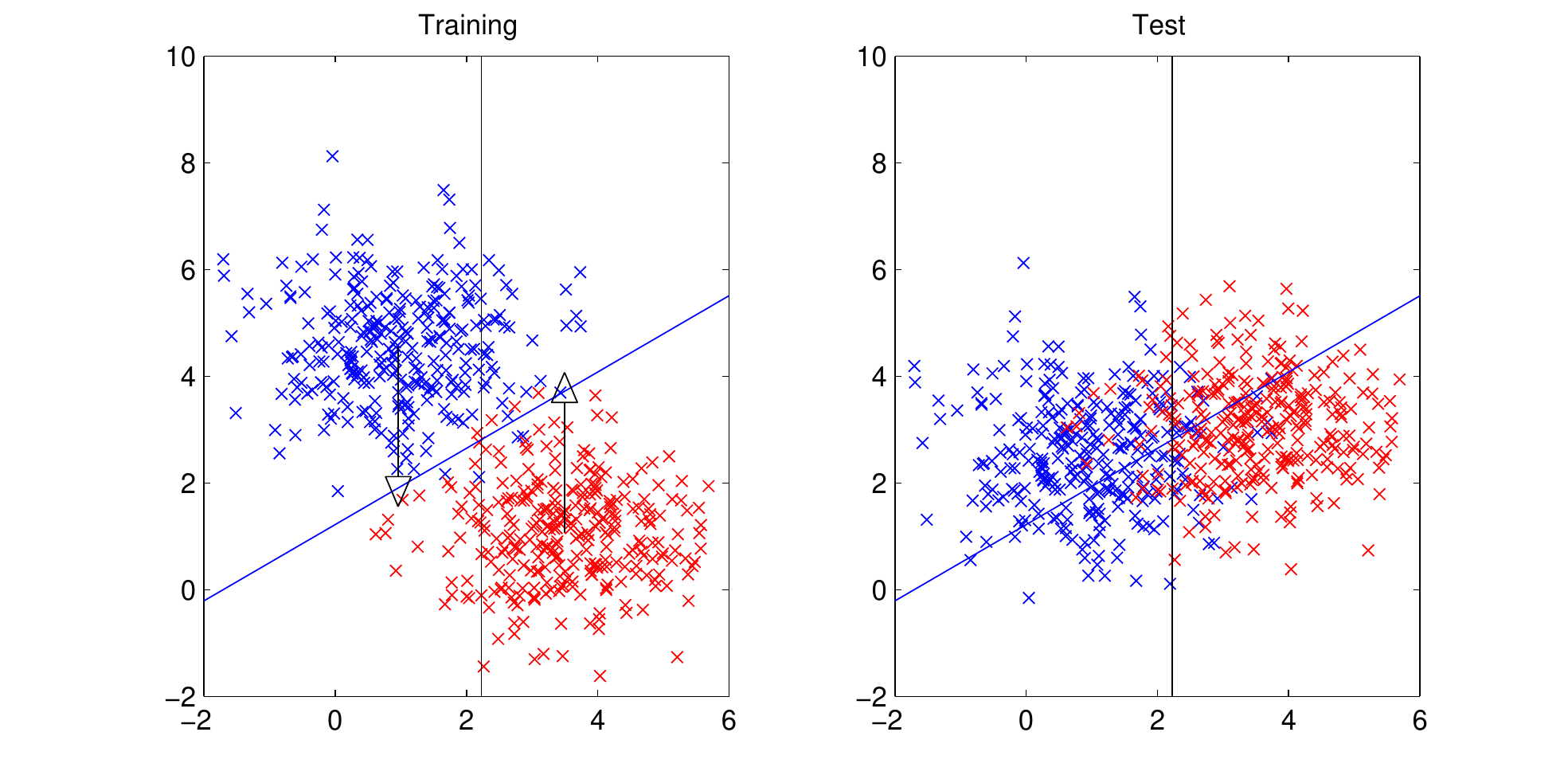}
  \caption{The figure displays an example in which the $y$-direction is non-stationary and the $x$-direction is stationary; although the $y$-direction has higher separation in training, the $x$-direction is to some extent separable and is more stationary. 
  In this case, the black decision line may yield a lower error in testing than the LDA solution in blue.  \label{fig:ToyExampleLDAnonstat}}
 \end{center}
\end{figure}

The simulations in Section~\ref{sec:transfersim} are aimed at achieving a partial intuition of the subtleties discussed in Section~\ref{sec:subspaceres}.
In particular, at a fixed parameter value for the trade-off parameter $\alpha = 0.1$ which achieves high subspace accuracy in the subspace simulations, we observe in Figure~\ref{fig:ClassificationSimulations} that this does not necessarily entail improvement of sLDA over LDA in the classification task, even under arbitrary non-stationarity. Rather, it is shown that, if the stationary but
discriminative directions (\ref{sec:transfersim} \emph{\s-space small}) are not significantly more separable than the non-stationary but discriminative direction, then improvement is not possible. On the other hand, if there are a number of stationary directions which are discriminative and there is one non-stationary (\ref{sec:transfersim} \emph{\s-space} large) but moderately discriminative direction, then improvement is possible using sLDA over LDA. In addition, it seems that gradLDA fares even worse
than LDA in this case: this suggests that gradLDA loses classification accuracy through poor optimization. This suggests, in addition, that further improvement, in this case, is plausible, given a tidier optimization approach for sLDA.

An example in which sLDA yields improvement over LDA is displayed in Figure~\ref{fig:ToyExampleLDAnonstat}. Whilst the direction chosen by LDA on the training data displays clear separation, the opposite is true for the LDA test data. On the other hand, while the separation on the direction chosen by sLDA on the training data is not so clear, the distributions remain more stable in the transition to the test data.

\section{Experiments}
\label{sec:experiments}

\subsection{Description of Experiments}

We evaluate the performance of the above method, sLDA, on BCI data combined with preprocessing with CSP making a reduction to a fixed number, $h_{CSP}$, of features before classification. The data consist of 80 BCI subjects taken from the study \cite{NeuroPred}. 
The baseline procedure for classification using CSP with LDA, which we use, follows the standard template for processing EEG for motor imagery classification \cite{SpatFilt}. See Algorithm~\ref{alg:CSPLDA}.

\begin{algorithm}
\begin{algorithmic}[1]
\caption{ \label{alg:CSPLDA} \emph{CSP+LDA}}
\item Find the discriminative frequency band. \label{freq}
\item Compute a number, $h_{CSP}$, of feature directions. \label{CSP}
\item For each trial compute the variance giving $2m$ data points, one for each trial, with $m$ in each class.
\item Train the LDA classifier on the entire training data consisting of these $2m$ data points.
\item Test the classifier's performance on the test set.
\end{algorithmic}
\end{algorithm}

For sLDA the procedure is defined as per Algorithm~\ref{alg:CSP+alpha_sLDA}:

\begin{algorithm}
\begin{algorithmic}[1]
\caption{ \label{alg:CSP+alpha_sLDA} \emph{CSP+sLDA} with $\alpha = \hat{\alpha}$}
\item Find the discriminative frequency band. \label{freq}
\item Compute a number, $h_{CSP}$, of feature directions. \label{CSP}
\item For each trial compute the variance giving $2m$ data points, one for each trial, with $m$ in each class.
\item Train the sLDA with $\alpha = \hat{\alpha}$ classifier on the entire training data consisting of these $2m$ data points.
\item Test the classifier's performance on the test set.
\end{algorithmic}
\end{algorithm}

In addition we test selection of $\alpha$ via cross validation as in Algorithm~\ref{alg:CSP+sLDA}.

\begin{algorithm}
\begin{algorithmic}[1]
\caption{\emph{CSP+sLDA c.v.} \label{alg:CSP+sLDA}}
\item Find the discriminative frequency band. \label{freq}
\item Compute a number, $h_{CSP}$, of feature directions using CSP. \label{CSP}
\item For each trial compute the variance giving $2m$ data points, one for each trial, with $m$ in each class.
\item Perform $k$-fold classification on the training data for trade-off parameter settings $\alpha_1,...,\alpha_R$, on each fold evaluating 
the performance using sLDA at the chosen parameter setting $\alpha_r$.
\item Retrain the sLDA classifier on the entire training data consisting of these $2m$ data points using the best parameter setting $\alpha$.
\item Test the classifier's performance on the test set.
\end{algorithmic}
\end{algorithm}

Although we argued above that by the nature of the problem, SSA is an unsuitable method for classification, we include results
for SSA for completeness and by way of a check. In particular we preprocess with SSA as per Algorithm~\ref{alg:SSA+CSP+LDA}.
\begin{algorithm}
\caption{\label{alg:SSA+CSP+LDA} \emph{SSA+CSP+LDA}}
\begin{algorithmic}[1]
\State Find the discriminative frequency band. \label{freq}
\For {$d_s = D-30, \dots, D-1$}
\State 
\begin{minipage}[t]{140mm}
Perform a $k$-cross validation fold by reducing the number of dimensions to $d_s$ with SSA: perform the
CSP+LDA (Algorithm~\ref{alg:CSPLDA}) on $P^{\s}X$ where X is the input data and $P^{\s}$ is the projection to the 
stationary sources found using SSA.
\end{minipage}
\EndFor
\State Chosen the optimal $d_s$ as the $d_s$ which minimizes the cross validation error.
\State Retrain the standard CSP+LDA method (Algorithm~\ref{alg:CSPLDA}) 
on $P^{\s}X$ for the projection, $P^{\s}$, to the stationary sources found with SSA at $d_s$.
at this optimal parameter setting.
\end{algorithmic}
\end{algorithm}

By way of an extra sanity check, we test against rLDA \cite{rLDA}: rLDA works by regularizing the covariance estimates against extreme eigenvalues which occur when estimating covariance from small sample sizes. By testing the difference in performance between the stationarity-penalizing method sLDA and the regularizing method rLDA,
we aim to check that any increases in performance are not (completely) due to implicit regularization (for small sample size) of the covariance estimates. 
The steps are displayed in Algorithm~\ref{alg:CSP_rLDA}.

\begin{algorithm}
\caption{\emph{CSP+rLDA} \label{alg:CSP_rLDA}}
\begin{algorithmic}[1]
\item Find the discriminative frequency band. \label{freq}
\item Compute a number, $h_{CSP}$, of feature directions using CSP. \label{CSP}
\item For each trial compute the variance giving $2m$ data points, one for each trial, with $m$ in each class.
\item Train the LDA classifier on the entire training data consisting of these $2m$ data points using regularized estimates 
of the covariance matrix at the optimal shrinkage parameter.
\item Test the classifier's performance on the test set.
\end{algorithmic}
\end{algorithm}

As yet another sanity check, we test the following method: we perform LDA via gradient based optimization: that is to say sLDA with the trade off parameter set to unity: $\alpha = 1$. We will 
often refer to this approach as "gradLDA".
The steps are displayed in Algorithm~\ref{alg:CSP_gradLDA}.

\begin{algorithm}
\caption{\emph{CSP+gradLDA} \label{alg:CSP_gradLDA}}
\begin{algorithmic}[1]
\item Find the discriminative frequency band. \label{freq}
\item Compute a number, $h_{CSP}$, of feature directions using CSP. \label{CSP}
\item For each trial compute the variance giving $2m$ data points, one for each trial, with $m$ in each class.
\item Train the sLDA classifier on the entire training data consisting of these $2m$ data points using $\alpha = 1$
\item Test the classifier's performance on the test set.
\end{algorithmic}
\end{algorithm}

The rational for this is to test whether any improvements in performance are due to, for instance, implicit regularization at local minima 
or in early stopping of gradient based optimization \cite{EarlyStopping}.

As a final sanity check we test the performance of the following method:
instead of performing sLDA and penalizing non-stationarity, we test the performance of a trade-off between 
the Fisher error function and a random homogeneous polynomial of degree 2 in the coefficients of the 
putatively discriminative direction $w$.

That is, we use the error function:

 \begin{equation}
 L_{rand}(w) = \alpha \sqrt{\frac{(w^T (\hat{\mu_1}-\hat{\mu_2})^2)}{w^T (\hat{\Sigma_1} + \hat{\Sigma_2}) w}} + (1-\alpha) w^T R w
 \end{equation}

Here, $R$ is a random square matrix with coefficients drawn uniformly from $[0,1]$.
So the method runs as per Algorithm~\ref{alg:CSP_randLDA}:

\begin{algorithm}
\caption{\emph{CSP+randLDA}  with $\alpha = \hat{\alpha}$ \label{alg:CSP_randLDA}}
\begin{algorithmic}[1]
\item Compute a random matrix $R$.
\item Find the discriminative frequency band. \label{freq}.
\item Compute a number, $h_{CSP}$, of feature directions using CSP. \label{CSP}.
\item For each trial compute the variance giving $2m$ data points, one for each trial, with $m$ in each class.
\item Train the randLDA classifier with $\alpha = \hat{\alpha}$ on the entire training data consisting of these $2m$ data points.
\item Test the classifier's performance on the test set.
\end{algorithmic}
\end{algorithm}

The rationale for testing randLDA, is that it may be due to a data dependent property that improvement using sLDA is possible: a trade off with a random 
error function may also achieve performance increases. We therefore include a random trade off for comparison's sake.

In summary we test the performance of the following basic methods: 

\begin{enumerate}
\item CSP + LDA (Algorithm~\ref{alg:CSPLDA})
\item CSP + rLDA (Algorithm~\ref{alg:CSP_rLDA})
\item CSP + sLDA c.v. (Algorithm~\ref{alg:CSP+sLDA})
\item CSP + sLDA with $\alpha = \hat{\alpha}$ (Algorithm~\ref{alg:CSP+alpha_sLDA})
\item SSA + CSP + LDA (Algorithm~\ref{alg:SSA+CSP+LDA})
\item CSP + gradLDA (Algorithm~\ref{alg:CSP_gradLDA})
\item CSP + randLDA (Algorithm~\ref{alg:CSP_randLDA}) with $\alpha = \hat{\alpha}$.
\end{enumerate}

We test the classification performance of all of these methods on the 80 subjects available and report the results below. In particular $h_{CSP} = 6$. The number of epochs
chosen for evaluating stationarity in sLDA is 7; this guarantees the determinacy of any 1 dimensional projection for $h_{CSP} \leq 12$ (for details see \cite{PRL:SSA:2009}). We choose $k = 5$, for cross validation for sLDA
and for SSA. The training data in each case consists of 75 trials from each class and the test data consists of 150 trials from each class.  

\section{Results and Discussion of Experiments}

\subsection{Layout of Figures}

\begin{figure}[ht]
 \begin{center}
  \includegraphics[width = 140mm]{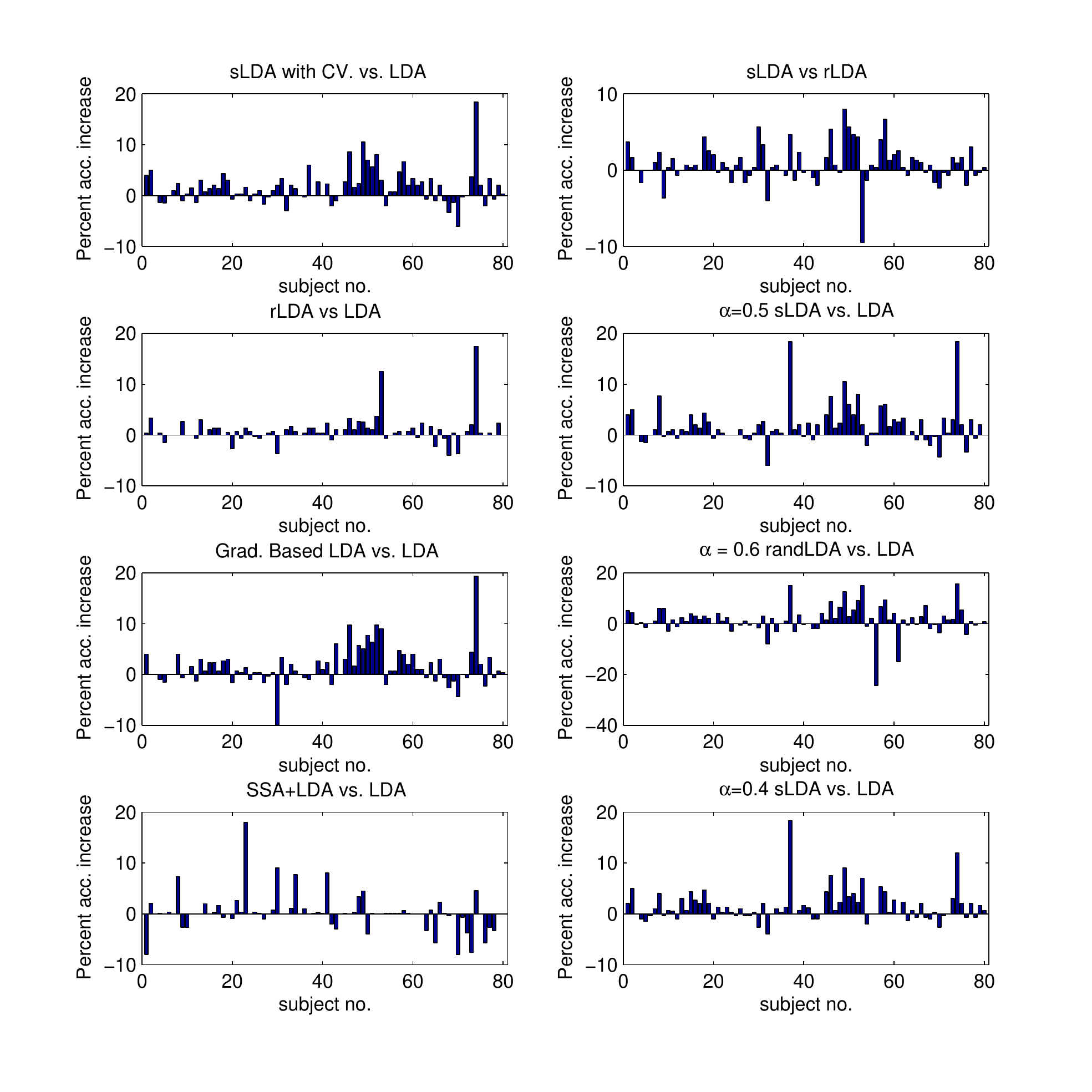}
  \caption{The figure displays results given as differences in error-rates in percent on individual subjects between the methods, LDA, sLDA with $\alpha = 0.5,0.4$ and rLDA, gradLDA, randLDA and sLDA with cross validation with CSP processing on the 80 BCI motor-imagery subjects. In each case the comparison is displayed in the title to the corresponding panel. 
  The $x$-axis corresponds to individual BCI motor imagery subjects and the $y$-axis corresponds to the difference in performance between the methods compared.
  The results
  show that sLDA yields an improvement over LDA which cannot be explained in terms of implicit eigenvalue spectrum related regularization but may be explained
  in terms of a different form of regularization or otherwise, due to improvement observed for randLDA. sLDA with $\alpha = 0.5$ has the highest mean accuracy due to the presence of fewer lower tail 
  outliers.   \label{fig:perchan1}
    }
 \end{center}
\end{figure}

\begin{figure}[ht]
 \begin{center}
  \includegraphics[width = 160mm]{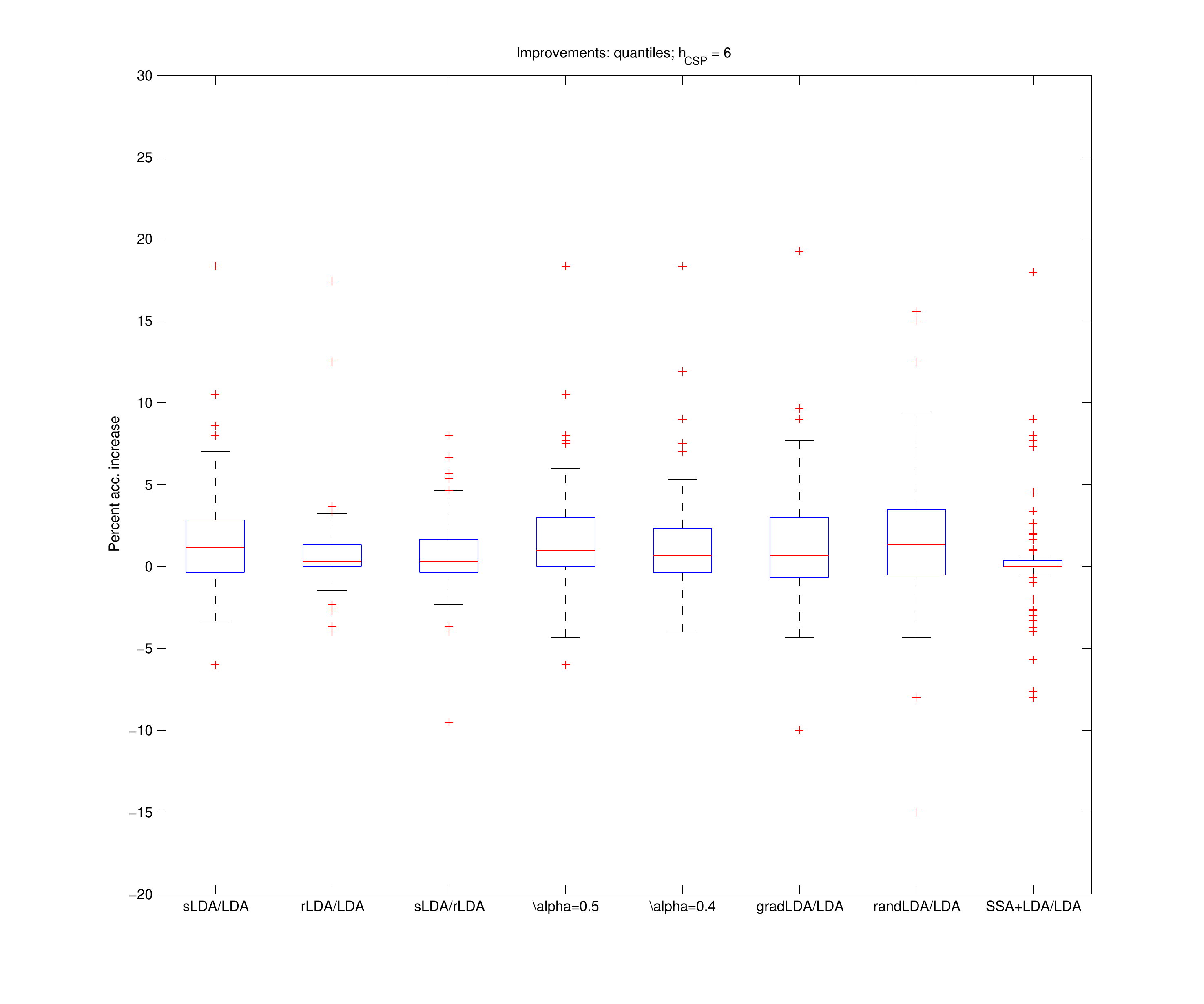}
  \caption{The figure displays the quantiles (the whiskers describe the full extent of the data, the box corresponds to the lower quartile, median and upper quartile; the crosses correspond to outliers) over improvement on the 80 BCI motor- imagery subjects in classification accuracy for the methods  LDA, sLDA with $\alpha = 0.5$ and rLDA, gradLDA, randLDA and sLDA with cross validation, in percentage terms. The comparison drawn is displayed on the $x$-axis. The $y$-values display the corresponding 
  increases in accuracy.     \label{fig:heavytails}}
 \end{center}
\end{figure}

\begin{figure}[ht]
 \begin{center}
  \includegraphics[width = 160mm]{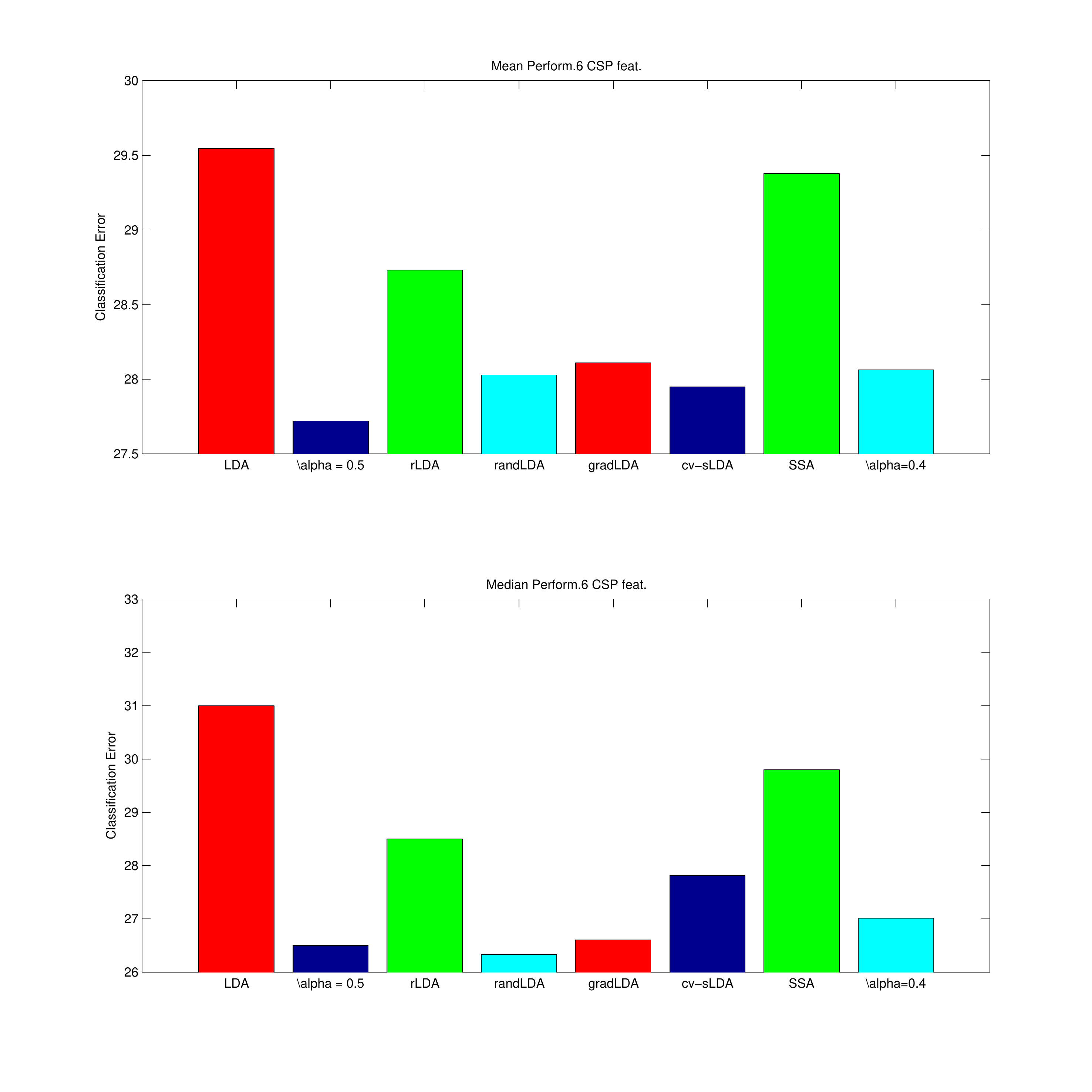}
  \caption{ The upper panel figure displays the mean average classification accuracy on the 80 BCI motor imagery data sets in percentage terms for LDA, sLDA with $\alpha = 0.5$ and rLDA, gradLDA, randLDA and sLDA with cross validation. The $x$-values represent the individual methods and the $y$-values the corresponding accuracies. The lower panel displays the corresponding medians.
  Whilst sLDA with $\alpha = 0.4$ obtains the highest mean error, randLDA obtains the highest median error. (More trials for randLDA are lower tail outliers.)   \label{fig:avper} 
    }
 \end{center}
\end{figure}

\begin{figure}[ht]
 \begin{center}
  \includegraphics[width = 120mm]{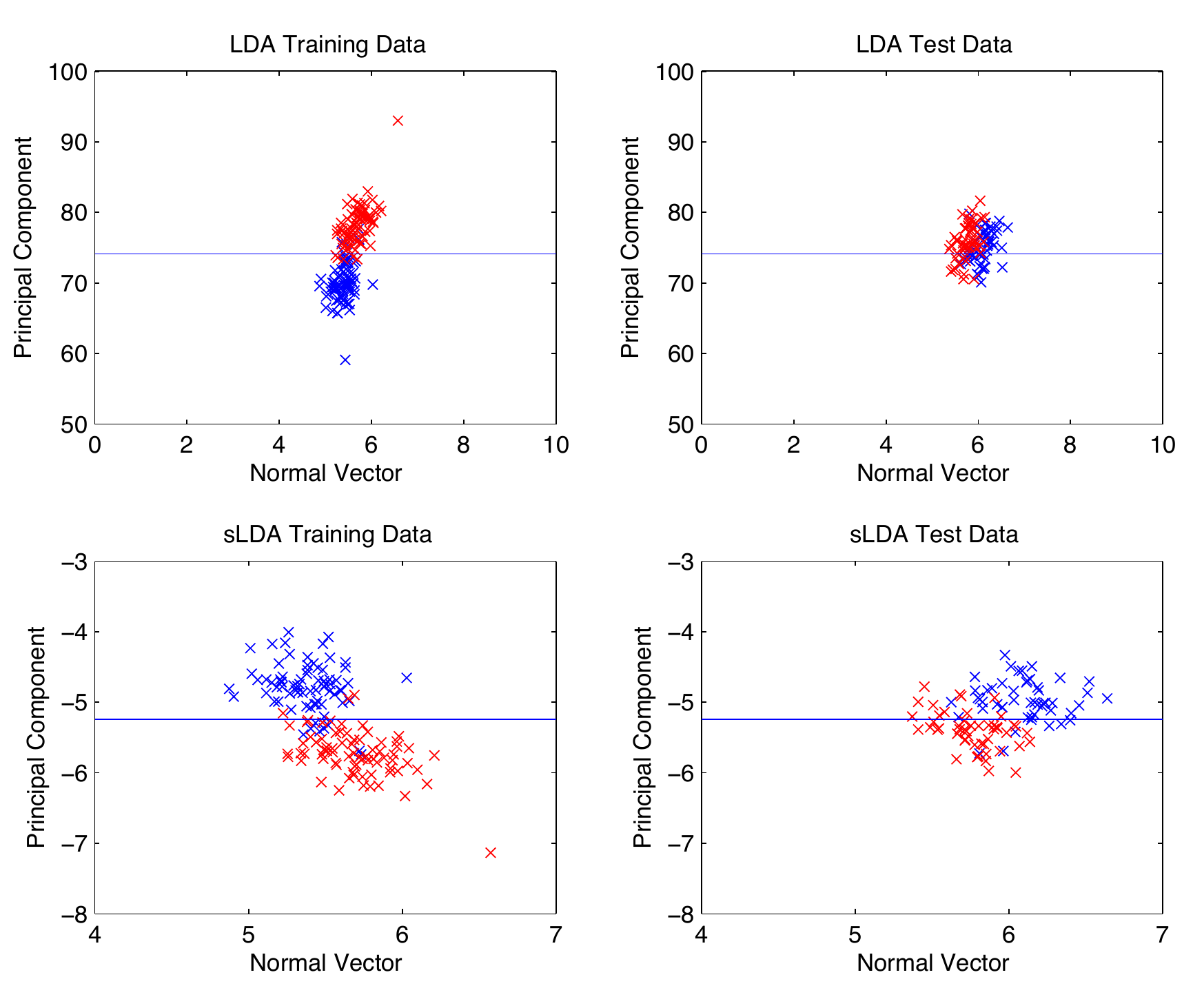}
  \caption{ The top two panels display the difference modulo the LDA hyperplane between training and testing for a single BCI motor imagery subject (subject no. 74). The bottom two panels display the difference modulo the sLDA hyperplane between training and testing for the same subject.   The blue lines display the decision boundary for each classifier. For clarity we plot the principal component of the pooled data against 
  the normal vector to each respective hyperplane. Whilst the class distributions shift significantly towards the hyperplane for the LDA solution, the shift is less pronounced for sLDA. \label{fig:bestimprovement}
    }
 \end{center}
\end{figure}

\begin{figure}[ht]
\begin{center}
\includegraphics[width=120mm]{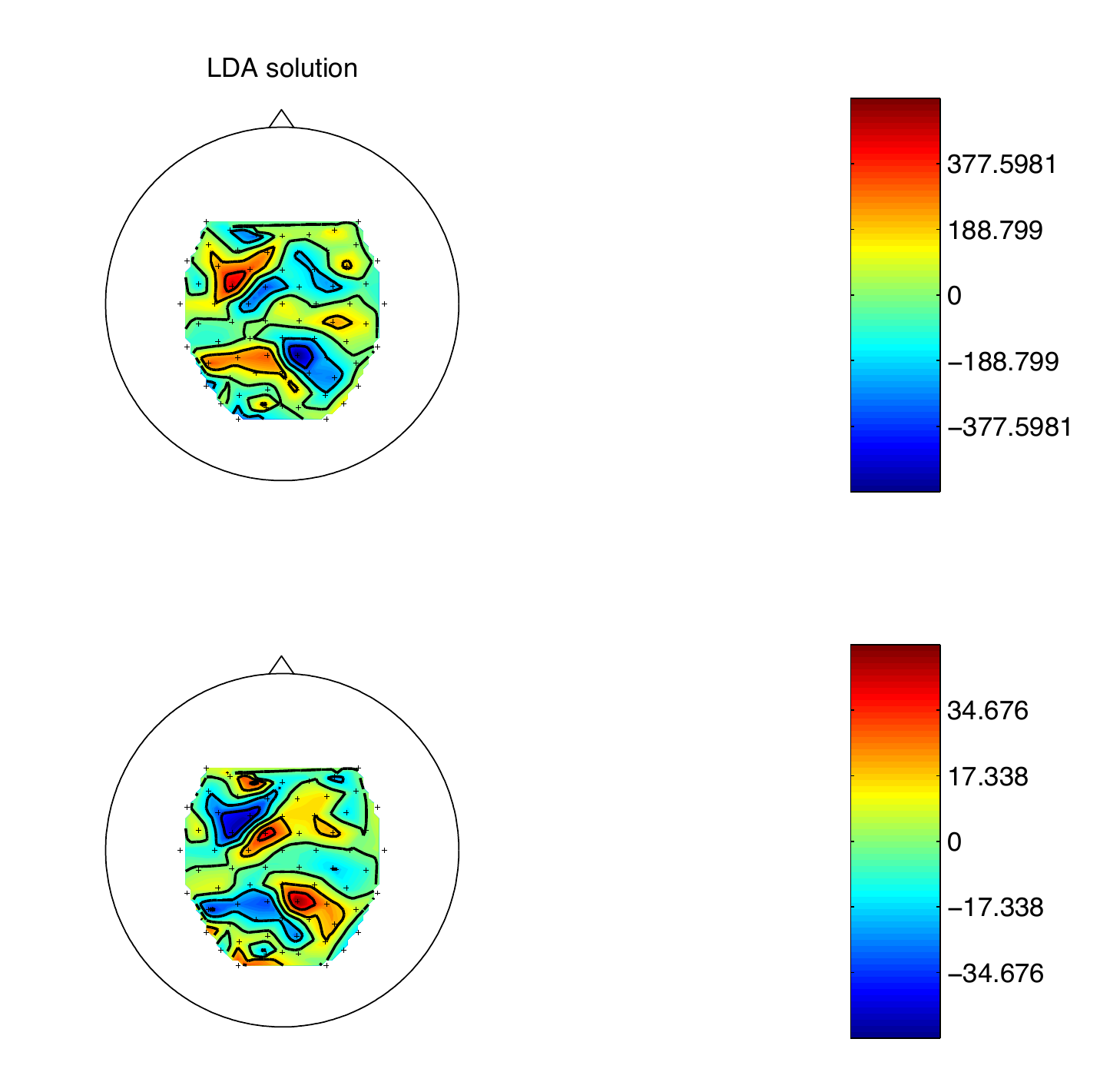}
\end{center}
\caption{\label{fig:bestscalp} The figure displays scalp plots for the LDA and sLDA solutions for the subject whose improvement was greatest using sLDA (subject no.74). In upper panel, the 
field pattern for LDA is displayed, whereas in the bottom panel, the difference for LDA is displayed. The difference in topography is most striking at the right central to parietal electrode positions.} 
\end{figure}

\begin{figure}[ht]
\begin{center}
\includegraphics[width=120mm]{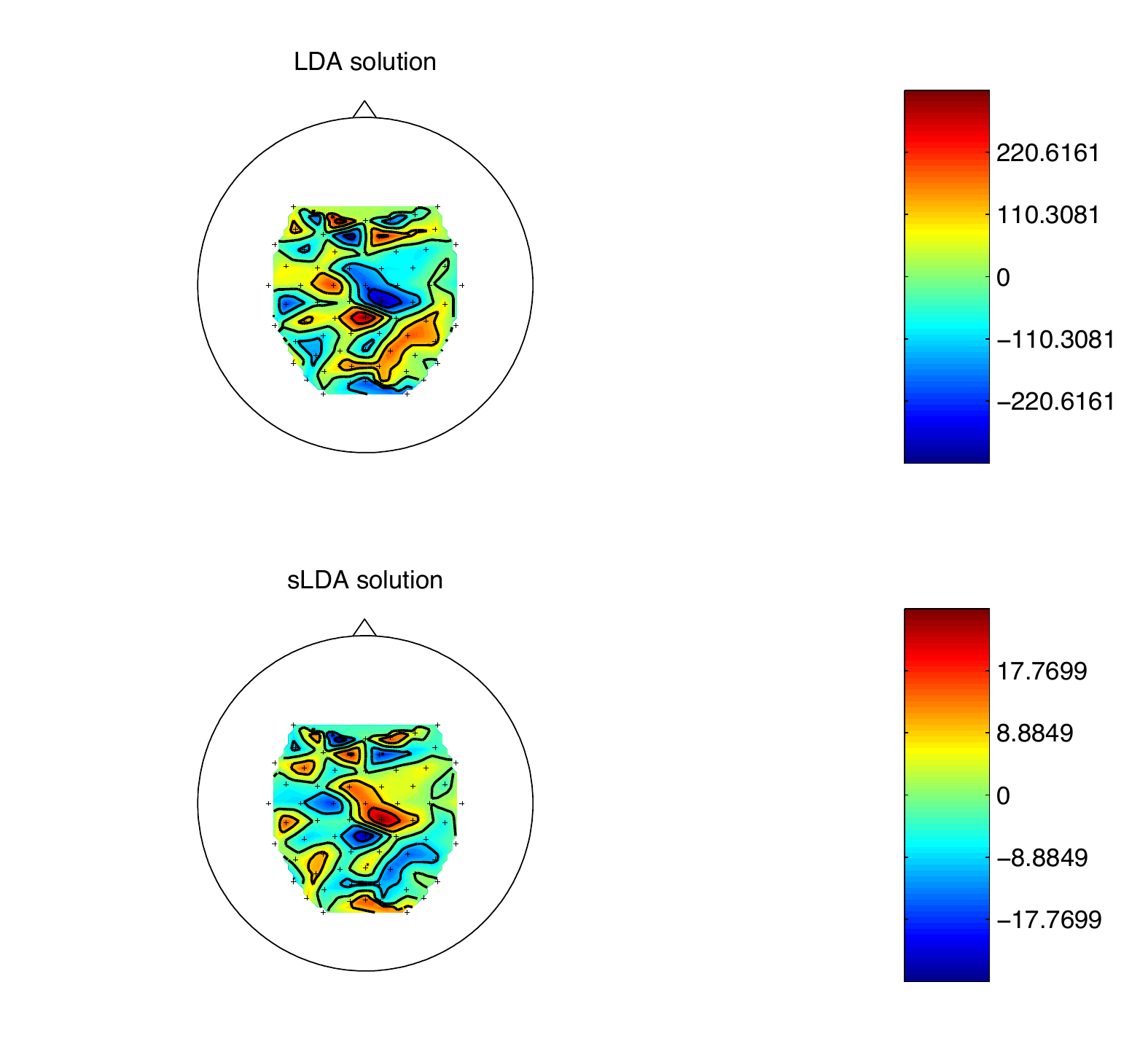}
\end{center}
\caption{\label{fig:worstscalp} The figure displays scalp plots for the LDA and sLDA solutions for the subject whose improvement was most inferior using sLDA (subject no.71). In upper panel, the 
field pattern for LDA is displayed, whereas in the bottom panel, the difference for LDA is displayed. The difference in topography is most striking at the right central to frontal electrode positions.}
\end{figure}

Figure~\ref{fig:perchan1} displays the individual differences in subject performances between a selection of the methods tested for the setting $h_{CSP}=6$.
Figure~\ref{fig:avper} displays the average performance of each of the methods at a selection of parameter settings.
Figure~\ref{fig:heavytails} displays the quantiles of the improvement in performance for the comparisons made individually in Figure~\ref{fig:perchan1}. 
Figure~\ref{fig:bestimprovement} displays an example (subject no. 74) where sLDA induces a large improvement over LDA.
Figure~\ref{fig:bestscalp} displays the corresponding scalp plots for subject no. 74 and Figure~\ref{fig:worstscalp} displays the scalp plots of the sLDA and LDA solutions
for the subject (no.~71) for which sLDA performed worst. 
The corresponding $p-values$ for:
\begin{equation*}
\text{``}H_0 = \{\text{CSP+$y$LDA has an error-rate $\leq$ to the error-rate of CSP+LDA}\} \text{ for }y = s, r \text{''}
\end{equation*}
are displayed using student's paired one sided $t$-test in
Table~\ref{ttest} and using the Wilcoxon sign rank test for equality of medians in Table~\ref{signrank}.


\begin{figure}[ht]
 \begin{center}
  \includegraphics[width = 140mm]{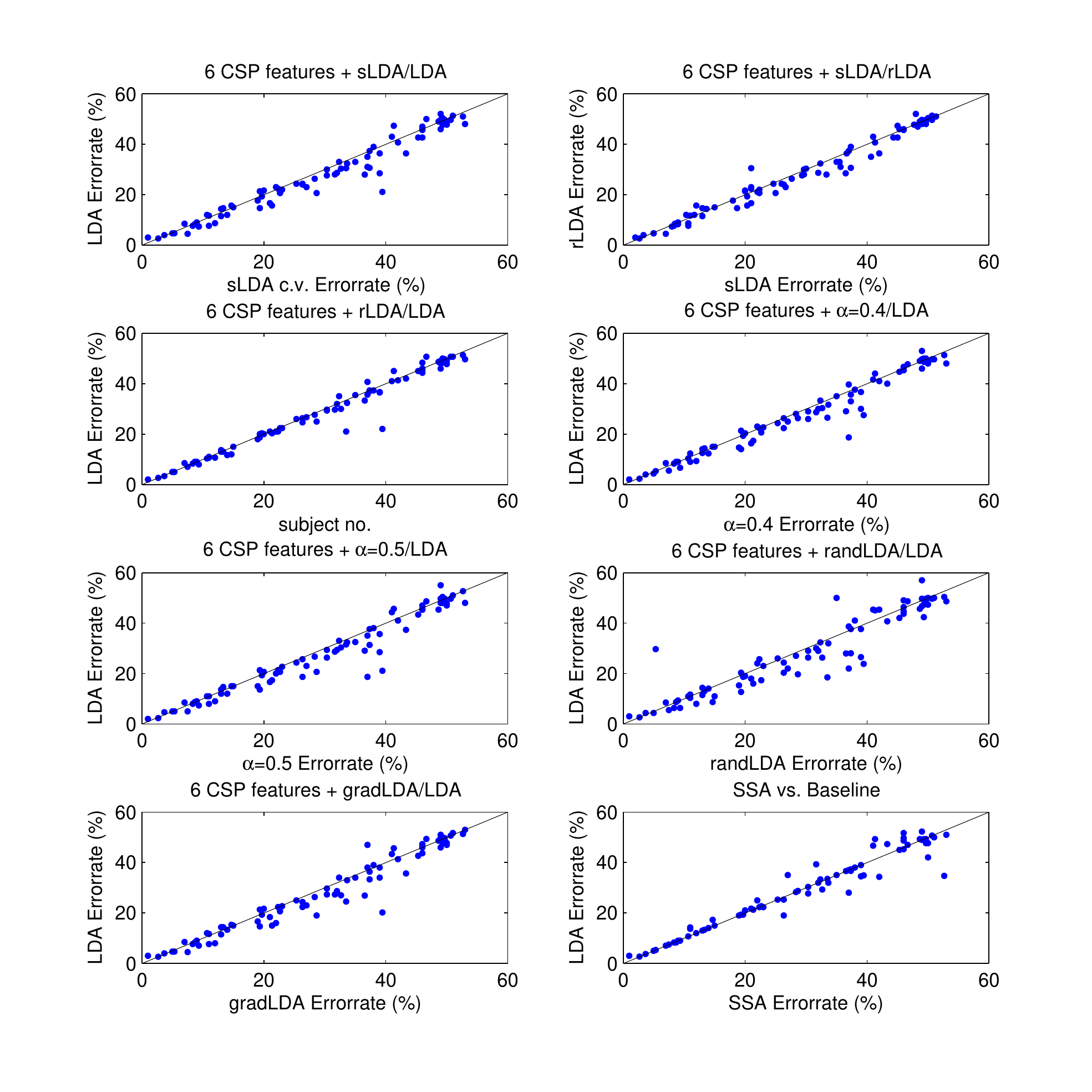}
  \caption{The figure displays results given as error-rates scatter plotted in percent between the methods, LDA, sLDA with $\alpha = 0.5$ and rLDA, gradLDA, randLDA and sLDA with cross validation, with CSP processing on the 80 BCI motor imagery subjects. A dot below the diagonal represents an increase in performance (for eg. sLDA over LDA). The comparison made is displayed in the title and each of the methods classification errors are displayed on the $x$ and $y$-axes. Each 
  dot corresponds to a single BCI subject. 
  The plots show that that the improvement of sLDA over LDA is most marked when the error rate of the subject under LDA is high, although improvements in performance are observed
  at lower error-rates also.
     \label{fig:scatter}}
 \end{center}
\end{figure}

%

\begin{table}
\begin{center}
\begin{tabular}{| c | c || l |}
\hline
Algorithm 1 & Algorithm 2 & $p$-value \\
\hline
\hline
c.v. sLDA & LDA & $p =7.60 \times 10^{-6}$ \\
\hline
rLDA &LDA & $p =1.96 \times 10^{-4}$ \\
\hline
c.v. sLDA & rLDA &  $p =0.0153$ \\
\hline
c.v. sLDA & gradLDA & $p =1$ \\
\hline
sLDA $\alpha = 0.4$ & LDA & $p =1.03 \times 10^{-5}$ \\
\hline
sLDA $\alpha = 0.5$ & LDA & $p = 1.76 \times 10^{-6}$ \\
\hline
gradLDA & LDA & $p = 2.05 \times 10^{-5}$ \\

\hline
sLDA $\alpha = 0.5$ & gradLDA & $p = 1$ \\
\hline
randLDA $\alpha = 0.6$ & LDA & $p =4.48 \times 10^{-4}$ \\
\hline
SSA & LDA & $p =1$ \\
\hline
\end{tabular}
\caption{\label{signrank}
$p$-values for the 80 BCI motor imagery data sets. The $p$-values given are obtained using a one sided paired Wilcoxon sign rank test using a Bonferroni correction factor $n=9$ equal to the number of tests performed. Each test tests the null hypothesis, effectively that the algorithm in the left hand column performs better than the algorithm in the middle column. Each row of the table corresponds to a single comparison between methods. The corresponding $p$-values are displayed in the right hand column.}
\end{center}
\end{table}

\begin{table}
\begin{center}
\begin{tabular}{| c | c || l |}
\hline
Algorithm 1 & Algorithm 2 & $p$-value \\
\hline
\hline
c.v.~sLDA &LDA & $p =5.04 \times 10^{-5}$ \\
\hline
rLDA & LDA & $p =0.0045$ \\
\hline
c.v.~sLDA & rLDA &  $p =0.0405$ \\
\hline
c.v.~sLDA & gradLDA& $p =1$ \\
\hline
sLDA $\alpha = 0.4$ & LDA & $p = 3.73 \times 10^{-5}$ \\
\hline
sLDA $\alpha = 0.5$ & LDA & $p = 1.94 \times 10^{-5}$ \\
\hline
sLDA $\alpha = 0.5$ & gradLDA & $p = 1$ \\
\hline
gradLDA & LDA & $p = 3.611 \times 10^{-5}$ \\
\hline
randLDA $\alpha = 0.6$ & LDA & $p =0.0684$ \\
\hline
SSA & LDA & $p =1$ \\
\hline
\end{tabular}
\caption{\label{ttest}
One sided paired $t$-tests using a Bonferroni correction factor $n=9$ (for the 80 BCI classification errors) equal to the number of tests performed. Each test tests the null hypothesis, effectively that the algorithm in the left hand column performs better than the algorithm in the middle column. Each
row corresponds to a comparison of methods. The corresponding $p$-values are displayed in the right hand column.}
\end{center}
\end{table}

\subsection{Discussion of Results}

The results displayed in Figure~\ref{fig:perchan1} show that application of SSA leads to no significant improvement in performance against the baseline method (CSP+LDA with $h_{CSP} = 6$), despite choice of the parameter $d_s$ (no. of stationary sources) by cross validation. Although SSA improves the performance on a subset of individual trials, preprocessing leads to no significant increase in overall performance of classification. 
Therefore we concentrate on comparison with the baseline method CSP+LDA henceforth.

Firstly, we note that both sLDA with cross validation, sLDA with $\alpha = 0.5$, sLDA with $\alpha = 0.4$, gradLDA and randLDA with $\alpha = 0.6$ lead to significant improvements ($p \ll 0.01$) in performance over the baseline method LDA and over rLDA; see Tables~\ref{signrank} and~\ref{ttest}.

Thus, we may conclude for these methods, that significant improvement is observed, not due to effects equivalent to covariance estimate regularization.

The method yielding the highest performance is sLDA with $\alpha = 0.5$. A few datasets are observed where gradLDA and randLDA show a decrease in performance but sLDA does not. However, the improvement of sLDA with $\alpha = 0.5$ over, for instance, gradLDA is not significant (see Table~\ref{ttest} and Table~\ref{signrank}).

Whether sLDA's incorporation of non-stationarity considerations, in fact, contributes to the observed improvement is also doubtful in light of the data analysis in Section~\ref{sec:dataanalysis}: here we observe that the non-stationary directions between training and test and the non-stationary direction within training reflect each other poorly. Suppose, therefore, that sLDA's effectiveness, here, does not rest on its stationarity incorporation:
we must suggest alternative explanations for the improvement observed.

The first possibility, on which improvement may be based, is regularization. If the data is correctly modeled as Gaussians, then covariance regularization using the optimal shrinkage parameter (i.e. rLDA) should be optimal, since LDA is optimal for Gaussians.
So the claim that improvement is solely due to regularization is equivalent to claiming that the Gaussian assumption is false.
That regularization is the sole explanation for improvement may be argued as implausible, in addition, when one takes the simulations from Section~\ref{sec:sanitysim} into account. In particular, when there
is more than one discriminative direction in the data set then regularization effects are lower. This is often the case for the data sets at hand, since the feature extraction step, CSP,
extracts a number of discriminative directions. Additionally, regularization effects are never observed to exceed 1$\%$ in simulation, whereas the performance increases observed, far exceed 1$\%$
(see Figure~\ref{fig:OO} for detailed results of our simulated investigation of regularization).

Thus, suppose that regularization does not fully explain improvement using sLDA, randLDA and gradLDA. Are there other possible explanations? 
One candidate explanation is that improvement depends on the BCI data specific non-stationarities between the training and test data. In particular we will observe in the data analysis of Section~\ref{sec:dataanalysis} that there are highly non-random dependencies between the most non-stationary directions and the most discriminative.

According to this possibility, there is a data dependent feature which implies that slight deviation from the direction chosen using LDA results in an increase in performance.
The increase in performance over LDA using gradLDA and randLDA using $\alpha = 0.6$ may be attributed to the slight difference in direction chosen using gradient based optimization.
The proposed explanation we suggest is that sLDA with gradient based optimization returns a slightly different direction for classification than that produced
by LDA, either due to local mimima or due to the stochastic nature of the gradient based approach. This slight difference in direction from the LDA solution accounts for
the increase in performance due to BCI data dependency.

Thus, we conclude the following:

\begin{enumerate}
\item sLDA produces a significant increase in performance over the baseline method.
\item There is possibly a data dependent feature under which performance increases are possible using algorithms which perturb LDA randomly.
\item sLDA outperforms sLDA with $\alpha =1$ but not significantly. ($p = 0.26$ paired one sided $t$-test; $p = 0.45$ one sided paired Wilcoxon sign-rank without correction for multiple testing --- see
Tables~\ref{ttest} and~\ref{signrank} for Bonferroni corrected $p$-values.)
\end{enumerate}
   
\subsection{Analysis of the 80 BCI datasets using group wise SSA} 
 \label{sec:dataanalysis}
In this section, we aim to gain insight into the non-stationarity present in BCI datasets.
The aim is to understand the improvement in performance observed above despite using putatively suboptimal classification methods for the task (for example randLDA and sLDA with $\alpha = 1$).
If non-stationarity is correlated with the direction of highest discriminability on the training data, then it may be the case that choosing a slightly different but nonetheless discriminative direction than chosen by LDA, 
results in a more stationary direction between training and test data.

In the first step of the analysis we compute the most non-stationary direction between the training and test data and the most non-stationary direction within the training data using group wise SSA \cite{GroupWiseSSA}.
The aim here is to gain insight into how consistent any attempt to infer the stationarity of directions between training and test data using only the training data can be.
The method used is as follows:

\begin{enumerate}
\item The data are fixed as the 6 CSP features from the experiments above.
\item The most non-stationary direction are computed on the training data using 7 epochs.
\item The most non-stationary direction between training and test data are computed using 
2 epochs, one for training and one for test.
\item The angle is computed between the two directions.
\end{enumerate}

In the next step of the analysis, we compute the angles between the most non-stationary direction between training and test and the direction chosen using LDA on the training data, as follows. 
 
\begin{enumerate}
\item The data are fixed as the 6 CSP features from the experiments above.
\item The LDA direction is computed within these features.
\item The most non-stationary direction between training and test data is computed within these features using 
2 epochs group wise SSA, one for training and one for test.
\item The angle is computed between the two directions.
\end{enumerate} 
 
\begin{figure}[ht]
 \begin{center}
  \includegraphics[width = 120mm]{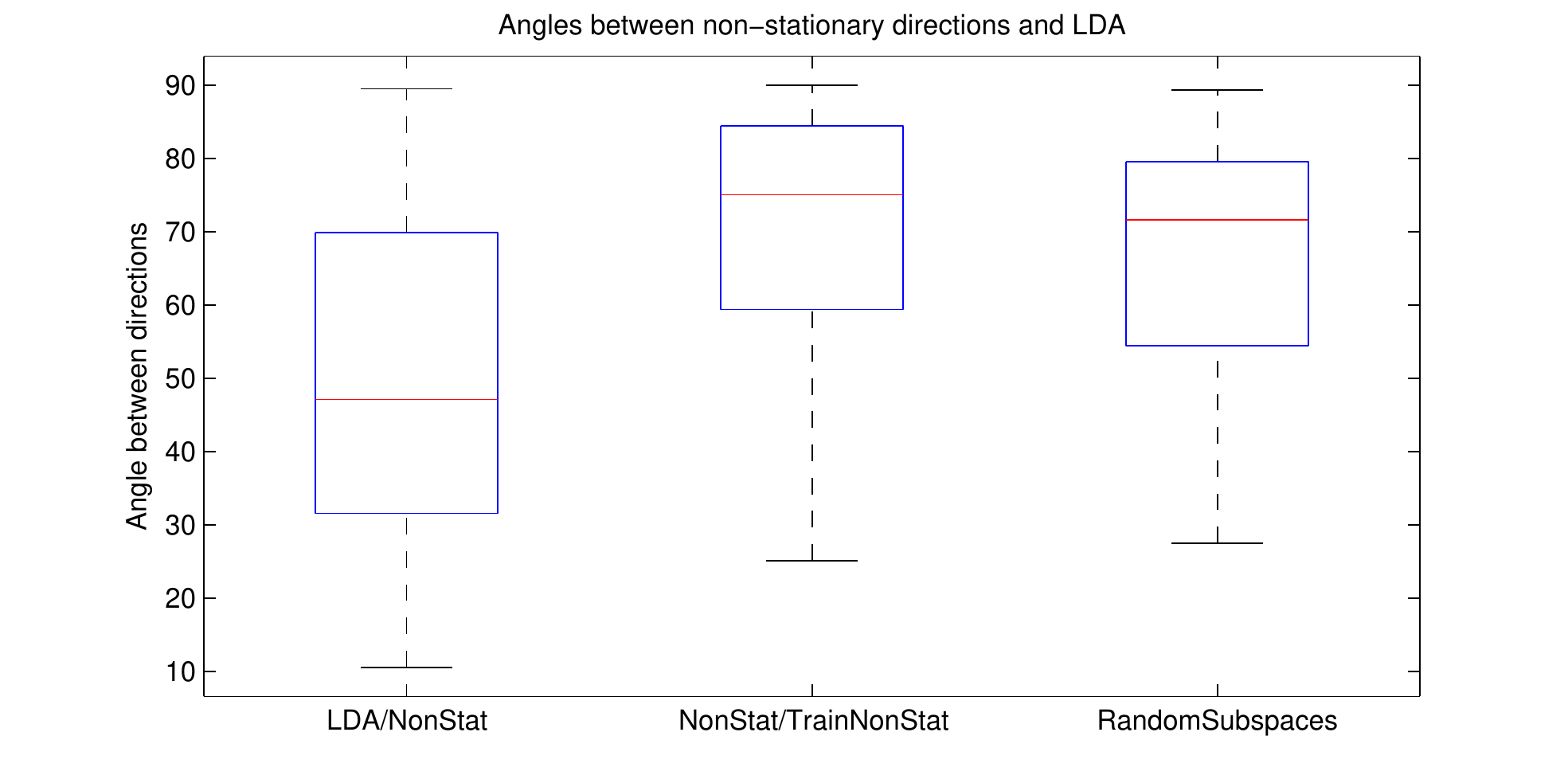}
  \caption{ The figure displays the angles as a boxplot (the whiskers describe the full extent of the data, the box corresponds to the lower quartile, median and upper quartile) for the 80 BCI motor imagery subjects between a selection of 1 dimensional subspaces contained within the CSP features. The left hand column represents the angle chosen by LDA on the training data and the direction maximizing the non-stationarity between training and test data. The second column displays the quantiles over the angles between the most non-stationary direction within the training data and the most non-stationary direction between training and test data. The third column displays the results one would expect if two subspaces are chosen at random, for comparison's sake. The results show that, there is little correlation between the non-stationarities within training and between training and test but that there is correlation between the
  the subspace chosen by LDA and the direction maximizing training to test non-stationarity. This leaves the particular suitability of sLDA for the BCI data sets doubtful.  \label{fig:NonStatAngles}}
 \end{center}
\end{figure}

The results are displayed in Figure \ref{fig:NonStatAngles}. Firstly, the results show that the attempt to infer distributions over non-stationarity is difficult at best purely on the basis of the training data.
The results show further that we may reject the hypothesis $\HH_0 = \text{\emph{The angle between the two directions is random}}$ at a highly significant confidence level using the Kolgomorov-Smirnov test ($p \ll 0.0001$) \cite{KSmirnoff} for testing whether the sample angles between the directions measured are drawn from the distribution over angles between two one-dimensional subspaces defined by vectors drawn from a 6 dimensional multivariate uniform distribution: i.e. the distribution defined on $[0,\frac{\pi}{2}]$ by the density function $p(\theta) = \frac{1}{Z} \text{sin}^4(\theta)$ \cite{PrincipleAngle}.
  
\subsection{Discussion of the Analysis}

The analysis above shows that the most non-stationary direction between the training and test data for BCI datasets often corresponds closely to the direction chosen by LDA.
This may explain why gradLDA improves performance in the BCI classification task: if we randomly perturb the direction given by LDA slightly, then we may obtain a more stationary direction
but nevertheless a discriminative direction. That is to say, gradLDA slightly perturbs the LDA solution in a random direction.

Moreover, a further potential explanation for gradLDA's improvement is the following: if the discriminative subspace of the data space which is stationary has higher dimensionality than the non-stationary subspace and we perturb a direction within this discriminative subspace,
then the odds are higher that we move further away from the non-stationary subspace than from the stationary subspace.

 A further, neuroscientific, question is: what constitutes, from a neurological point of view, the correlation between the most non-stationary direction and the most discriminative direction? A potential answer is that the test phase requires the engagement
 of a slightly different set of cognitive abilities than the training phase: for instance, during training, the subject \emph{sees} the cue to perform a particular motor imagery: see \cite{NeuroPred} for details of the experimental setup. Visual areas may thus
 provide information in training not available in testing and thus a difference in the most discriminative direction which presents a mode of non-stationarity which is not susceptible to 
 extraction from the training data alone. This implies that there may often be a non-stationary direction which coincides largely with the most discriminative direction on the training data. If this direction is simultaneously the \emph{most} non-stationary direction, then any deviation \emph{away} from the direction chosen by LDA is highly probably \emph{more} stationary. Although it may appear 
 that this data pecularity is merely an artifact of the experimental setup, the phenomenon, whereby different states of the subject, including discriminative information, occur in the training and test phases,
 is pertinent to the design of \emph{any} Brain Computer Interface. The task, therefore, remains, to examine the structure of this systematic difference for selected paradigms. 
  
\section{Outlook}

We have demonstrated that sLDA leads to significant improvements in performance for BCI classification over the standard approach which cannot be explained as the result of implicit eigenvalue-based regularization.

As we stated above, the aim in defining sLDA was to derive a linear classifier which is robust against non-stationarity under the assumption that the classes may be coarsely modeled
as Gaussians. We noted that specific assumptions must be fulfilled
for sLDA to work successfully: non-stationarity in the training set must be constant in topography between the test and training phases, but itself non-stationary in distribution over time. However
when fulfilled, we observed that sLDA leads to improvement in classification performance.

However, we have observed, in addition. that the improvement for BCI may be data dependent. In particular, we successfully showed that there are highly non-random patterns in the BCI non-stationarity. The challenge remains, therefore, to describe explain exactly what type of non-stationarity must be present to make such improvement possible and 
whether this understanding may be levered to design methods which yield higher improvements in the face of non-stationarity. The exciting task of understanding domain specific BCI non-stationarity
is also an inviting road for further research.

\chapter{Conclusion}

This thesis has contributed to research into non-stationarity in Machine Learning by proposing and testing two linear projection algorithms for dealing with non-stationarity. Both are inspired by Stationary Subspace Analysis.
The first algorithm maximizes non-stationarity under its projection of the dataset and is thus suited as a preprocessing method for high-dimensional sequential change-point detection, although
its application is not in principle limited as such: we conjecture, additionally, that maximizing non-stationarity will prove of interest in the analysis of neural data resulting from learning paradigm experimental settings.
The second algorithm proposed yields improvements under non-stationarity in 
Brain Computer Interfaces. Although we cannot conclude that the improvement in performance observed is a results of non-stationarity penalization, we have identified, using SSA and the method developed, sLDA, features of BCI datasets which may represent valuable prior information in 
coping with non-stationarity.

\addcontentsline{toc}{chapter}{Bibliography}
\bibliography{supervised,stable,segmentation,ssa_pubs}        
\bibliographystyle{plain}  
\end{document}

%% file: ssa_defs.tex
\usepackage{bm}
\usepackage{amsmath}
\usepackage{amssymb}  

\usepackage{dsfont}

\usepackage[normalem]{ulem}

\usepackage{dsfont}

\newcommand{\R}{\ensuremath{\mathds{R}}}		

\newcommand{\Gauss}{\ensuremath{\mathcal{N}}}	

\newcommand{\s}{\ensuremath{\mathfrak{s}}}		
\newcommand{\n}{\ensuremath{\mathfrak{n}}}		

\DeclareMathOperator*{\E}{\mathds{E}}					
\DeclareMathOperator*{\Var}{Var}				
\DeclareMathOperator*{\Cov}{Cov}				
\DeclareMathOperator*{\argmin}{argmin}		    
\DeclareMathOperator*{\argmax}{argmax}		    

\newcommand{\KLD}{D_{\text{KL}}}				
\newcommand{\esi}{\hat{\Sigma}}					
\newcommand{\emu}{{\ensuremath{\hat{\mu}}}}	

\newcommand{\NN}{\mathbb{N}}


%% file: Thesis.bbl
\begin{thebibliography}{10}

\bibitem{PrincipleAngle}
P.A. Absil, A.~Edelman, and P.~Koev.
\newblock On the largest principal angle between random subspaces.
\newblock {\em Linear Algebra and its Applications}, 414:288--294, 2006.

\bibitem{Speech1}
R.~Andre-Obrecht.
\newblock A new statistical approach for the automatic segmentation of
  continous speech signals.
\newblock {\em IEEE Trans.~Acoustics, Speech, Signal Processing},
  ASSP-36(1):29--40, 1988.

\bibitem{Spectral}
U.~Appelai and A.V. Brandta.
\newblock Adaptive sequential segmentation of piecewise stationary time series.
\newblock {\em Information Sciences}, 29:27--56, 1983.

\bibitem{ChangePoint}
M~Basseville and I.V. Nikiforov.
\newblock {\em Detection of Abrupt Changes - Theory and Application}.
\newblock Prentice-Hall, Inc., Englewood Cliffs, N.J., 1993.

\bibitem{oai:eprints.pascal-network.org:3317}
B.~Blankertz, M.~Kawanabe, R.~Tomioka, F.~Hohlefeld, V.~Nikulin, and K.R.
  M{\"u}ller.
\newblock Invariant common spatial patterns: Alleviating nonstationarities in
  brain-computer interfacing.
\newblock {\em Advances in Neural Information Processing Systems}, 20, 2008.

\bibitem{rLDA}
B.~Blankertz, S.~Lemm, M.~Treder, S.~Haufe, and K.-R. M{\"u}ller.
\newblock Single-trial analysis and classification of {ERP} components - a
  tutorial.
\newblock {\em Neuroimage}, 2011.

\bibitem{NeuroPred}
B.~Blankertz, C.~Sannelli, S.~Halder, E.M. Hammer, A.~K\"ubler, K.R. M\"uller,
  G.~Curio, and T.~Dickhaus.
\newblock Neurophysiological predictor of smr-based bci performance.
\newblock {\em Neuroimage}, 51:1303--1309, 2010.

\bibitem{SpatFilt}
B.~Blankertz, R.~Tomioka, S.~Lemm, M.~Kawanabe, and K.-R. M\"uller.
\newblock Optimizing spatial filters for robust eeg single-trial analysis.
\newblock {\em Signal Processing Magazine, IEEE}, 21:41 -- 56, 2008.

\bibitem{oai:biomedcentral.com:1471-2202-10-S1-P85}
B.~Blankertz and C.~Vidaurre.
\newblock Towards a cure for {BCI} illiteracy: machine learning based
  co-adaptive learning.
\newblock {\em BMC Neuroscience}, page~85, July~13 2009.

\bibitem{CPpreprint}
D.A.J. Blythe, P.~von B{\"u}nau, F.C Meinecke, and K.R. M{\"u}ller.
\newblock Feature extraction for high dimensional change point detection using
  stationary subspace analysis.
\newblock {\em Online Pre-Print}, 2011.
\newblock arXiv:1108.2486.

\bibitem{Pattern}
F.L. Chung, T.C. Fu, Ng.V., and R.W.P. Luk.
\newblock An evolutionary approach to pattern-based time series segmentation.
\newblock {\em IEEE Transactions on evolutionary computation}, 8:471--489,
  2004.

\bibitem{bb57382}
C.~Cortes and V.~Vapnik.
\newblock Support-vector networks.
\newblock {\em Machine Learning}, 20(3):273--297, 1995.

\bibitem{Econ1}
H.~Cs\"org\"o and L.~Horv{\'a}rth.
\newblock Nonparametric methods for change point problems.
\newblock In P.R. Krishnaiah and C.R. Rao, editors, {\em Handbook of
  statistics}, volume~7, pages 403--425. Elsevier, New York, 2009.

\bibitem{KSmirnoff}
W.T. Eadie, D.~Drijard, F.E. James, M.~Roos, and B.~Sadoulet.
\newblock {\em Statistical Methods in Experimental Physics.}
\newblock Amsterdam: North Holland, 1971.

\bibitem{LDA}
R.~Fisher.
\newblock The use of multiple measurements in taxonomic problems.
\newblock {\em Annals Eugen.}, 7:170--188, 1937.

\bibitem{Foerstner:1994q}
W.~Foerstner.
\newblock A framework for low level feature extraction.
\newblock {\em Computer Vision-ECCV'94-Springer}, 801:383--394, 1994.

\bibitem{Fault1}
P.M. Frank.
\newblock Fault diagnosis in dynamic systems using analytical and knowledge
  based redundency - a survey and new ressults.
\newblock {\em Automatica}, 26:459--474, 1990.

\bibitem{Gower69SingleLinkage}
J.~C. Gower and G.~J.~S. Ross.
\newblock Minimum spanning trees and single linkage cluster analysis.
\newblock {\em Journal of the Royal Statistical Society}, 18(1):54--64, 1969.

\bibitem{Biomed2}
D.E. Gustafson, A.S. Willsky, J.Y. Wang, M.C. Lancaster, and J.H. Triebwasser.
\newblock {ECG/VCG} rhythm diagnosis using statistical signal analysis. part i:
  Identification of persistent rhythms. part {II}: Identification of transient
  rhythms.
\newblock {\em IEEE Trans. Biomedical Engineering}, BME-25:344--353, 353--361,
  1978.

\bibitem{Guyon:2003q}
I.~Guyon and A.~Elisseeff.
\newblock An introduction to variable and feature selection.
\newblock {\em Journal of Machine Learning Research}, 3:1157--1182, 2003.

\bibitem{BayesOptimal}
O.C. Hamsici and A.M. Martinez.
\newblock Bayes optimality in linear discriminant analysis.
\newblock {\em IEEE Transactions on Pattern Analysis and Machine Intelligence},
  30:647--657, 2008.

\bibitem{HarKawWasBun10SSA}
S.~Hara, Y.~Kawahara, T.~Washio, and P.~von B\"{u}nau.
\newblock Stationary subspace analysis as a generalized eigenvalue problem.
\newblock In {\em Proceedings of the 17th international conference on Neural
  information processing: theory and algorithms - Volume Part I}, ICONIP'10,
  pages 422--429, Berlin, Heidelberg, 2010. Springer-Verlag.

\bibitem{HaykinSig}
S.~Haykin.
\newblock {\em Kalman Filtering and Neural Networks (Adaptive and Learning
  Systems for Signal Processing, Communications and Control)}.
\newblock John Wiley and Sons, 605 Third Avenue, New York, NY, 10158-0012,
  2001.

\bibitem{ICABook}
A.~Hyv\"arinen, J.~Karhunen, and E.~Oja.
\newblock {\em Independent Component Analysis}.
\newblock Wiley, New York, 2001.

\bibitem{DistBased1}
A.~Iyer, U.~Ofoegbu, R.~Yantorno, and B.~Smolenski.
\newblock Speaker distinguishing distances: a comparative study.
\newblock {\em International Journal of Speech Technology}, 10:95--107, 2009.

\bibitem{DistBased4}
Y.~Kawahara and M.~Sugiyama.
\newblock Change-point detection in time-series data by density-ratio
  estimation.
\newblock In {\em Proceedings of the 9th {SIAM} Int.~Conf.~on Data Mining},
  2009.

\bibitem{InfoSSA}
M.~Kawanabe, W.~Samek, P.~von B{\"u}nau, and F.C. Meinecke.
\newblock An information geometrical view of stationary subspace analysis.
\newblock {\em Lecture Notes in Computer Science}, 6792:397--404, 2011.

\bibitem{JMLRpreprint}
F.J. Kiraly, P.~von B{\"u}nau, F.C. Meinecke, D.A.J. Blythe, and K.R.
  M{\"u}ller.
\newblock Algebraic geometric comparison of probability distributions.
\newblock {\em Online Pre-Print}, 2011.
\newblock arXiv:1108.1483.

\bibitem{Kohlmorgen:2003fk}
J.~Kohlmorgen.
\newblock On optimal segmentation of sequential data.
\newblock {\em Proceedings of the 13th IEEE workshop on Neural Networks for
  Signal Processing}, pages 449 -- 458, 2003.

\bibitem{DistBased2}
J.~Kohlmorgen and S.~Lemm.
\newblock A dynamic {HMM} for on-line segmentation of sequential data.
\newblock In T.G. Dietterich, S.~Becker, and Z.~Ghahramani, editors, {\em
  Advances in Neural Information Processing Systems 14}, pages 793--800. MIT
  Press, 2002.

\bibitem{KohlmorgenKybernetics}
J.~Kohlmorgen, J.~Rittweger, and K.~Pawelzik.
\newblock Identication of nonstationary dynamics in physiological recordings.
\newblock {\em Biological Cybernetics}, 83:73--84, 2000.

\bibitem{LeCun90GG8}
Y.~{LeCun}, B.~Boser, J.S. Denker, D.~Henderson, R.E. Howard, W.~Hubbard, and
  L.D. Jackel.
\newblock Handwritten digit recognition with a back-propagation network.
\newblock In D.~Touretzky, editor, {\em Neural Information Processing Systems},
  volume~2. Morgan Kaufman, 1990.

\bibitem{Lewis:1991q}
D.D. Lewis.
\newblock Feature selection and feature extraction for text categorization.
\newblock {\em HLT '91 Proceedings of the workshop on Speech and Natural
  Language}, pages 212--217, 1991.

\bibitem{Li:2006q}
H.~Li, T.~Jiang, and K.~Zhang.
\newblock Efficient and robust feature extraction by maximum margin criterion.
\newblock {\em IEEE Transactions on Neural Networks}, pages 157 -- 165, 2006.

\bibitem{Geophysics}
W.~Menke.
\newblock {\em Geophysical Data Analysis: Discrete Inverse Theory}.
\newblock Academic Press, 1989.

\bibitem{Morris:2005q}
J.S. Morris, K.R. Coombes, J.~Koomen, E.A. Baggerly, and R.~Kobayashi.
\newblock Feature extraction and quantification for mass spectrometry in
  biomedical applications using the mean spectrum.
\newblock {\em Bioinformatics}, 21:1764--1775, 2005.

\bibitem{MurataOnline}
Noboru Murata, Motoaki Kawanabe, Andreas Ziehe, Klaus-Robert M{\"u}ller, and
  Shun ichi Amari.
\newblock On-line learning in changing environments with applications in
  supervised and unsupervised learning.
\newblock {\em Neural Networks}, 15(4-6):743--760, 2002.

\bibitem{Narendra}
K.S. Narendra and J.~Balakrishnan.
\newblock Improving transient response of adaptive control systems using
  multiple models and switching.
\newblock {\em IEEE Transactions on Automatic Control}, 39:1861--1866, 1994.

\bibitem{EEGPrin}
E.~Niedermeyer and F.~H. Lopes~da Silva.
\newblock {\em Electroencephalography: basic principles, clinical applications,
  and related fields}.
\newblock Lippincott, Williams and Wilkins, 530 Walnut Street, Philadephia, PA
  19106 USA, 2005.

\bibitem{TricksOfTheTrade}
G.~B. Orr and K.-R.M{\"u}ller, editors.
\newblock {\em Neural Networks: Tricks of the Trade, this book is an outgrowth
  of a 1996 {NIPS} workshop}, volume 1524 of {\em Lecture Notes in Computer
  Science}. Springer, 1998.

\bibitem{Page:1954fk}
E.S. Page.
\newblock Continuous inspection schemes.
\newblock {\em Biometrika}, 41(1/2):100--115, 1954.

\bibitem{oai:repository.ust.hk:1783.1/6830}
S.J. Pan, I.W. Tsang, J.T. Kwok, and Q.~Yang.
\newblock Domain adaptation via transfer component analysis.
\newblock {\em IEEE Transactions on Neural Networks}, pages 1187--1192, July
  2009.

\bibitem{Plu05}
M.D. Plumbley.
\newblock Geometrical methods for non-negative {ICA}: Manifolds, {L}ie groups
  and toral subalgebras.
\newblock {\em Neurocomputing}, 67(161--197), 2005.

\bibitem{Pri83Spectral}
M.~B. Priestley.
\newblock {\em Spectral Analysis and Time Series}.
\newblock Academic Press, 1983.

\bibitem{Speech2}
L.R. Rabiner.
\newblock A tutorial on hidden markov models and selected applications in
  speech recognition.
\newblock In A.~Waibel and K.~Lee, editors, {\em Readings in Speech
  Recognition}, pages 267--296. Morgan Kaufmann, 1990.

\bibitem{AdaptNN}
L.~Rutkowski.
\newblock Adaptive probabilistic neural networks for pattern classification in
  time-varying environment.
\newblock {\em IEEE Transactions on Neural Networks}, 15, 2004.

\bibitem{Saeys:2007q}
Y.~Saeys, I.~Inza, and P.~Larranaga.
\newblock A review of feature selection techniques in bioinformatics.
\newblock {\em Bioinformatics}, 23:2507--2517, 2007.

\bibitem{GroupWiseSSA}
W.~Samek, M.~Kawanabe, and C.~Vidaurre.
\newblock Group-wise stationary subspace analysis - a novel method for studying
  non-stationarities.
\newblock In {\em Proceedings of the 5th International BCI Conference - Graz},
  2011.

\bibitem{Schoelkopf:1998q}
B.~Sch\"{o}lkopf, A.~Smola, and K.-R. M\"{u}ller.
\newblock Nonlinear component analysis as a kernel eigenvalue problem.
\newblock {\em Neural Computation}, 10:1299--1319, 1998.

\bibitem{CovariateShift}
M.~Sugiyama, K.~Krauledat, and K.-R. M{\"u}ller.
\newblock Covariate shift adaptation by importance weighted cross validation.
\newblock {\em Journal of Machine Learning Research}, 8:985--1005, 2000.

\bibitem{Torkkolla:2003q}
K.~Torkkolai.
\newblock Feature extraction by non-parametric mutual information maximization.
\newblock {\em Journal of Machine Learning Research}, 3:1415--1438, 2003.

\bibitem{bb58133}
V.~Vapnik.
\newblock {\em Statistical Learning Theory}.
\newblock Wiley, 1998.

\bibitem{journals/neco/VidaurreSMB11}
C.~Vidaurre, C.~Sannelli, K.-R. M{\"u}ller, and B.~Blankertz.
\newblock Machine-learning-based coadaptive calibration for brain-computer
  interfaces.
\newblock {\em Neural Computation}, 23(3):791--816, 2011.

\bibitem{PRL:SSA:2009}
P.~von B\"unau, F.C. Meinecke, F.J. Kir\'aly, and K.-R. M\"uller.
\newblock Finding stationary subspaces in multivariate time series.
\newblock {\em Phys. Rev. Lett.}, 103(21):214101, Nov 2009.

\bibitem{Likelihood}
S.S. Wilks.
\newblock The large-sample distribution of the likelihood ratio for testing
  composite hypotheses.
\newblock {\em Ann. Math. Statist.}, 9(1):60--62, 1938.

\bibitem{5946469}
W.~Wojcikiewicz, C.~Vidaurre, and M.~Kawanabe.
\newblock Stationary common spatial patterns: Towards robust classification of
  non-stationary eeg signals.
\newblock In {\em Acoustics, Speech and Signal Processing (ICASSP), 2011 IEEE
  International Conference on}, pages 577 --580, may 2011.

\bibitem{ZieheBio}
A.~Ziehe, K.-R. Muller, G.~Nolte, B.-M. Mackert, and G.~Curio.
\newblock Artifact reduction in magnetoneurography based on time-delayed
  second-order correlations.
\newblock {\em IEEE Transactions on Biomedical Engineering}, 47:75 -- 87, 2000.

\bibitem{EarlyStopping}
H.~Zou and T.~Hastie.
\newblock Regularization and variable selection via the elastic net.
\newblock {\em Journal of the Royal Statistical Society}, 67:301--320, 2005.

\end{thebibliography}
